\documentclass[10pt,twocolumn,letterpaper]{article}
\usepackage{cvpr}              

\usepackage{graphicx}
\usepackage[most]{tcolorbox} 

\usepackage{amsmath}    

\usepackage{circledsteps}
\pgfkeys{/csteps/inner color=white}
\pgfkeys{/csteps/outer color=green!50!black}
\pgfkeys{/csteps/fill color=green!50!black}

\newcommand{\hlcyan}[1]{{\sethlcolor{cyan}\hl{#1}}}
\newcommand{\hlcyanlow}[1]{{\sethlcolor{cyan!12}\hl{#1}}}
\newcommand{\hlredhigh}[1]{{\sethlcolor{red!55}\hl{#1}}}
\definecolor{darkred}{RGB}{154,23,8}  
\definecolor{lightorange}{RGB}{255,200,150} 
\definecolor{darkpink}{RGB}{255,153,153} 
\definecolor{darkviolet}{RGB}{94,23,235}  
\definecolor{lightbrown}{RGB}{45,37,35}  

\usepackage{colortbl}
\usepackage{varwidth}
\usepackage{CJKutf8}  
\usepackage{caption}   
\usepackage{float}     
\usepackage{calc}
\usepackage{balance}

\definecolor{Gray}{gray}{0.85}
\definecolor{SkyBlue}{rgb}{0.88,1,1}
\definecolor{PasteGreen}{RGB}{204, 226, 215}
\definecolor{PastePink}{RGB}{253, 223, 236}

\newcolumntype{a}{>{\columncolor{Gray}}c}

\definecolor{LightYellow}{RGB}{252, 255, 125}
\definecolor{PasteYellow}{RGB}{254, 255, 214}
\definecolor{PasteLavender}{RGB}{229, 228, 244}
\definecolor{NewGreen}{RGB}{17,207,156}

\newcommand{\hlGray}[1]{{\sethlcolor{Gray}\hl{#1}}}
\newcommand{\hlpastelavender}[1]{{\sethlcolor{PasteLavender}\hl{#1}}}
\newcommand{\hlpastegreen}[1]{{\sethlcolor{PasteGreen}\hl{#1}}}
\newcommand{\hlpastepink}[1]{{\sethlcolor{PastePink}\hl{#1}}}

\newcommand{\hlpink}[1]{{\sethlcolor{darkpink}\hl{#1}}}

\usepackage{multirow}
\usepackage{multicol}
\usepackage{makecell}
\usepackage{pifont}
\usepackage{textcomp}
\usepackage{amssymb}

\usepackage{booktabs}
\usepackage{array}
\definecolor{heatmap1}{RGB}{255,103,121}     
\definecolor{heatmap2}{RGB}{255,121,109}
\definecolor{heatmap3}{RGB}{255,135,99}
\definecolor{heatmap4}{RGB}{255,147,88}
\definecolor{heatmap5}{RGB}{255,161,75}
\definecolor{heatmap6}{RGB}{255,174,65}
\definecolor{heatmap7}{RGB}{255,186,56}
\definecolor{heatmap8}{RGB}{255,196,47}
\definecolor{heatmap9}{RGB}{255,205,41}   
\definecolor{heatmap10}{RGB}{255,217,37}
\definecolor{heatmap11}{RGB}{255,227,34}
\definecolor{heatmap12}{RGB}{241,233,41}
\definecolor{heatmap13}{RGB}{224,239,47}
\definecolor{heatmap14}{RGB}{202,246,56}
\definecolor{heatmap15}{RGB}{184,251,64}
\definecolor{heatmap16}{RGB}{162,255,74}
\definecolor{heatmap17}{RGB}{135,255,86}    
\definecolor{heatmap18}{RGB}{111,255,96}
\definecolor{heatmap19}{RGB}{71,255,106}
\definecolor{heatmap20}{RGB}{0,255,130}

\newcommand{\roadsociallogo}[1]{\includegraphics[scale={#1}] {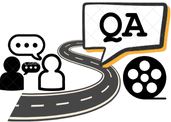}}
\usepackage{soul}
\usepackage{tabularx}
\definecolor{applegreen}{rgb}{0.55, 0.71, 0.0}
\definecolor{atomictangerine}{rgb}{1.0, 0.6, 0.4}

\definecolor{darkgreen}{rgb}{0.0, 0.5, 0.0}

\newcommand{\greenCircledNumber}[1]{%
    \tikz[baseline={(0,-.3em)}]{
        \node[circle, fill=darkgreen, inner sep=0.8pt, minimum size=1.0em, text=white, font=\footnotesize] {#1};
    }%
}

\newcommand{\blackCircledNumber}[1]{%
    \tikz[baseline={(0,-.3em)}]{
        \node[circle, fill=black!75, inner sep=0.8pt, minimum size=1.0em, text=white, font=\footnotesize] {#1};
    }%
}

\definecolor{cvprblue}{rgb}{0.21,0.49,0.74}
\tcbuselibrary{breakable,skins} 

\usepackage[pagebackref,breaklinks,colorlinks,citecolor=cvprblue]{hyperref}
\usepackage[normalem]{ulem}
\usepackage{tocloft}   

\hypersetup{
    colorlinks=true,
    linkcolor=blue,
    filecolor=magenta,      
    urlcolor=magenta,
}

\definecolor{headerback}{RGB}{70,70,70}    
\definecolor{headerblue}{RGB}{25,75,150}   
\definecolor{headergreen}{RGB}{0,100,60}   
\definecolor{headertext}{RGB}{255,255,255} 

\def\paperID{9496} 
\def\confName{CVPR}
\def\confYear{2025}

\title{
\vspace{-2mm} \roadsociallogo{0.25}
RoadSocial: A Diverse VideoQA Dataset and Benchmark for Road Event Understanding from Social Video Narratives}

\author{Chirag Parikh\thanks{Equal contribution.}, Deepti Rawat\footnotemark[1], Rakshitha R. T., Tathagata Ghosh, and Ravi Kiran Sarvadevabhatla\\
CVIT \& iHub-Data, IIIT Hyderabad, India\\
{\small\url{https://roadsocial.github.io}}
}

\begin{document}

\maketitle

\begin{abstract}
We introduce RoadSocial, a large-scale, diverse VideoQA dataset tailored for generic road event understanding from social media narratives. Unlike existing datasets limited by regional bias, viewpoint bias and expert-driven annotations, RoadSocial captures the global complexity of road events with varied geographies, camera viewpoints (CCTV, handheld, drones) and rich social discourse. Our scalable semi-automatic annotation framework leverages Text LLMs and Video LLMs to generate comprehensive question-answer pairs across 12 challenging QA tasks, pushing the boundaries of road event understanding. RoadSocial is derived from social media videos spanning 14M frames and 414K social comments, resulting in a dataset with 13.2K videos, 674 tags and 260K high-quality QA pairs.
We evaluate 18 Video LLMs (open-source and proprietary, driving-specific and general-purpose) on our road event understanding benchmark. We also demonstrate RoadSocial's utility in improving road event understanding capabilities of general-purpose Video LLMs.
\end{abstract}

\section{Introduction}

A road event typically refers to any incident, activity, or condition occurring on or around the roadway that affects traffic flow, safety, or road usage. The ability to recognize and interpret road events is essential for safe and reliable intelligent vehicles and transportation systems. In this regard, large-scale video datasets of road events are used to develop assistive models~\cite{kim2018textual, ramanishka2018toward,Parikh2024IDDXAM, chandra2023meteor, singh2022road, caesar2020nuscenes}. Many recent datasets contain videos with accompanying question-answer text pairs and other text metadata~\cite{marcu2312lingoqa, qian2024nuscenes, Qasemi2023TrafficDomainVQ, sima2023drivelm}. Such datasets have become a de facto choice for training Video Large Language Models (Video LLMs)~\cite{zhou2024embodied, sima2023drivelm, ma2023dolphins, Gopalkrishnan2024MultiFrameL}.

However, current video-based road event understanding approaches are limited by region-specific datasets, neglecting the diversity of global road scenarios. Most datasets focus on dashcam views for autonomous driving, overlooking other camera types such as CCTV, handheld, and drone-based. They also lack annotations on generic events (\eg defensive driving, near-misses). Due to the reliance on regionally-biased expert annotators, the broader and richer contextual insights from real-world social discourse on road events are absent.
Furthermore, existing evaluation frameworks fail to test the Video LLMs' ability to distinguish informative road event details from misleading information, essential for developing reliable, hallucination-resistant road event understanding systems.

To address these limitations and to enable foundational video language models for \textit{generic} road event understanding, we introduce \textbf{RoadSocial}, a large-scale and diverse Video Question Answer (VideoQA) dataset. RoadSocial is obtained by processing social media videos and the narratives accompanying these videos. The inherent diversity of social media in terms of geographical locations, camera viewpoints, road event types and social commentary addresses shortcomings of video datasets mentioned previously. Specifically, we make the following contributions:

\begin{table*}[!ht]
\centering
\footnotesize
\begin{tabular}{lcccccccccccccc}
\toprule
Dataset & \makecell{Viewpoint \\ Type} & \makecell{Video \\ Frames} & \makecell{Duration \\ (mins)} & \makecell{Social \\ Comments} & Countries & \makecell{Video \\ Tags} & QAs & \makecell{TG \\ QA} & \makecell{AV \\ QA}  & \makecell{IC \\ QA} &  \makecell{Loc. \\ QA} &  \makecell{Internet \\ sourced} \\
\midrule
\rowcolor{LightYellow} \textcolor{blue}{\textbf{RoadSocial (Ours)}} & \textbf{6} & \textbf{14M} & \textbf{7.9K} & \textbf{414K} & \textbf{100} & \textbf{674} & \textbf{260K} & \ding{52} & \ding{52} & \ding{52} & \ding{52} & \ding{52} \\
\textcolor{blue}{Lingo-QA}~\cite{marcu2312lingoqa} & 1 & $\;$.1M & 1.8K & - & 1 & 7 & 419K & - & -& - & - & - \\
\textcolor{blue}{SUTD-TrafficQA}~\cite{xu2021sutd} & 3 & 10M & 6.7K & - & $<$4 & - & 62.5K & - & -& - & - & \ding{51} \\
\textcolor{blue}{DRAMA}~\cite{malla2023drama} & 1 & .02M & $\;$.6K & - & 1 & - & 102K & - & -& - & - & - \\
\textcolor{blue}{Rank2Tell}~\cite{rank2tell} & 1 & .02M & 39 & - & 1 & - & $>$118 & - & -& - & - & - \\
\textcolor{orange}{ROAD}~\cite{singh2022road} & 1 & $\;$.1M & 170 & - & 1 & 43 & - & - & -& - & - & - \\
\textcolor{orange}{MM-AU}~\cite{fang2024abductive} & 1 & 2.2M & 1.2K & - & $>$50 & 58 & 58.6K & \ding{51} & -& - & - & \ding{51} \\
\textcolor{orange}{DriveLM}~\cite{sima2023drivelm, zhou2024embodied} & 1 & .03M & 5.7K & - & 43 & - & $\;$375K & - & -& - & - & - \\
\textcolor{orange}{BDD-OIA}~\cite{xu2020explainable} & 1 & .02M & 1.9K & - & 1 & 25 & - & - & -& - & - & - \\
\textcolor{orange}{BDD-X}~\cite{kim2018textual} & 1 & 8.4M & 4.6K & - & 1 & - & - & - & -& - & - & - \\
\bottomrule
\end{tabular}
\caption{\textbf{Comparison of RoadSocial with existing road event understanding datasets.} TG: Temporal Grounding, AV: Adversarial, IC: Incompatible, Loc: Geographical Location. Internet-sourced videos do not contain LiDAR or CAN bus data. \textcolor{orange}{Orange}: Additional annotations added to existing datasets. \textcolor{blue}{Blue}: New datasets.}
\label{tab:dataset-comparison}
\end{table*}

\begin{itemize}    
    \item RoadSocial: a large-scale, diverse VideoQA resource for road events, derived from social media videos spanning \textbf{14M} frames and \textbf{414K} social comments, resulting in a dataset with \textbf{13.2K} videos, \textbf{674} unique tags, and \textbf{260K} high-quality QA pairs.
    \item A semi-automatic annotation framework using Text LLM and Video LLM that processes social media video narratives and generates comprehensive QA  pairs across 12 distinct challenging tasks.
    \item A robust evaluation framework incorporating \textit{non-road event} videos and irrelevant questions to assess the robustness of Video LLMs to hallucinations.   
    \item A demonstration of RoadSocial's utility in improving road event understanding capabilities of general-purpose Video LLM.
    \item Critical insights into 18 Video LLMs' performance on road event understanding, obtained from their evaluation on our RoadSocial-QA benchmark. 
\end{itemize}

\section{Related Works}
\label{sec:related_works}

Several video datasets describe road events through actions of surrounding entities~\cite{singh2022road,chandra2023meteor}, interactions between traffic participants~\cite{malla2023drama,Parikh2024IDDXAM,xu2020explainable,rank2tell}, or explanations of normal or safety-critical driving scenarios~\cite{kim2018textual,sima2023drivelm,marcu2312lingoqa}, including dangerous driving behaviors or accidents~\cite{xu2021sutd,fang2024abductive}. 

However, the diversity of these datasets is often limited by their geographical scope. Although some datasets~\cite{zhou2024embodied,fang2024abductive, xu2021sutd} include crowd-sourced videos from a range of locations, their textual annotations reflect local expertise, which may lack a comprehensive understanding of global traffic norms and behaviors. In contrast, our dataset is sourced from global social media video posts which addresses this shortcoming. Existing works typically rely on a pool of manual annotators, a process that is labor-intensive and lacks scalability~\cite{xu2021sutd,sima2023drivelm,fang2024abductive,marcu2312lingoqa,xu2020explainable}. We propose a scalable, semi-automatic annotation framework that leverages the capabilities of powerful Video LLMs and Text LLMs to process social media content from around the world and generate high-quality QA pairs associated with road event videos. 

Existing video language models built on previous road understanding benchmarks are often trained on specific camera viewpoints, usually vehicle-mounted~\cite{zhou2024embodied, fang2024abductive, marcu2312lingoqa, rank2tell, malla2023drama, sima2023drivelm, Tian2024DriveVLMTC} or CCTV~\cite{Kong2024WTSAP}. Such models may not generalize well across different viewpoint types and geographical regions, limiting their effectiveness for understanding road events in a broad context. In contrast, our dataset contains videos captured in diverse and uncontrolled camera settings (drone, handheld, CCTV etc.) across the world. Coupled with social discourse, our dataset is a viable alternative.

Existing VideoQA datasets focus primarily on ego-centric tasks~\cite{fang2024abductive,marcu2312lingoqa,sima2023drivelm,malla2023drama,rank2tell,zhou2024embodied}, limiting perspectives to the ego-vehicle. While Xu \etal~\cite{xu2021sutd} explore complex traffic scenarios, none of the existing works assess model robustness against misleading inputs or hallucinations. We address this gap by introducing novel QA tasks to evaluate (a) robustness to hallucinations with non-road-event videos and irrelevant questions (b) comprehension across camera viewpoint types and (c) geographical awareness. These tasks enable holistic Video LLM evaluation for \textit{general-purpose} road event understanding. A comparison of RoadSocial with existing datasets is shown in ~\cref{tab:dataset-comparison}.


\section{RoadSocial Dataset}
\label{sec:dataset}

RoadSocial is a dataset created from social media videos in unconstrained, real-world environments. These videos are accompanied by rich social commentary that reflects facts and varied cultural perspectives on road events worldwide.

\begin{figure*}[!ht]
  \centering  
   \includegraphics[width=\textwidth]{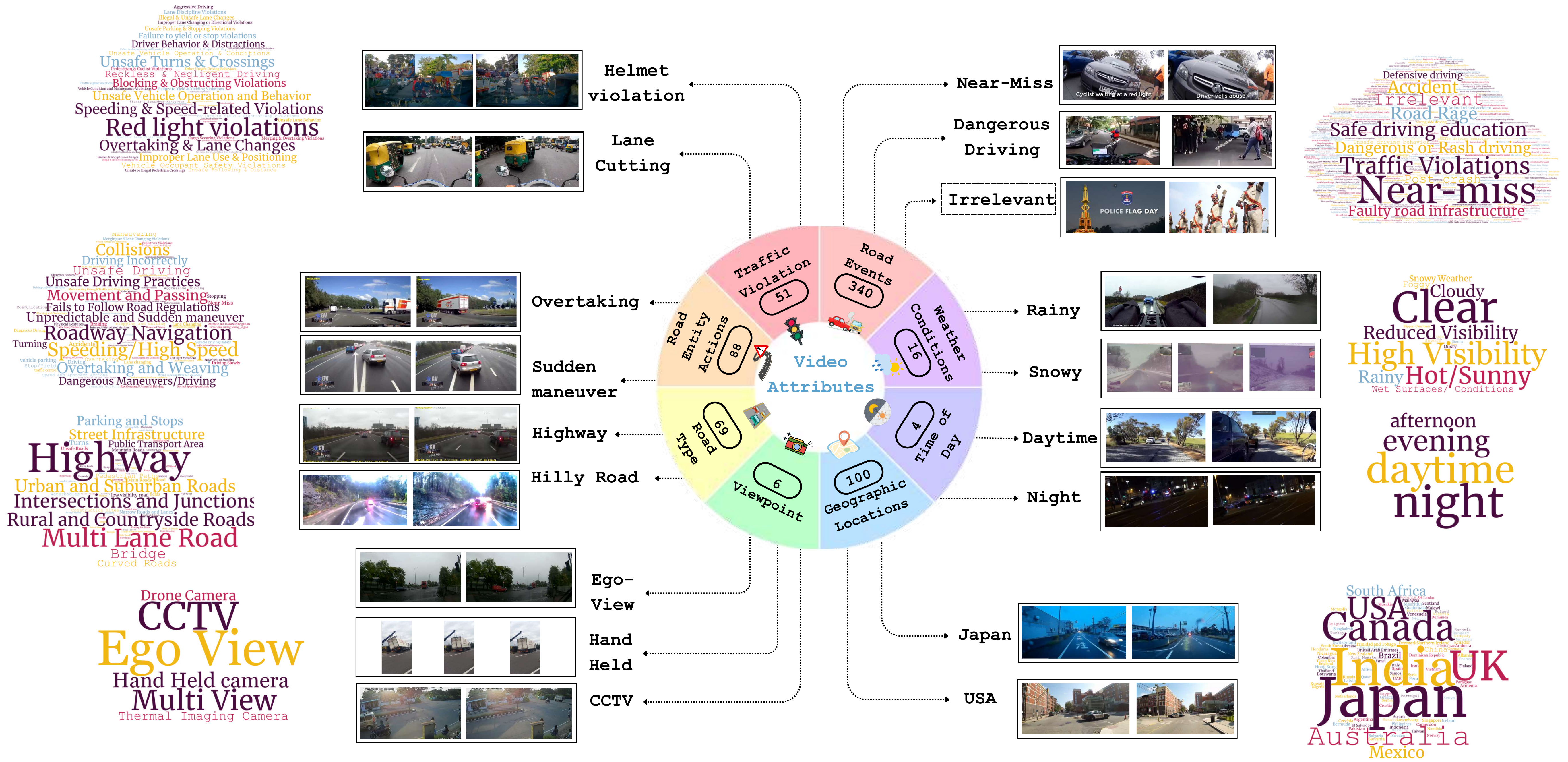}
   \caption{\textbf{Diverse Video Attributes in the RoadSocial Dataset:} The total count of unique tags for each attribute is shown in \CircledText[inner color=black, outer color=black, fill color=white]{circled boxes}, alongside word clouds highlighting these values. For each attribute, we display examples with 2-3 keyframes from videos. The figure captures the diversity of road events, environmental conditions, geographical locations, viewpoints, interactions between road entities, and traffic violations. The varied scenarios under each attribute showcase the rich complexity of our dataset.}
   \label{fig:vid_div}
\end{figure*}

\subsection{Data Collection}
\label{sec:video_collection}

We crowdsourced diverse road event data from X (formerly Twitter), leveraging its global community for real-world insights. Unlike other platforms, X is characterized by an active social discourse on road events that includes the general public, road event enthusiasts, and road enforcement authorities. Our strategy focused on popular road event related social media handles worldwide, using multilingual keywords to scrape tweet data from 2012 onwards, filtering for videos with substantial commentary. 
The resulting dataset captures varied road events—traffic violations, accidents, safe driving, and infrastructure awareness—across different environments and locations.

\begin{figure*}[t]
    \centering
    \includegraphics[width=\textwidth]{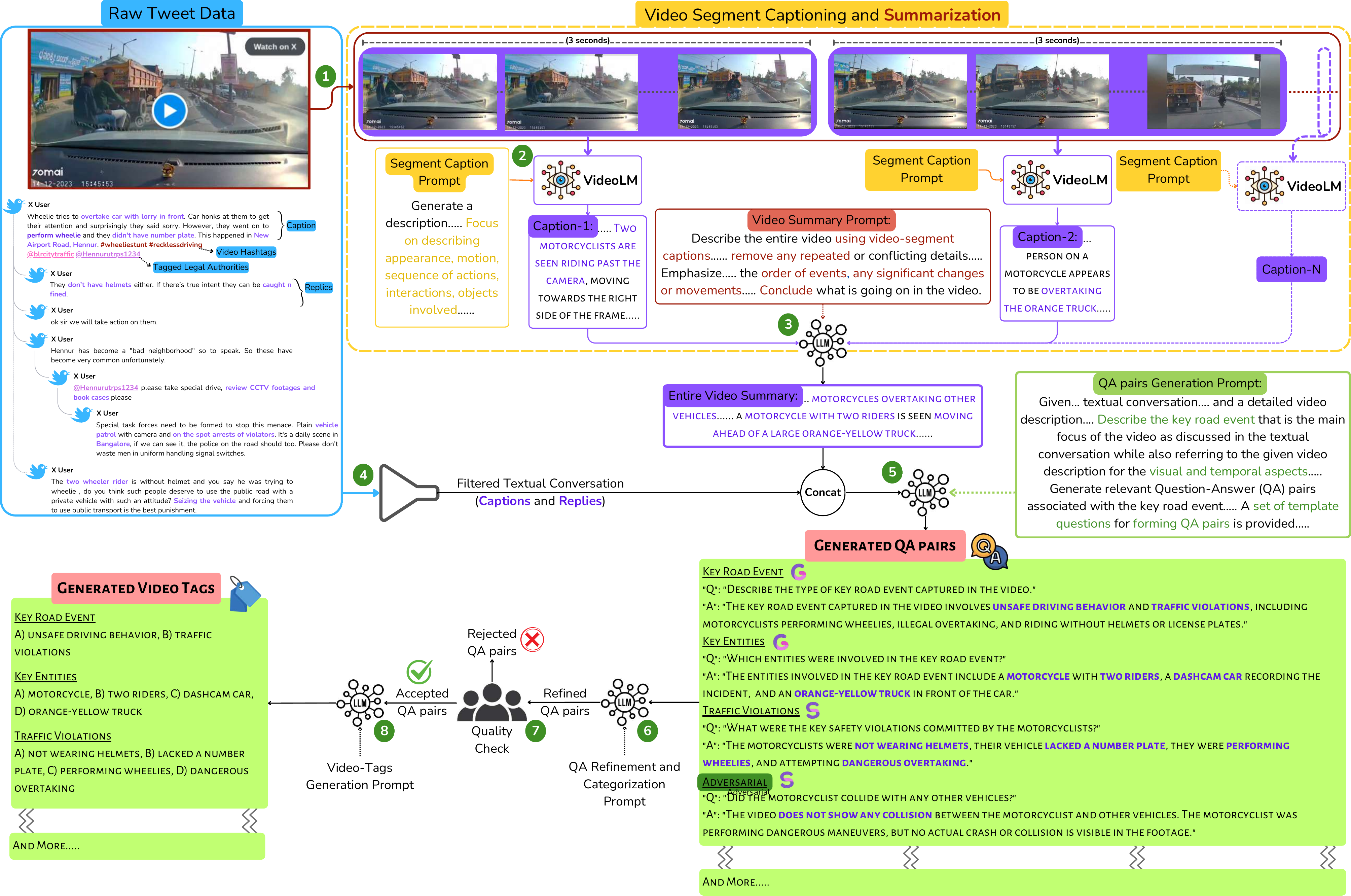}
    \caption{\textbf{RoadSocial Annotation Pipeline:} The steps involved in the annotation pipeline are depicted from \Circled{1} to \Circled{8}. \hlcyan{Raw Tweet Data} consists of the video and the Twitter conversation. Step \Circled{1} includes splitting the video into 3-second segments (in purple shaded boxes). Step \Circled{2} involves feeding the video segments to Video LLM and prompting it to generate corresponding captions numbered from 1 to N. These captions are aggregated and summarized by an LLM to generate \textcolor{darkred}{entire video summary} in Step \Circled{3}. Step \Circled{4} filters the raw tweet textual data and extracts the captions, replies, hashtags, and tagged legal authorities' user handles (highlighted in \hlcyan{blue}). This filtered conversation data and the entire video visual summary are fed to LLM and prompted to generate generic ({\includegraphics[width=0.02\textwidth]{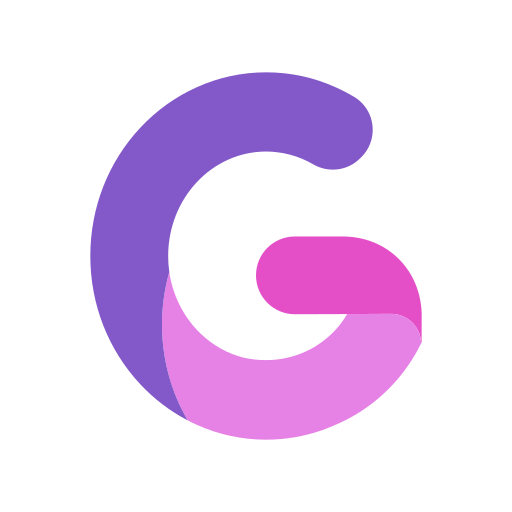}}) and specific ({\includegraphics[width=0.02\textwidth]{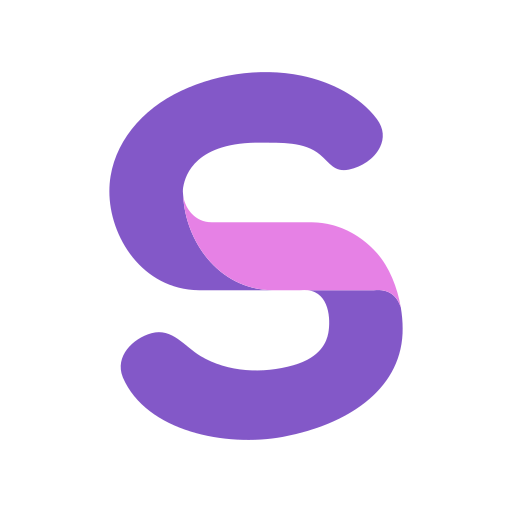}}) QA pairs in Step \Circled{5}. All important aspects of the key road event mentioned in the raw tweet text, video segment captions, the entire video summary, and the generated QA pairs are highlighted in \textcolor{darkviolet}{\textbf{purple}}. The \hlpink{generated QA pairs} are refined and categorized into pre-defined tasks in step \Circled{6}. These QA pairs are verified by expert annotators to either include or exclude them from the dataset in Step \Circled{7}. The human-verified QA pairs are then used as input to \hlpink{generate video-level tags} in Step \Circled{8}.}
    \label{fig:approach_diagram}
\end{figure*}

\subsection{Annotation Strategy: QAs and Tags}
\label{sec:qa_annotation_strategy}

Our annotation strategy merges LLM-based automation with expert verification to produce high-quality QA pairs and video tags. We start by identifying representative road event samples, then use a hybrid approach to generate QA pairs that blend video semantics with social media context. QA pairs are refined, categorized into predefined tasks, and supplemented with video-level tags, all verified by experts. Additionally, we create incompatible QA pairs for non-road event videos. Details of these steps follow.

\textbf{Identifying Representative Road Event Samples:} To design effective template questions for QA generation, we identified representative samples of diverse road events by embedding multilingual tweet text and hashtags using OpenAI's GPT-3 text embeddings~\cite{openai_api}. Hierarchical k-means clustering of these embeddings produced clusters of distinct road events (\eg UK cyclist near-misses, illegal truck overtaking in China, car hydroplaning in USA). We selected the top five samples closest to each cluster center as representatives, ensuring our QA generation is grounded in well-represented events.

\textbf{Hybrid Approach for QA Generation:} Twitter conversations often focus on unique events in the video but lack visual details (\eg, color of road entity, time of day, type of road). To create holistic QA pairs, we use a hybrid approach combining visual and contextual information from both video and conversation. Visual semantics are extracted by splitting videos into 3-second segments, prompting a Video LLM~\cite{wang2024qwen2} to generate segment captions (see \cref{fig:approach_diagram} - \greenCircledNumber{1}, \greenCircledNumber{2}). These captions are merged and passed to a Text-based Large Language Model (Text LLM)~\cite{claude_3_5} for a visually-rich summary (\greenCircledNumber{3}). Meanwhile, tweet conversations are cleaned of URLs and irrelevant data (\greenCircledNumber{4}). The Text LLM then integrates the enriched visual summary with the tweet conversation to generate QA pairs using template questions (\greenCircledNumber{5}).

To ensure a range of difficulty in QA pairs, we curate both generic and specific template questions for predefined QA tasks. Generic questions, such as \textit{What actions were performed by the road entities involved in the key road event?}, require complex reasoning and are harder to answer, while specific questions, such as \textit{How was the truck involved in the accident?}, directly reference the event and entities, making them easier. This approach ensures varied difficulty in QA pairs. Sample QA pairs generated by our hybrid approach are illustrated in \cref{fig:approach_diagram}. The prompts used in each QA stage are iteratively refined on representative samples for quality. \cref{sec:template_q_generation,sec:hybrid_approach,sec:specific_qa_generation} includes details for all template questions, prompts, and outcomes.


\textbf{QA Refinement and Categorization:} Social conversations often include non-visual information, like names or past experience. Relying on such data for QA generation may produce answers with irrelevant information. To address this, we prompt a Text LLM to refine QA pairs by removing specific details—such as names, past encounters, or dates—that aren’t directly inferable from the video. This refinement step (\greenCircledNumber{6}) ensures the QA pairs are answerable solely through video content, enhancing quality. To evaluate various aspects of road event understanding, we categorize QA pairs into predefined tasks (\greenCircledNumber{6}). A Text LLM assigns each QA pair to a task with a category score, and expert annotators review pairs with low scores for accuracy, reassigning or removing as needed. Verified QA pairs then undergo final quality checks for relevance to the video (\greenCircledNumber{7}). Detailed prompts and sample outputs are provided in \cref{sec:refinement_and_categorization}. 

\textbf{Video-level Tag Generation:} To categorize videos by key aspects of road events, we generate diverse video-level tags (\eg traffic violation, wheelie, unsafe overtaking) using verified answers from step \greenCircledNumber{7}. A Text LLM~\cite{claude_3_5} scans these answers to generate top-k tags most relevant to each QA task (\greenCircledNumber{8}). This structured tagging approach ensures that the generated QA pairs, tags are robust and reflect the diverse scenarios present in the dataset. Details about the tag generation prompt and the resulting tags distribution are provided in \cref{sec:video_tag_gen}.

\begin{figure*}[t]
    \centering
    \includegraphics[width=\textwidth]{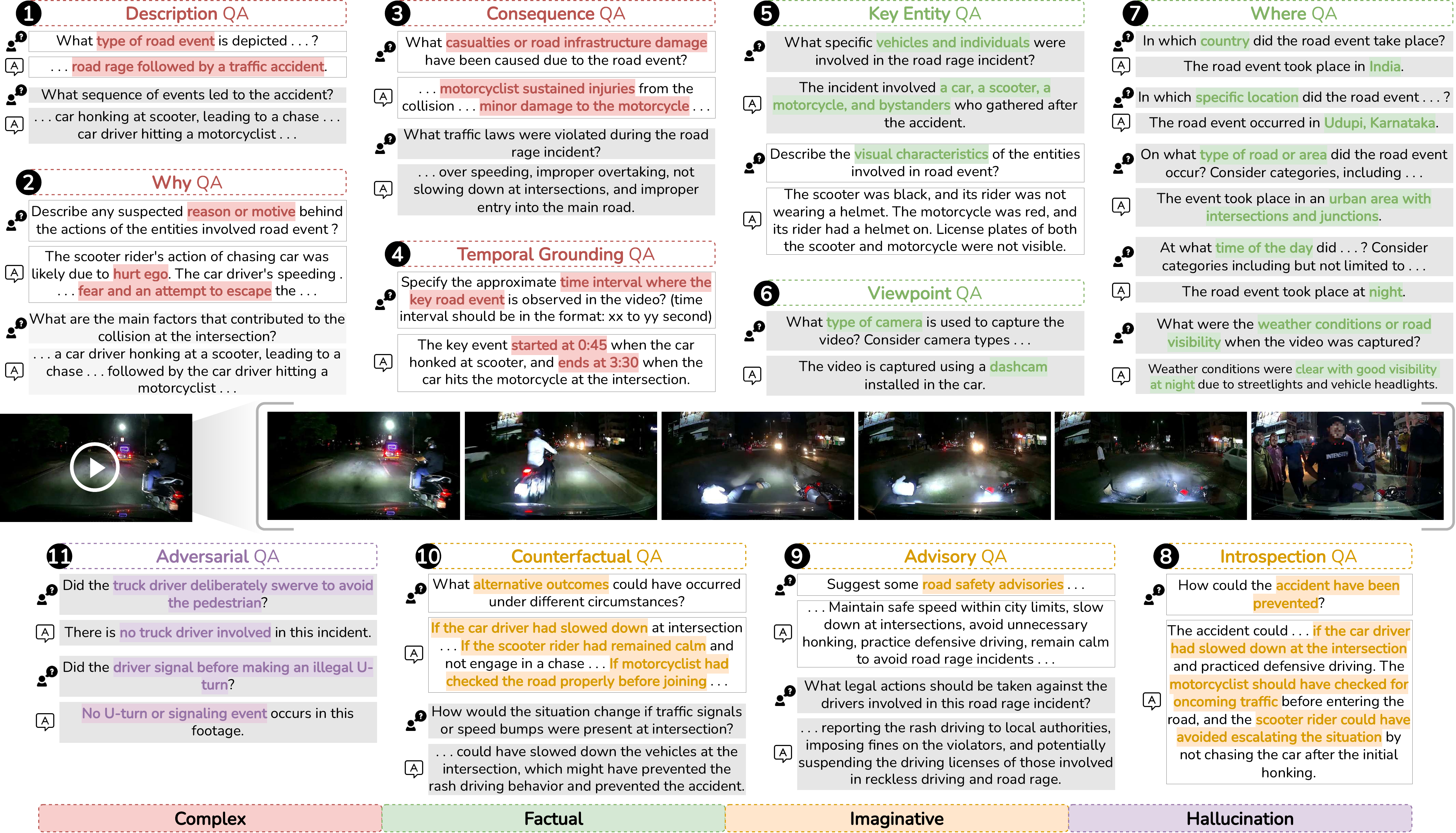}
    \caption{Examples of QA Pairs grouped by tasks and color-coded by task category. Gray outlined questions are generic while gray fill shading indicates specific questions. Highlighted text indicates key information. (\cref{sec:qa_annotation}).}
    \label{fig:task_qa_examples}
\end{figure*}

\begin{figure}[!t]
  \centering  
   \includegraphics[width=1.0\linewidth]{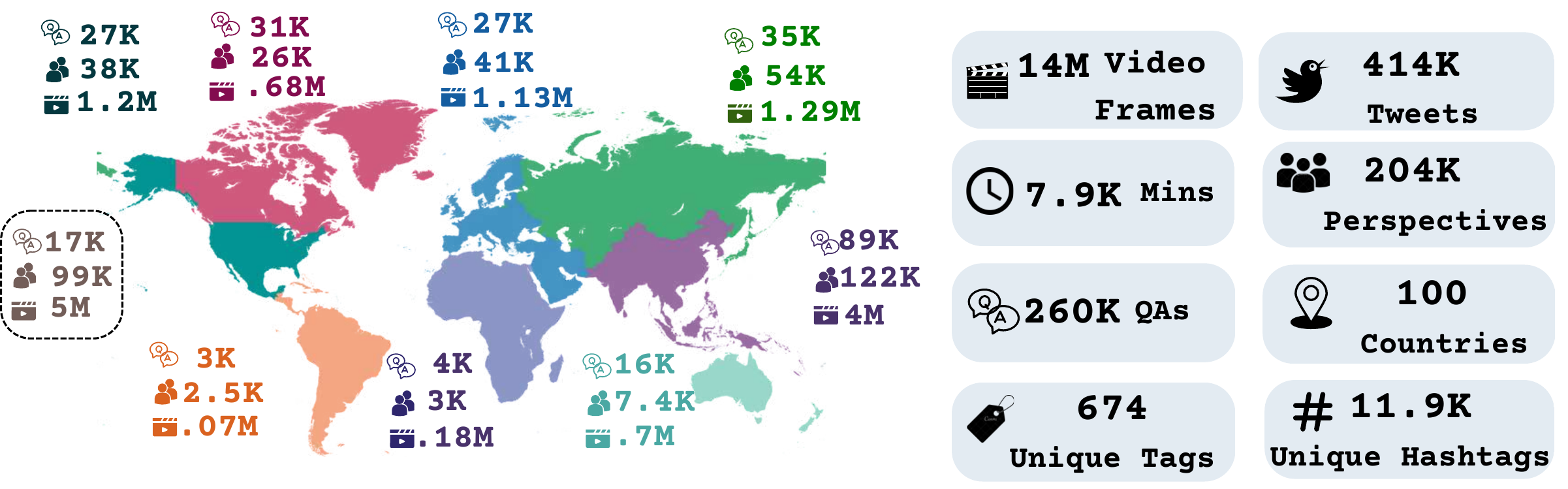}
   \caption{\textbf{The diversity of RoadSocial dataset:} The number of QA pairs, social commentary (tweets), and video frames spread across different regions is shown. Overall statistics of the raw tweet data, generated QA pairs, and tags in our dataset is also shown. Total incompatible QA pairs and related numbers for non-road event videos are specified inside a \textcolor{lightbrown}{light brown box} at left.}
   \label{fig:MapInfo}
\end{figure}

\textbf{Incompatible QA Generation:} To assess the reliability and resistance of Video LLMs to hallucinations, we generate incompatible QA pairs for non-road event videos. This involves sampling questions from road event QA pairs to create mismatched questions for unrelated videos. Answers are generated using the Hybrid Approach mentioned previously, with modified prompts treating these mismatched questions as templates. Further details on the prompt modifications are provided in \cref{sec:incompatible_qa_gen}.

\subsection{Dataset Statistics}
\label{sec:dataset_stats}

Our final dataset comprises over \textbf{14M} video frames from more than \textbf{13.2K} videos (totaling \textbf{7.9K} minutes of video footage) with \textbf{260K} QA pairs and \textbf{674} unique video tags (total \textbf{100K+}). 

The dataset exhibits significant diversity across several dimensions, including geographical distribution (\cref{fig:MapInfo}), QA types (\cref{fig:task_qa_examples}), and video tags (\cref{fig:vid_div}). \cref{fig:MapInfo} shows the global coverage of our dataset attributes, depicting the diverse perspectives involved in the QA pair generation process.

It includes \textbf{414K} multilingual tweet captions and replies corresponding to \textbf{204K} unique user handles (from across \textbf{100} countries) sharing facts and opinions about the road or traffic events. \cref{tab:dataset-comparison} compares key attributes of our dataset with related road event understanding datasets. The distribution of QA pairs corresponding to each task category is shown in \cref{fig:qa_tasks}. The distribution of video tags along different attributes is shown by word clouds in \cref{fig:vid_div}. The videos durations range from 0.13 seconds to 3885.44 seconds with an average of 35.6 seconds.

\subsection{QA Tasks Taxonomy}
\label{sec:qa_annotation}

\begin{figure}[!t]
    \centering
    \includegraphics[width=0.4\textwidth]{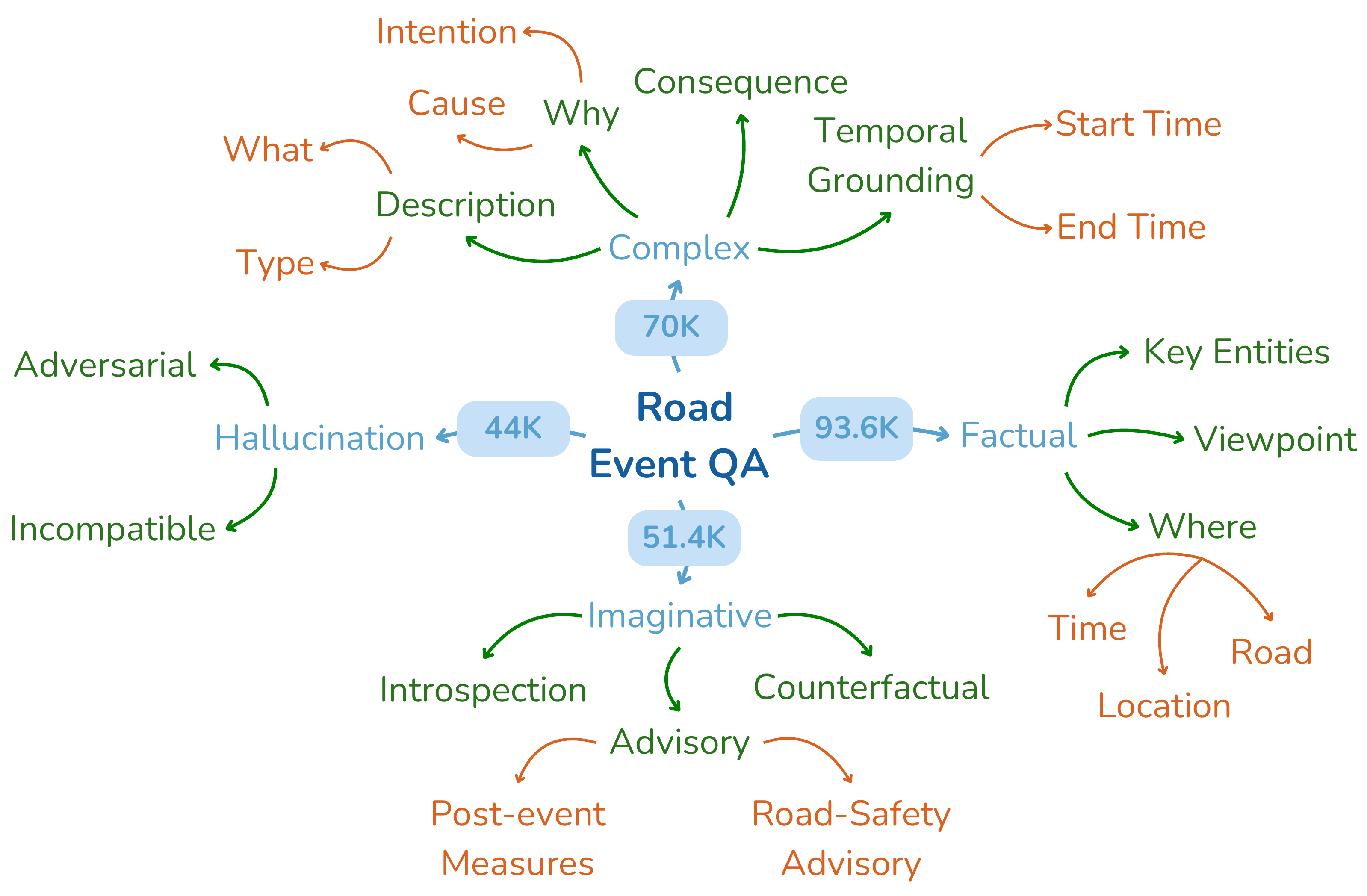}
    \caption{\textbf{QA Task Taxonomy:} The QA pairs in RoadSocial are broadly grouped into 4 categories (highlighted in blue) which are further subdivided into 12 tasks (shown in green).  Total QA pair count for each category is shown in blue squared box. Some of these tasks are further subdivided into granular sub-tasks (highlighted in orange) to facilitate coarse to fine-grained understanding of road events along different aspects.}
    \label{fig:qa_tasks}
\end{figure}

We developed a question-answer (QA) taxonomy for structured evaluation of Video Large Language Models (Video LLMs). The taxonomy consists of 12 distinct tasks organized into four reasoning categories: Complex, Factual, Imaginative, and Hallucination (\cref{fig:qa_tasks}). These categories assess various aspects of road events, ranging from key entity identification (see \blackCircledNumber{5} in \cref{fig:task_qa_examples}) to hypothetical scenario exploration (\cref{fig:task_qa_examples} \blackCircledNumber{10}). Our taxonomy extends beyond conventional road datasets by incorporating previously underrepresented tasks, such as Viewpoint QA (analysis of camera perspectives capturing road events) and Where QA (geographic location identification of road events). As an additional novelty, our approach uniquely incorporates Adversarial QA and Incompatible QA. Adversarial QA tests a model’s ability to recognize and reject misleading assumptions or false details in questions by identifying non-occurring road events \eg \cref{fig:task_qa_examples} \blackCircledNumber{11}). Incompatible QA on non-road-event videos helps evaluate models' robustness to hallucination by identifying irrelevant video-question pairs. \cref{fig:task_qa_examples} illustrates representative QA pairs for each category including the generic and specific questions (described in \cref{sec:qa_annotation_strategy}).  
For a detailed QA task description, please refer to \cref{sec:qa_taxonomy}.

\section{Experiments}
\label{sec:experiments}

We evaluate a wide range of Video LLMs (both open-source and proprietary, driving-specific and general-purpose) on our road event understanding benchmark. 

\subsection{Data Setup}
\label{sec:experimental_setup_data}

\textbf{Evaluation Benchmark:} RoadSocial-QA consists of \textbf{13.2K} videos encompassing \textbf{260K} QA pairs, with an average of \textbf{20} QA pairs per video. To evaluate zero-shot reasoning capabilities of Video LLMs, we split our dataset into \textbf{12K} training and \textbf{1.2K} test videos, resulting in \textbf{234K} and \textbf{26K} QA pairs respectively. The video splits maintain geographical diversity across the dataset, with the test set serving as our primary evaluation benchmark.

\begin{figure}[!t]
  \centering  
   \includegraphics[width=\linewidth]{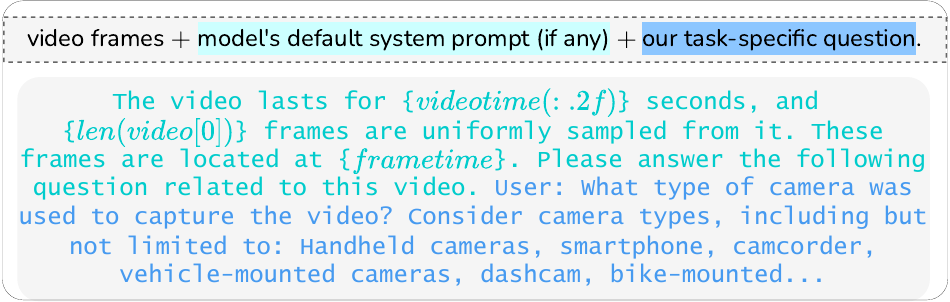}
   \caption{An example of prompting a Video LLM~\cite{zhang2024videoinstructiontuningsynthetic}.}
   \label{fig:std_prompt}
\end{figure}

For model evaluation, we provide the model with video frames and a task-specific question, following the format: video frames $+$ model's default system prompt (if any) $+$ our task-specific question (\cref{fig:std_prompt}). Detailed prompting structures are described in \cref{sec:data_setup}.

\subsection{Model Setup}
\label{sec:model_setup}

Our evaluation encompasses 18 Video LLMs, comprising 15 open-source general-purpose models, 2 proprietary general-purpose models, and 1 open-source driving-specific model. We evaluate their zero-shot performance on the test split of RoadSocial-QA using each model's official configuration for open-ended response generation. The results, presented in \cref{tab:eval_models_all}, analyze model performance across different tasks. All evaluation runs were conducted on a computing cluster equipped with NVIDIA H100 GPUs. Detailed information about model configurations, prompting templates and evaluation timelines is provided in \cref{sec:model_setup_supp}. 

\begin{table*}[!ht]
\footnotesize
\begin{center}
\setlength{\tabcolsep}{3pt}
\newcommand{\applyHeatmap}[1]{%
    \ifdim #1 pt < 20pt \cellcolor{cyan!12}#1\else%
    \ifdim #1 pt < 40pt \cellcolor{cyan!30}#1\else%
    \ifdim #1 pt < 60pt \cellcolor{yellow!35}#1\else%
    \ifdim #1 pt < 80pt \cellcolor{orange!45}#1\else%
    \cellcolor{red!55}#1%
    \fi\fi\fi\fi
}

\begin{tabular}{l|c|cccccccccccc|cccc}
\toprule
\multirow{2}{*}{Model} & \multirow{2}{*}{Params} & \multicolumn{3}{c}{Factual} & \multicolumn{4}{c}{Complex} & \multicolumn{3}{c}{Imaginative} & \multicolumn{2}{c|}{Hallucination} & Overall & Overall & Overall & Overall \\
 \cmidrule(rl){3-5} \cmidrule(rl){6-9} \cmidrule(rl){10-12} \cmidrule(rl){13-14}
& & WR & KE & VP & DS & WY & CQ & TG & AD & IN & CF & AV & IC & (ALL) & (RT) & (Generic) & (Specific) \\[1.2pt]
\cmidrule(r){1-1} \cmidrule(r){2-2} \cmidrule(rl){3-5} \cmidrule(rl){6-9} \cmidrule(rl){10-12} \cmidrule(rl){13-14} \cmidrule(rl){15-15} \cmidrule(rl){16-16} \cmidrule(rl){17-17} \cmidrule(l){18-18}
\cellcolor{PasteGreen} Dolphin~\cite{ma2023dolphins} & 9B & \applyHeatmap{61.3} & \applyHeatmap{34.5} & \applyHeatmap{67.8} & \applyHeatmap{35.8} & \applyHeatmap{25.2} & \applyHeatmap{37.2} & \applyHeatmap{0.01} & \applyHeatmap{49.8} & \applyHeatmap{39.1} & \applyHeatmap{45.5} & \applyHeatmap{71.8} & \applyHeatmap{21.3} & \applyHeatmap{40.8} & \applyHeatmap{42.5} & \applyHeatmap{29.8} & \applyHeatmap{46.5}  \\
\midrule
\cellcolor{PasteLavender} GPT-4o~\cite{openai2024gpt4o} & - & \applyHeatmap{77.0} & \textbf{\applyHeatmap{66.6}} & \applyHeatmap{84.3} & \textbf{\applyHeatmap{70.2}} & \textbf{\applyHeatmap{70.8}} & \textbf{\applyHeatmap{72.1}} & \applyHeatmap{7.8} & \textbf{\applyHeatmap{77.7}} & \textbf{\applyHeatmap{76.4}} & \textbf{\applyHeatmap{77.0}} & \textbf{\applyHeatmap{90.0}} & \textbf{\applyHeatmap{67.6}} & \textbf{\applyHeatmap{69.8}} & \textbf{\applyHeatmap{70.0}} & \textbf{\applyHeatmap{69.5}} & \textbf{\applyHeatmap{74.4}} \\ 
\cellcolor{PasteLavender} Gemini-1.5-Pro~\cite{team2023gemini} & - & \textbf{\applyHeatmap{77.7}} & \applyHeatmap{56.7} & \textbf{\applyHeatmap{85.4}} & \applyHeatmap{61.9} & \applyHeatmap{61.4} & \applyHeatmap{60.1} & \textbf{\applyHeatmap{18.6}} & \applyHeatmap{72.1} & \applyHeatmap{70.2} & \applyHeatmap{75.7} & \applyHeatmap{72.3} & \applyHeatmap{48.7} & \applyHeatmap{63.4} & \applyHeatmap{64.7} & \applyHeatmap{60.1} & \applyHeatmap{68.3} \\ 
\midrule
\cellcolor{PastePink} InternVL2~\cite{Chen2023InternVS} & 76B & \applyHeatmap{72.4} & \applyHeatmap{51.3} & \applyHeatmap{81.4} & \applyHeatmap{57.1} & \applyHeatmap{59.0} & \applyHeatmap{62.1} & \applyHeatmap{1.07} & \applyHeatmap{70.5} & \applyHeatmap{67.0} & \applyHeatmap{69.2} & \applyHeatmap{58.6} & \applyHeatmap{27.6} &  \applyHeatmap{56.4} & \applyHeatmap{59.1} & \applyHeatmap{55.5} & \applyHeatmap{65.1}  \\ 
\cellcolor{PastePink} Qwen2-VL~\cite{wang2024qwen2} & 72B & \applyHeatmap{76.8} & \applyHeatmap{56.6} & \applyHeatmap{85.1} & \applyHeatmap{60.2} & \applyHeatmap{64.0} & \applyHeatmap{67.6} & \applyHeatmap{0.01} & \applyHeatmap{71.9} & \applyHeatmap{72.4} & \applyHeatmap{71.6} & \applyHeatmap{37.0} & \applyHeatmap{40.2} &  \applyHeatmap{58.6} & \applyHeatmap{60.3} & \applyHeatmap{58.3} & \applyHeatmap{68.8}  \\ 
\cellcolor{PastePink} LLaVA-Video~\cite{zhang2024videoinstructiontuningsynthetic} & 72B & \applyHeatmap{75.8} & \applyHeatmap{52.4} & \applyHeatmap{76.8} & \applyHeatmap{52.4} & \applyHeatmap{55.0} & \applyHeatmap{52.2} & \textbf{\applyHeatmap{9.94}} & \applyHeatmap{68.3} & \applyHeatmap{63.7} & \applyHeatmap{64.9} & \applyHeatmap{83.5} & \applyHeatmap{24.7} &  \applyHeatmap{56.7} & \applyHeatmap{59.6} & \applyHeatmap{51.1} & \applyHeatmap{63.3}  \\ 
\cellcolor{PastePink} LLaVA-OV~\cite{li2024llava} & 72B & \applyHeatmap{75.1} & \applyHeatmap{54.1} & \applyHeatmap{78.7} & \applyHeatmap{53.0} & \applyHeatmap{53.3} & \applyHeatmap{54.1} & \applyHeatmap{3.99} & \applyHeatmap{67.8} & \applyHeatmap{61.9} & \applyHeatmap{63.1} & \applyHeatmap{45.1} & \applyHeatmap{19.9} &  \applyHeatmap{52.5} & \applyHeatmap{55.5} & \applyHeatmap{51.8} & \applyHeatmap{63.0}  \\ 
\cellcolor{PastePink} VITA~\cite{fu2024vita} & 8x7B & \applyHeatmap{66.6} & \applyHeatmap{52.1} & \applyHeatmap{71.6} & \applyHeatmap{48.1} & \applyHeatmap{55.6} & \applyHeatmap{56.3} & \applyHeatmap{2.27} & \applyHeatmap{66.7} & \applyHeatmap{66.0} & \applyHeatmap{62.4} & \applyHeatmap{56.3} & \applyHeatmap{22.0} & \applyHeatmap{52.2} & \applyHeatmap{54.9} & \applyHeatmap{49.8} & \applyHeatmap{60.4}  \\ 
\cellcolor{PastePink} Tarsier~\cite{Wang2024TarsierRF} & 34B & \applyHeatmap{73.7} & \applyHeatmap{58.1} & \applyHeatmap{78.2} & \applyHeatmap{58.2} & \applyHeatmap{59.0} & \applyHeatmap{58.8} & \applyHeatmap{0.32} & \applyHeatmap{71.6} & \applyHeatmap{71.1} & \applyHeatmap{67.4} & \applyHeatmap{83.2} & \textbf{\applyHeatmap{82.3}} & \textbf{\applyHeatmap{63.5}} & \applyHeatmap{61.8} & \applyHeatmap{58.4} & \applyHeatmap{66.1}  \\ 

\cellcolor{PastePink} ARIA~\cite{li2024aria} & 25.3B & \applyHeatmap{75.4} & \applyHeatmap{53.1} & \textbf{\applyHeatmap{86.2}} & \applyHeatmap{58.4} & \applyHeatmap{56.9} & \textbf{\applyHeatmap{70.2}} & \applyHeatmap{8.96} & \textbf{\applyHeatmap{75.1}} & \applyHeatmap{74.7} & \applyHeatmap{74.0} & \textbf{\applyHeatmap{86.4}} & \applyHeatmap{29.2} & \applyHeatmap{62.4} & \applyHeatmap{65.4} & \applyHeatmap{56.7} & \applyHeatmap{68.5} \\
\cellcolor{PastePink} InternVL2~\cite{Chen2023InternVS} & 8B & \applyHeatmap{67.7} & \applyHeatmap{51.7} & \applyHeatmap{78.0} & \applyHeatmap{55.7} & \applyHeatmap{59.3} & \applyHeatmap{60.9} & \applyHeatmap{0.77} & \applyHeatmap{66.7} & \applyHeatmap{66.8} & \applyHeatmap{70.0} & \applyHeatmap{68.1} & \applyHeatmap{26.1} & \applyHeatmap{56.0} & \applyHeatmap{58.7} & \applyHeatmap{53.7} & \applyHeatmap{64.0}  \\ 
\cellcolor{PastePink} Mini-CPM-V 2.6~\cite{yao2024minicpm} & 8B & \applyHeatmap{77.7} & \applyHeatmap{57.6} & \applyHeatmap{80.6} & \applyHeatmap{55.0} & \applyHeatmap{50.5} & \applyHeatmap{57.5} & \applyHeatmap{0.4} & \applyHeatmap{61.6} & \applyHeatmap{52.3} & \applyHeatmap{59.3} & \applyHeatmap{73.5} & \applyHeatmap{30.0} & \applyHeatmap{54.7} & \applyHeatmap{56.9} & \applyHeatmap{51.0} & \applyHeatmap{62.0}  \\ 
\cellcolor{PastePink} IXC-2.5~\cite{Zhang2024InternLMXComposer25AV} & 7B & \textbf{\applyHeatmap{78.5}} & \textbf{\applyHeatmap{58.7}} & \applyHeatmap{85.4} & \textbf{\applyHeatmap{61.7}} & \textbf{\applyHeatmap{65.3}} & \applyHeatmap{68.5} & \applyHeatmap{0.69} & \applyHeatmap{73.9} & \textbf{\applyHeatmap{75.6}} & \textbf{\applyHeatmap{75.7}} & \applyHeatmap{85.8} & \applyHeatmap{29.2} & \applyHeatmap{63.3} & \textbf{\applyHeatmap{66.4}} & \textbf{\applyHeatmap{60.7}} & \textbf{\applyHeatmap{70.3}}  \\ 
\cellcolor{PastePink} Tarsier~\cite{Wang2024TarsierRF} & 7B & \applyHeatmap{69.9} & \applyHeatmap{54.7} & \applyHeatmap{72.3} & \applyHeatmap{52.0} & \applyHeatmap{53.4} & \applyHeatmap{55.2} & \applyHeatmap{0.11} & \applyHeatmap{69.5} & \applyHeatmap{69.3} & \applyHeatmap{63.5} & \applyHeatmap{79.1} & \applyHeatmap{67.3} & \applyHeatmap{58.9} & \applyHeatmap{58.1} & \applyHeatmap{54.0} & \applyHeatmap{61.7}  \\ 
\cellcolor{PastePink} LongVU~\cite{shen2024longvu} & 7B & \applyHeatmap{73.0} & \applyHeatmap{53.0} & \applyHeatmap{76.3} & \applyHeatmap{51.1} & \applyHeatmap{50.2} & \applyHeatmap{55.0} & \applyHeatmap{0.84} & \applyHeatmap{59.7} & \applyHeatmap{55.8} & \applyHeatmap{58.2} & \applyHeatmap{48.9} & \applyHeatmap{32.7} & \applyHeatmap{51.2} & \applyHeatmap{52.9} & \applyHeatmap{47.7} & \applyHeatmap{59.7} \\
\cellcolor{PastePink} Qwen2-VL~\cite{wang2024qwen2} & 7B & \applyHeatmap{75.5} & \applyHeatmap{52.8} & \applyHeatmap{76.1} & \applyHeatmap{52.7} & \applyHeatmap{57.7} & \applyHeatmap{56.4} & \applyHeatmap{0.59} & \applyHeatmap{69.2} & \applyHeatmap{71.6} & \applyHeatmap{65.9} & \applyHeatmap{37.5} & \applyHeatmap{39.6} & \applyHeatmap{54.6} & \applyHeatmap{56.0} & \applyHeatmap{52.6} & \applyHeatmap{63.9}  \\ 
\cellcolor{PastePink} LLaVA-Video~\cite{zhang2024videoinstructiontuningsynthetic} & 7B & \applyHeatmap{74.6} & \applyHeatmap{50.1} & \applyHeatmap{76.7} & \applyHeatmap{52.1} & \applyHeatmap{50.1} & \applyHeatmap{50.3} & \applyHeatmap{1.43} & \applyHeatmap{60.4} & \applyHeatmap{53.8} & \applyHeatmap{58.7} & \applyHeatmap{61.8} & \applyHeatmap{23.5} & \applyHeatmap{51.1} & \applyHeatmap{53.6} & \applyHeatmap{47.6} & \applyHeatmap{59.7}  \\ 
\cellcolor{PastePink} LLaVA-OV~\cite{li2024llava} & 7B & \applyHeatmap{73.4} & \applyHeatmap{51.2} & \applyHeatmap{77.2} & \applyHeatmap{50.7} & \applyHeatmap{51.7} & \applyHeatmap{51.2} & \applyHeatmap{0.97} & \applyHeatmap{62.8} & \applyHeatmap{55.4} & \applyHeatmap{58.6} & \applyHeatmap{45.4} & \applyHeatmap{21.1} & \applyHeatmap{50.0} & \applyHeatmap{52.6} & \applyHeatmap{48.4} & \applyHeatmap{59.8}  \\ 
\midrule \midrule
\cellcolor{Gray} LLaVA-OV ft. & 7B & \applyHeatmap{80.9} & \applyHeatmap{64.1} & \applyHeatmap{85.7} & \applyHeatmap{64.1} & \applyHeatmap{68.7} & \applyHeatmap{65.1} & \applyHeatmap{4.49} & \applyHeatmap{74.2} & \applyHeatmap{70.9} & \applyHeatmap{71.7} & \applyHeatmap{95.4} & \applyHeatmap{87.6} & \applyHeatmap{69.4} & \applyHeatmap{67.8} & \applyHeatmap{65.1} & \applyHeatmap{69.7} \\ 
\bottomrule
\end{tabular}
\end{center}
\caption{\textbf{Video LLMs benchmarked on RoadSocial-QA.} Standard prompting with task-specific instructions were employed for zero-shot evaluation of Video LLMs on 12 QA tasks. Video LLMs are grouped as open-source (\hlpastegreen{driving-specific} and \hlpastepink{general-purpose}), and \hlpastelavender{closed-source} models. Further, we \hlGray{fine-tune} a Video LLM - LLaVA-OV-7B and report its performance at the end of the table. Abbreviations used for QA tasks include Factual (F), Complex (C), Imaginative (I), Hallucination (H), Where (WR), Key Entities (KE), Viewpoint (VP), Description (DS), Why (WY), Consequence (CQ), Temporal Grounding (TG), Advisory (AD), Introspection (IN), Counterfactual (CF), Adversarial (AV), Incompatible (IC), and Road-event related Tasks (RT). RT includes all tasks except IC which corresponds to non-road event videos. GPT-3.5 score is reported for all tasks except Temporal Grounding (TG) for which average mAP@.3:.7 (\%) is reported. Overall average scores are reported for ALL QA tasks (F, C, I, and H), Road-event related Tasks (RT), Generic QAs, and Specific QAs under each task. All reported scores (scale 0 to 100) are colored based on their value from \hlcyanlow{low} to \hlredhigh{high}. VideoLLMs show per-query latencies of 1-25s (7B-76B) on H100 GPUs.}
\label{tab:eval_models_all}
\end{table*}

\subsection{Evaluation Metrics} 
\label{sec:metrics}

To assess the similarity between model-generated and ground-truth open-ended responses, we adopt GPT-3.5 score~\cite{openai2022gpt35} as our primary evaluation metric for all tasks (except Temporal Grounding), following established practices in recent literature~\cite{sima2023drivelm,Xu2023DriveGPT4IE,lin2021truthfulqa,maaz2023video}. 
To ensure statistical robustness, we conduct multiple evaluation runs and report mean of the GPT-3.5 scores. 
For Temporal Grounding QAs, time interval is extracted from the model-generated response and compared with the ground-truth time interval range using the mean Average Precision (mAP) evaluation metric.
The complete evaluation protocols, including prompt templates, and scoring criteria are provided in \cref{sec:model_setup_supp}.

\subsection{Analysis}

\begin{figure}[!t]
  \centering  
   \includegraphics[width=0.9\linewidth]{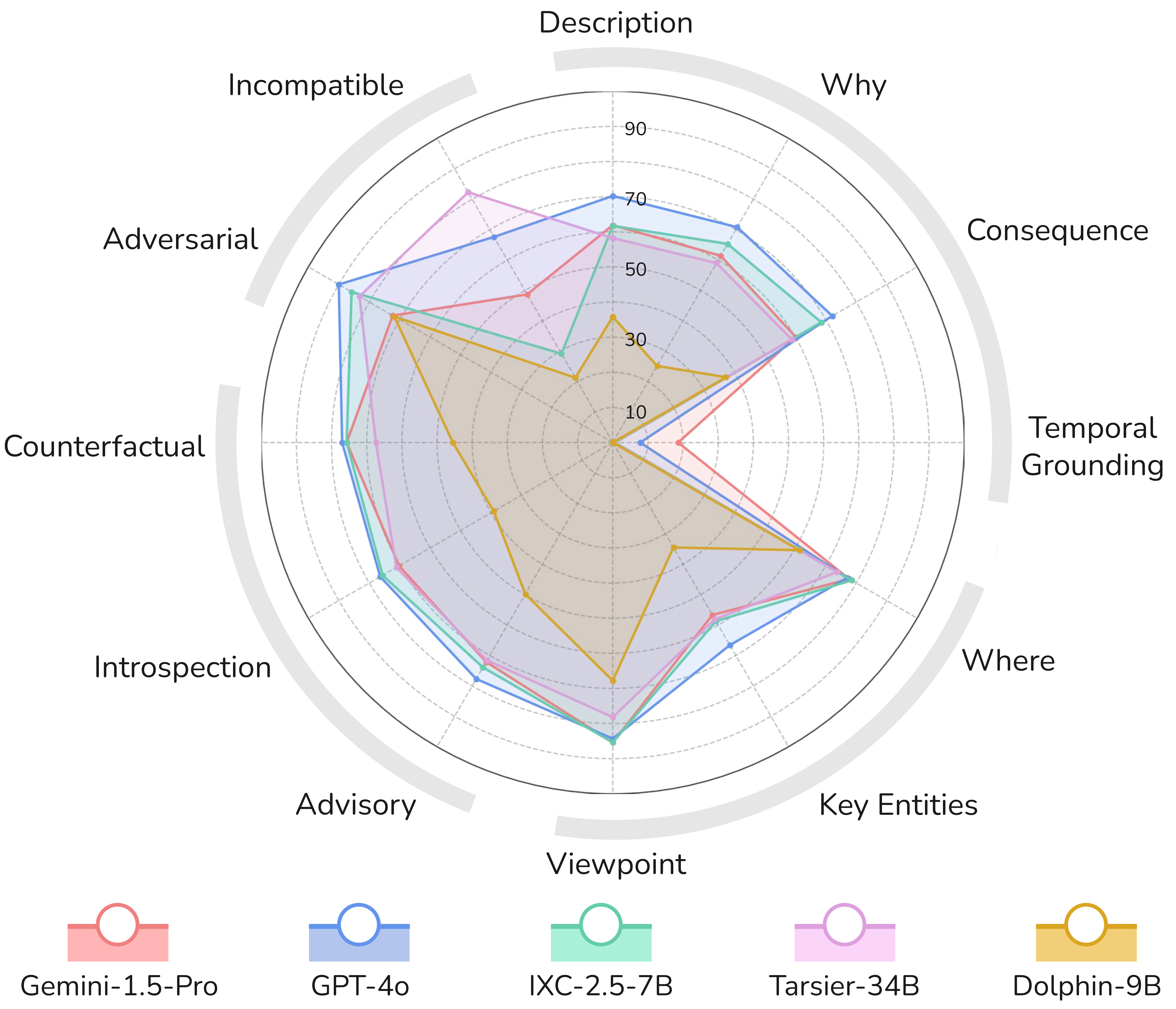}
   \caption{ Comparison between representative Video LLMs on RoadSocial benchmark across different QA tasks.}
   \label{fig:radar}
\end{figure}

\noindent \textbf{Overall Performance Trends:} Refer to last 3 columns of \cref{tab:eval_models_all}. Tarsier-34B~\cite{Wang2024TarsierRF} achieves the highest overall score (63.5) across ALL QA tasks whereas IXC-2.5-7B~\cite{Zhang2024InternLMXComposer25AV} leads the benchmark on road-event related tasks (RT) (66.4) among open-source models. These models even outperform larger models such as InternVL2-76B~\cite{Chen2023InternVS} and Qwen2-VL-72B~\cite{wang2024qwen2}. Additionally, all general-purpose models surpass driving-specific Video LLM across all tasks, revealing significant performance gaps in general road event understanding within the driving-focused model. Predictably, Video LLMs face greater difficulty with Generic QAs compared to Specific QAs because generic questions require the model to infer the context independently, unlike specific questions. Among closed-source models, GPT-4o~\cite{openai2024gpt4o} stands out as the top performer, achieving the highest scores across all models. A radar plot with representative Video LLMs can be viewed in \cref{fig:radar}. 

\noindent \textbf{Performance Across Task Categories:}
The analysis reveals distinct patterns across different reasoning categories, highlighting strengths and weaknesses among models in various types of reasoning tasks.

In factual reasoning, models perform well in Where (WR) and Viewpoint (VP) QA tasks, both of which yield consistently high scores. For VP tasks, this may be partly due to our prompt that offers a limited set of viewpoint options, essentially transforming the question into a multiple-choice format rather than a free-form open-ended question. Empirically, performance declines when these choices are absent from the prompt, as noted in \cref{sec:data_setup}. Meanwhile, WR tasks perform well due to their inherently specific questions.

Most Video LLMs encounter difficulties with complex reasoning tasks, such as Description (DS), Why (WY), Consequence (CQ), and Temporal Grounding (TG) reasoning, as well as Key Entity (KE) tasks. These results indicate that many models struggle with identifying key road event that is the main focus of the video.

Temporal Grounding (TG) proves to be particularly challenging, with most models achieving average mAP scores below 1\%, highlighting a major limitation in temporal localization for Video LLMs. The highest-performing model, Gemini-1.5-Pro~\cite{team2023gemini}, achieves 18.6\%. In comparison, LLaVA-Video-72B~\cite{zhang2024videoinstructiontuningsynthetic} leads among open-source models with 9.94\%, potentially benefiting from its default prompt, which incorporates time-based instructions (\cref{fig:std_prompt}). Empirical analysis shows two common reasons for TG underperformance: some models, such as Tarsier-34B~\cite{Wang2024TarsierRF}, struggle with instruction following, leading to unexpected or incoherent answers, while others, such as LLaVA-OV~\cite{li2024llava}, lack the capability to associate the sequence of events with time in the video (details in \cref{sec:qualitative}).

In imaginative reasoning, models show promising capabilities, with several models achieving over 70\% accuracy in Advisory (AD) and Introspection (IN) tasks. This indicates that models can effectively use their pre-trained knowledge to reason about hypothetical scenarios.

\noindent \textbf{Robustness and Hallucination Assessment:}
The evaluation of model robustness through Adversarial (AV) and Incompatible (IC) QAs reveals interesting behavioral patterns. Some models, such as GPT-4o~\cite{openai2024gpt4o} and IXC-2.5-7B~\cite{Zhang2024InternLMXComposer25AV} demonstrate exceptional robustness to adversarial queries, suggesting effective mechanisms for identifying misleading information. However, most models struggle on Incompatible QAs indicating their tendency to generate hallucinated responses for irrelevant Video and QA pairs. Notably, Tarsier-34B~\cite{Wang2024TarsierRF} outperforms all models by a good margin indicating inherent capability to identify misleading information and reject out-of-domain queries.

\noindent \textbf{Error Analysis and Future Directions:} (1) \textit{Temporal Confusion}: Models frequently struggle with temporal localization, particularly evident in the poor Temporal Grounding (TG) scores. (2) \textit{Complex Reasoning Gaps}: While many models perform well in factual reasoning tasks, they often struggle with QAs requiring in-depth contextual understanding. (3) \textit{Context Integration}: The observed performance gap between Generic and Specific QAs suggests that models struggle to autonomously infer context for generic questions. Future models could benefit from improved mechanisms to integrate prior domain knowledge with visual data for more accurate general context recognition. (4) \textit{Hallucination in Response Generation}: Although some models demonstrate resilience to adversarial queries, hallucination remains a problem for Incompatible (IC) QAs, where irrelevant or out-of-domain answers are generated. Enhancing model training with stricter grounding mechanisms may reduce hallucinations, especially when faced with ambiguous or misleading inputs.

\noindent \textbf{RoadSocial improves road event understanding capability of general-purpose Video LLM:} We utilize the train split of our dataset and fine-tune a general-purpose Video LLM. Specifically, we selected LLaVA-OV-7B~\cite{li2024llava} parameter model as our baseline and employed standard instruction fine-tuning strategy wherein QA pairs are structured into instruction-tuned triplets (question, video, response). We adhere to the official training guidelines and optimized the model using a global batch size of 16 distributed over 16 NVIDIA H100 GPUs. During this phase, all key components (Vision tower, MLP adapter and LLM) were fine-tuned to optimize performance. Our evaluation results (last row of \cref{tab:eval_models_all}) shows that the fine-tuned LLaVA-OV-7B~\cite{li2024llava} model attains a significant jump of \textbf{19.4}\% in overall average score across all QA tasks and stands on par with the best performers. Specifically, the performance gains are significant across complex reasoning (DS, WY, CQ), introspection (IN), and Hallucination (AV, IC) tasks, showcasing our dataset's utility for improving road event understanding capabilities of general-purpose Video LLM.

\noindent \textbf{Ethical and Privacy Considerations:} Our data collection adheres to ethical guidelines, using only publicly available social media content. Our QA generation process includes rigorous checks to exclude potentially harmful, biased or inappropriate content in QA pairs, ensuring the dataset supports fair and responsible research in road event understanding. Additional details can be found in \cref{sec:refinement_and_categorization}.

\section{Conclusion}

RoadSocial redefines the landscape for general-purpose road event understanding. With a first-of-its-kind VideoQA dataset spanning \textbf{14M} frames and \textbf{414K} social comments, our dataset provides \textbf{13.2K} videos with \textbf{260K} high-quality QA pairs and \textbf{674} unique video tags (total \textbf{100K+}). By capturing diverse camera viewpoints, geographical contexts, and socially-informed QAs, RoadSocial delivers a comprehensive dataset that captures the complexity of real-world road scenarios across varied cultural and environmental contexts. Leveraging social media content, it addresses the limitations of traditional datasets by incorporating unique perspectives and nuanced social discourse. Our scalable semi-automatic annotation framework, powered by Text and Video LLMs, supports the creation of rich QA pairs across 12 challenging tasks. Given its scalable nature, our annotation framework can easily ingest and process social media posts generated continuously over time, enabling even larger dataset size with sustained quality. Our robust evaluation framework tests model resilience to irrelevant inputs, hallucinations, cross-viewpoint comprehension, and geographical awareness. Our evaluation across 18 Video LLMs provides critical performance insights across a spectrum of road event QA tasks. 

While RoadSocial is a rich resource for road event understanding, its reliance on social media data may introduce biases, skewing coverage towards regions with higher social media use. Apart from addressing these concerns, we envision several future directions for expanding RoadSocial such as increasing language diversity and establishing additional benchmark tasks. We believe RoadSocial will be instrumental in driving progress towards safer and more inclusive intelligent transportation systems.

\textbf{Acknowledgement.} Our sincere gratitude goes to late B.V. Khadiravana for his invaluable help in optimizing the execution of our experiments. The project was supported by iHub-Data and Mobility at IIIT Hyderabad.

{
    \small
    \bibliographystyle{ieeenat_fullname}
    \bibliography{main}
}


\clearpage
\appendix
\setcounter{page}{1}
\maketitlesupplementary

\onecolumn
\tableofcontents
\clearpage
\addcontentsline{toc}{section}{List of Figures}
\listoffigures
\clearpage

\twocolumn


\section{Data Collection}
\begin{figure*}[!t]
    \centering
    \includegraphics[width=0.98\textwidth]{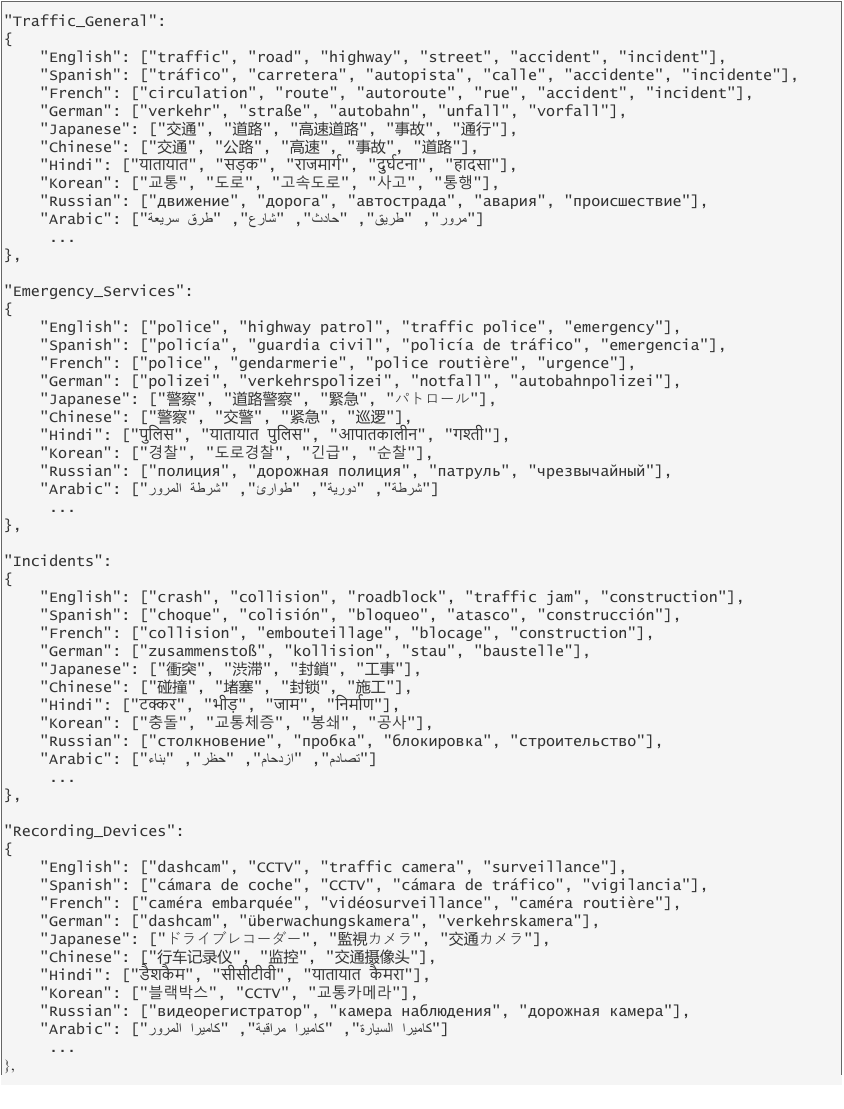}
    \label{fig:multi_lingual_1}
\end{figure*}

\begin{figure*}[!t]
    \centering
    \includegraphics[width=0.98\textwidth]{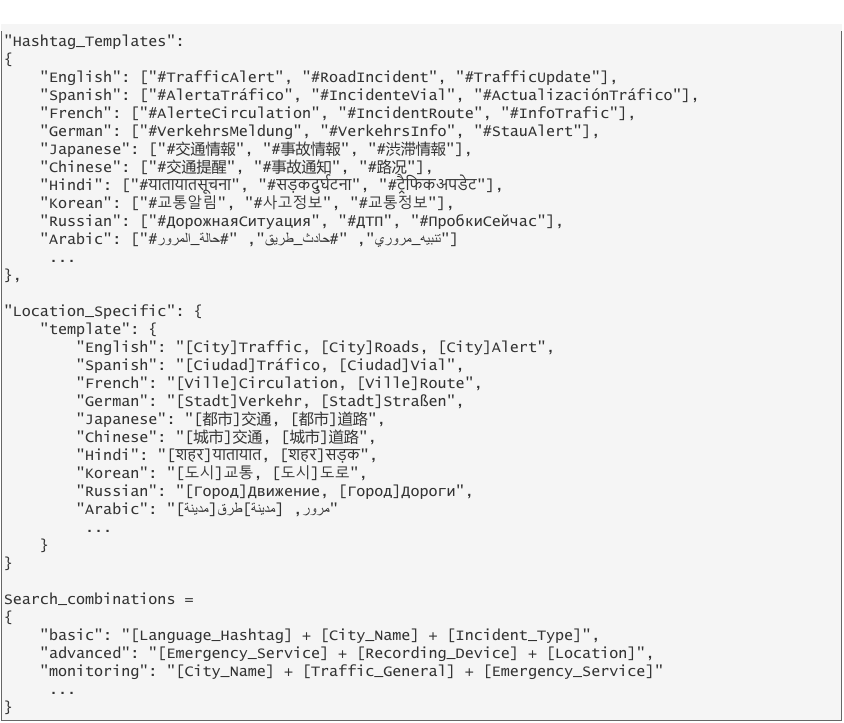}
    \caption{\textbf{Multilingual Traffic Keyword Dictionary for Tweet Mining}: A comprehensive dictionary of traffic-related keywords and hashtags, designed for identifying road event content on social media. Terms span traffic incidents, emergency services, recording devices, and location-specific templates. Effective usage involves combining terms across categories \textit{[Traffic\_General] + [Incidents]} and creating location-specific searches.}
    \label{fig:multi_lingual_2}
\end{figure*} 

\begin{figure*}[!t]
    \centering
    \includegraphics[width=\textwidth]{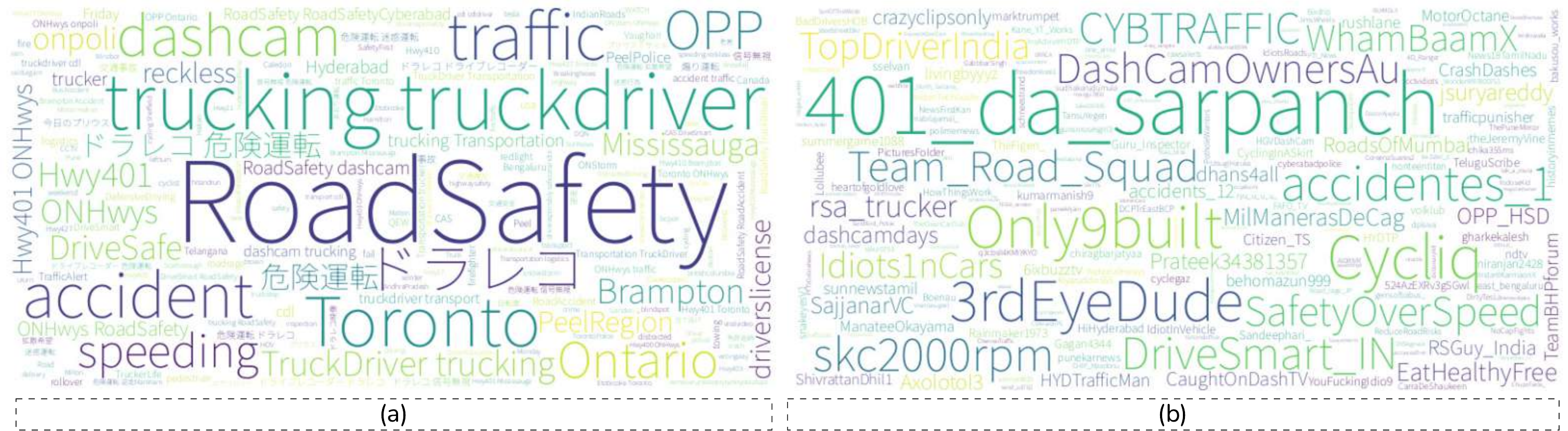}
    \caption{\textbf{Visualization of Our Dataset's Social Media Sources}: (a) Wordcloud of 3,385 unique hashtags mined iteratively from Twitter handles in our dataset, starting from initial accounts and expanding through network analysis of commonly used hashtags. (b) Wordcloud of Twitter handles from the 2,382 accounts discovered through this iterative hashtag mining process.}
    \label{fig:multi_lingual_keywords}
\end{figure*}

To identify relevant handles, we first created a multilingual keyword dictionary covering traffic terminology, emergency services, and regional variations (examples in \cref{fig:multi_lingual_2}). Using this dictionary, we manually identified key handle and analyzed their commonly used hashtags. Through hashtag mining and network analysis of these accounts, we discovered related accounts. This approach resulted in a total 2,382 accounts. We then scraped their content (videos, captions, and replies) from 2012 onwards. We programmatically filtered out tweets with fewer than four replies, retaining only those with substantial discussion. Representative hashtags and the handles are shown in \cref{fig:multi_lingual_keywords}. This systematic approach ensured the collection of road event content with significant community interaction. Full list of keywords, hashtags and handles will be released with dataset.

\section{Annotation Strategy: QAs and Tags}
\subsection{Identifying Representative Road Event Samples}
\label{sec:identify_representative_road_events}

\begin{figure*}[!t]
    \centering
    \includegraphics[width=\textwidth]{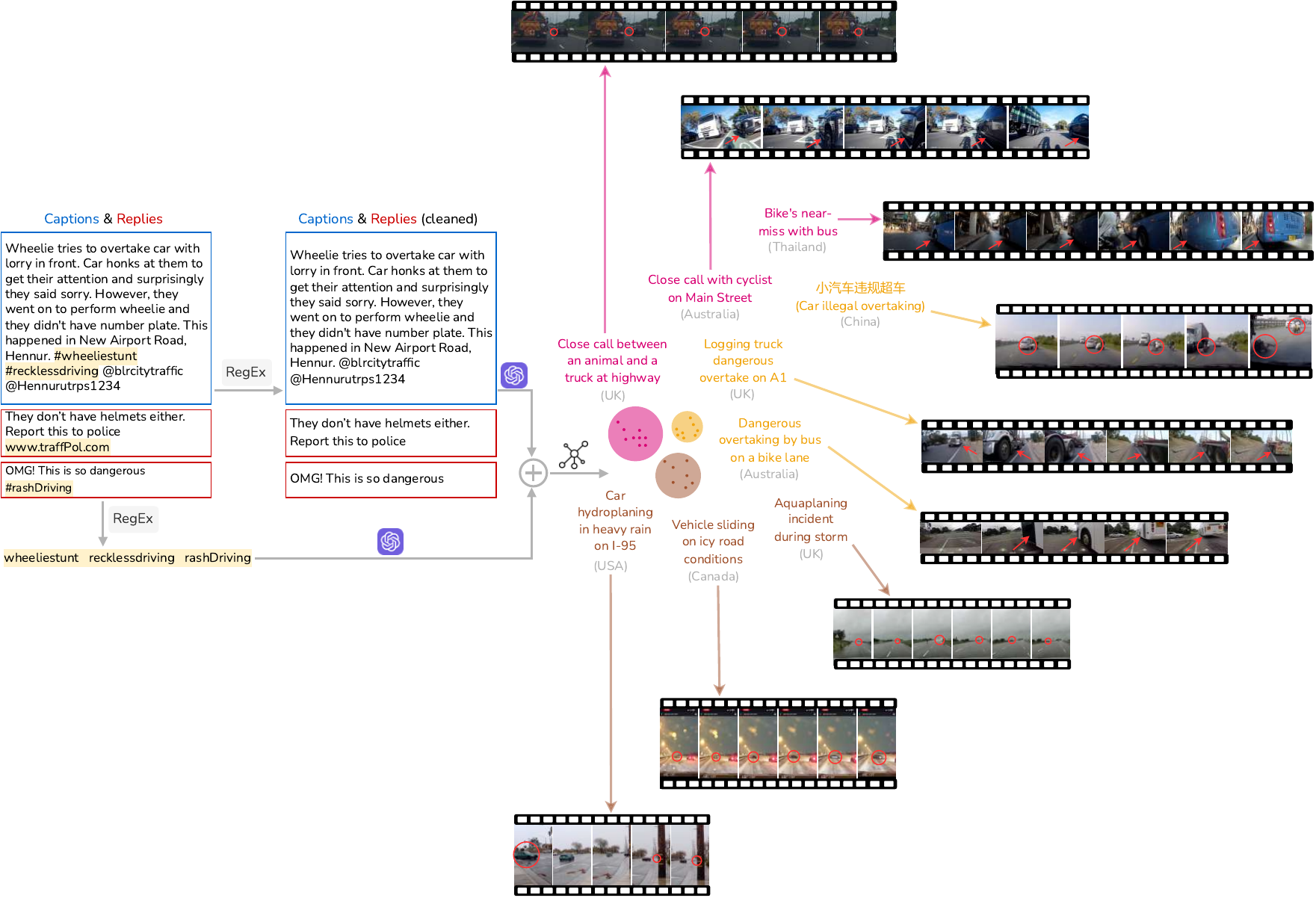}
    \caption{
        \textbf{Overview of Our Text Embedding and Clustering Pipeline}: 
        Left: RegEx-based cleaning is performed to separate tweet text from hashtags and URLs. Then GPT-3 embeddings (\raisebox{-1.9pt}{\includegraphics[width=0.02\textwidth, height=0.02\textwidth]{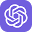}}) were computed separately for both cleaned text and hashtags before combination. Right: Resulting multilingual clusters of semantically similar road events via hybrid hierarchical k-means clustering (\raisebox{-2pt}{\includegraphics[width=0.02\textwidth, height=0.02\textwidth]{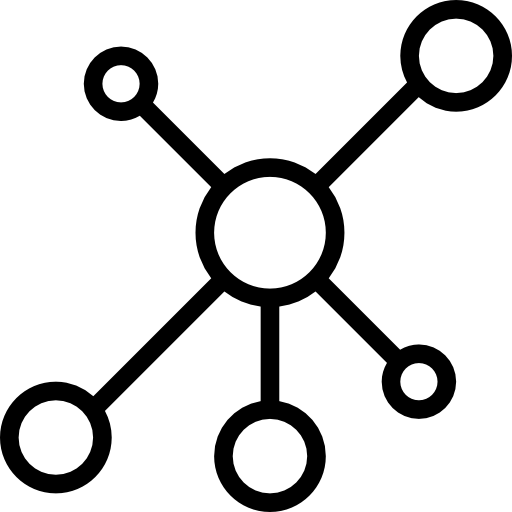}}). Refer back to \cref{sec:identify_representative_road_events}
    }
    \label{fig:hierar_clustering}
\end{figure*}

Our annotation strategy begins with identifying representative samples that capture the diversity of road events in our multilingual dataset. As shown in \cref{fig:hierar_clustering}, we first implement a text preprocessing pipeline where tweets undergo cleaning to remove URLs while preserving essential content. For instance, a tweet \texttt{Cyclist nearly hit by car \#OxfordStreet @MetPolice https://t.co/xyz} is reduced to \texttt{Cyclist nearly hit by car @MetPolice}. Concurrently, we extract and process hashtags separately, maintaining their semantic value by removing only the \texttt{\#} symbol (\eg \texttt{\#RoadSafety \#CyclingUK \#NearMiss} becomes \texttt{RoadSafety CyclingUK NearMiss}). For tweets lacking hashtags, we introduce a placeholder \texttt{\#NoHashTag}.
Using OpenAI's GPT-3 text embeddings API~\cite{openai2022gpt35}, we generate separate embeddings for cleaned text and processed hashtags. Our empirical analysis suggested that separately computing embeddings for cleaned text and hashtags, followed by their combination through averaging, yielded more representative sample clusters compared to alternatives such as embedding raw text or cleaned text alone.

These combined embeddings then undergo a hierarchical k-means clustering with a divisive approach (\cref{fig:hierar_clustering}). The process begins with a single cluster and iteratively creates sub-clusters based on silhouette scores. 
Specifically, after each k-means step, if the score improves or remained stable, we proceed to divide sub-clusters further; if it decreases significantly (indicating poor separation), we halt further splits for that branch of the hierarchy. This recursive process continues until reaching either a minimum cluster size or a predefined depth, with empirical analysis suggesting optimal results at 95 clusters.
This approach effectively groups similar road events across languages. For example, one cluster combines near-miss incidents like \texttt{Bike's near-miss with bus} (Thailand) and \texttt{Close call with cyclist on Main Street} (Australia), while another groups illegal overtaking events such as 
\texttt{Car illegal overtaking} from China and \texttt{Dangerous overtaking by bus on a bike lane} from Australia. Weather-related incidents form distinct clusters including \texttt{Car hydroplaning in heavy rain on I-95} (USA) and \texttt{Vehicle sliding on icy road conditions} (Canada).
From each cluster, we select five representative samples using a center-based approach. By computing the Euclidean distance between each sample and its cluster center, we identify the samples that best represent the cluster's core characteristics while maintaining linguistic and regional diversity. This systematic approach, validated through manual review, ensures our QA generation is grounded in well-represented events across our dataset.

\subsection{Template Question Generation}
\label{sec:template_q_generation}

\begin{figure*}[!t]
    \centering
    \includegraphics[width=\textwidth]{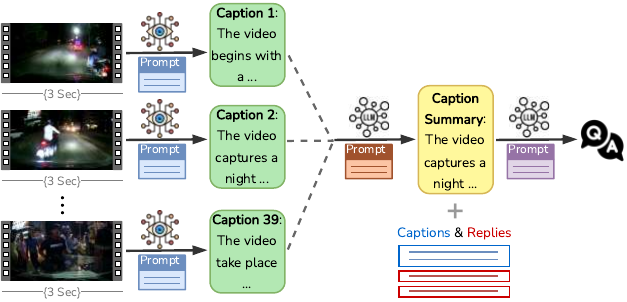}
    \caption{\textbf{Hybrid Approach for QA Generation Combining Visual and Social Context}: Left: Input video is segmented into 3-second clips, with Qwen2-VL (\raisebox{-2pt}{\includegraphics[width=0.02\textwidth, height=0.02\textwidth]{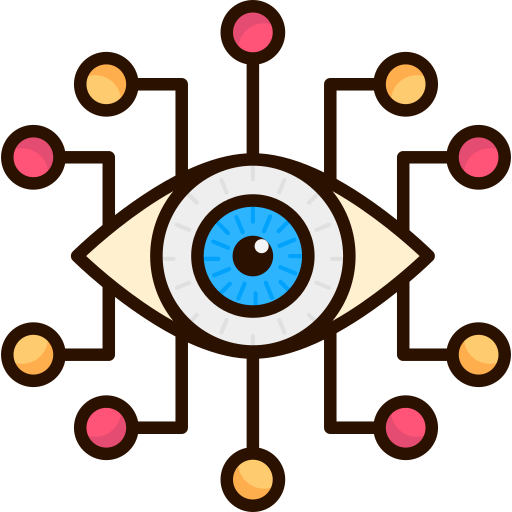}}) generating captions (\raisebox{-2pt}{\includegraphics[width=0.015\textwidth, height=0.017\textwidth]{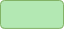}}) for each segment. Middle: Claude 3.5 Sonnet (\raisebox{-5pt}{\includegraphics[width=0.028\textwidth, height=0.028\textwidth]{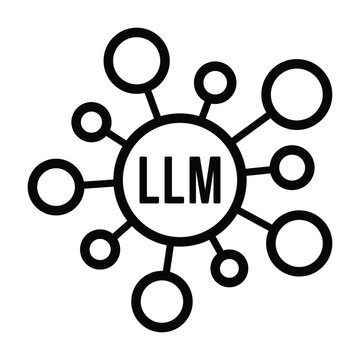}}) synthesizes these captions into a comprehensive video summary (\raisebox{-2pt}{\includegraphics[width=0.015\textwidth, height=0.017\textwidth]{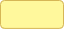}}). Right: Final prompt combines this summary with cleaned social media text (caption \& replies) to generate relevant QA pairs using template questions. The prompt to generate caption for a video-segment (\raisebox{-3.6pt}{\includegraphics[width=0.025\textwidth, height=0.025\textwidth]{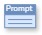}}) is illustrated in \cref{fig:prompts_1}. The full caption output (\raisebox{-2pt}{\includegraphics[width=0.015\textwidth, height=0.017\textwidth]{images/caption.png}}) for a video in our dataset is illustrated in \cref{fig:caption_1} - \ref{fig:caption_7}. The prompt to generate summary of a video from its segment captions (\raisebox{-3.6pt}{\includegraphics[width=0.025\textwidth, height=0.025\textwidth]{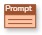}}) is illustrated in \cref{fig:prompts_2}. The full summary output (\raisebox{-2pt}{\includegraphics[width=0.015\textwidth, height=0.017\textwidth]{images/caption_summary.png}}) for the same video is illustrated \cref{fig:video_summary}. Also, the prompt that utilizes video summary, clean tweet text and template questions, to generate QA pairs corresponding to a video (\raisebox{-3.6pt}{\includegraphics[width=0.025\textwidth, height=0.025\textwidth]{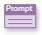}}), is illustrated in \cref{fig:prompt_3_a} - \ref{fig:prompt_3_b}. \raisebox{-2.8pt}{\includegraphics[width=0.02\textwidth, height=0.02\textwidth]{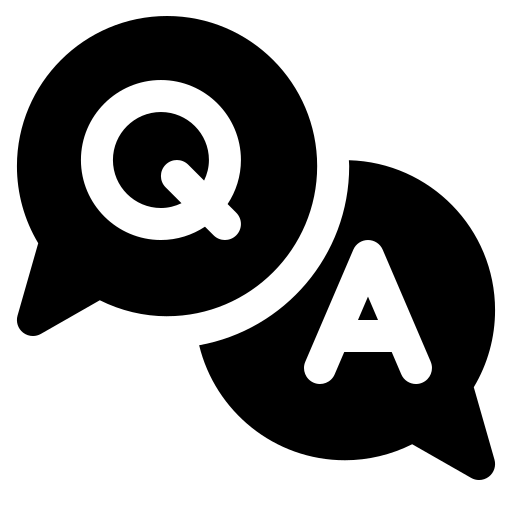}} represents the initially generated QA pairs which will further be modified and refined as discussed in upcoming subsections.
    }
    \label{fig:qa_gen_initial}
\end{figure*}

To develop comprehensive template questions for our dataset, we implemented an iterative approach based on analysis of representative video samples and their associated social media discourse. Following our hierarchical clustering process (\cref{fig:hierar_clustering}), we selected 5 representative videos from each of the 95 distinct clusters, creating a diverse corpus of 475 videos for detailed examination.

\noindent \textit{Formulating fundamental questions}: In the initial phase, we conducted manual analysis of the selected videos and their associated tweet conversations, focusing on fundamental aspects of road events. We began by formulating basic questions such as \texttt{What road event took place in the video?}

\noindent \textit{Formulating analysis questions}: We expanded our template set based on patterns observed in social media discussions. For example, in videos involving accidents and near-misses, conversations were frequently centered on causal analysis. This observation led us to develop questions specifically probing the potential causes and motivations behind road events, such as \texttt{What was the primary reason behind the occurrence of this incident?} Similarly, discussion around post-crash measures in relevant scenarios, led the inclusion of template questions addressing response actions such as \texttt{What measures should be taken after witnessing an event like this?}

\noindent \textit{Template refinement}: The template refinement process was inherently iterative, with each round of video analysis contributing to the evolution of our question set. A key consideration was maintaining question generalizability while preserving specificity where necessary. For instance, certain questions (\eg those about accident causation) were not universally applicable across our diverse video corpus. This recognition prompted us to reformulate the questions to ensure broader applicability. For instance, accident-related questions were reframed conditionally: \texttt{If the road event involves an accident or a near-miss incident, explain how it could have been prevented.} We also incorporated universally applicable questions about recording devices (\eg \texttt{What type of camera was used to capture the video?}) and geographical context (\eg \texttt{In which country did this road event take place?}).

\noindent \textit{Spatial and temporal aspects in questions}: Furthermore, we carefully structured the questions to address both spatial and temporal aspects of road events. Spatial questions could be answered through single-frame analysis (\eg \texttt{In which country did this road event take place?} or \texttt{What were the weather conditions or road visibility when the video was captured?}). While temporal questions inquire about event sequences and interactions (\eg \texttt{Describe the actions performed by the road entities involved in the road event} or \texttt{Specify the approximate time interval where the key road event is observed in the video?}). This dual approach ensures comprehensive coverage of both spatial and temporal dimensions of road events.

The final set of 18 carefully curated questions, shown in \cref{fig:prompt_3_a} - \ref{fig:prompt_3_b}, were integrated into our LLM prompting strategy which is described in the next subsection.

\subsection{QA Generation via Hybrid Approach}
\label{sec:hybrid_approach}

\begin{figure*}[!t]
    \centering
    \includegraphics[width=\textwidth]{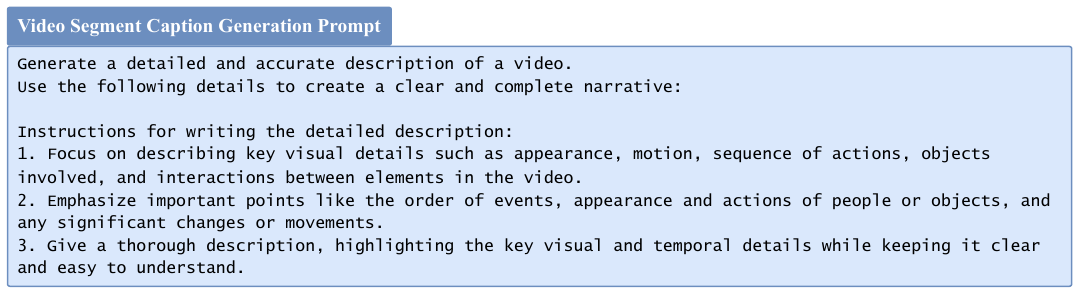}
    \caption{Prompt design for generating segment-wise captions using Qwen2-VL Video LLM. The model generates detailed descriptions for each 3-second video segment, capturing temporal visual information. Refer back to \cref{fig:qa_gen_initial} or \cref{sec:hybrid_approach}.}
    \label{fig:prompts_1}
\end{figure*}

\begin{figure*}[!t]
    \centering
    \includegraphics[width=\textwidth]{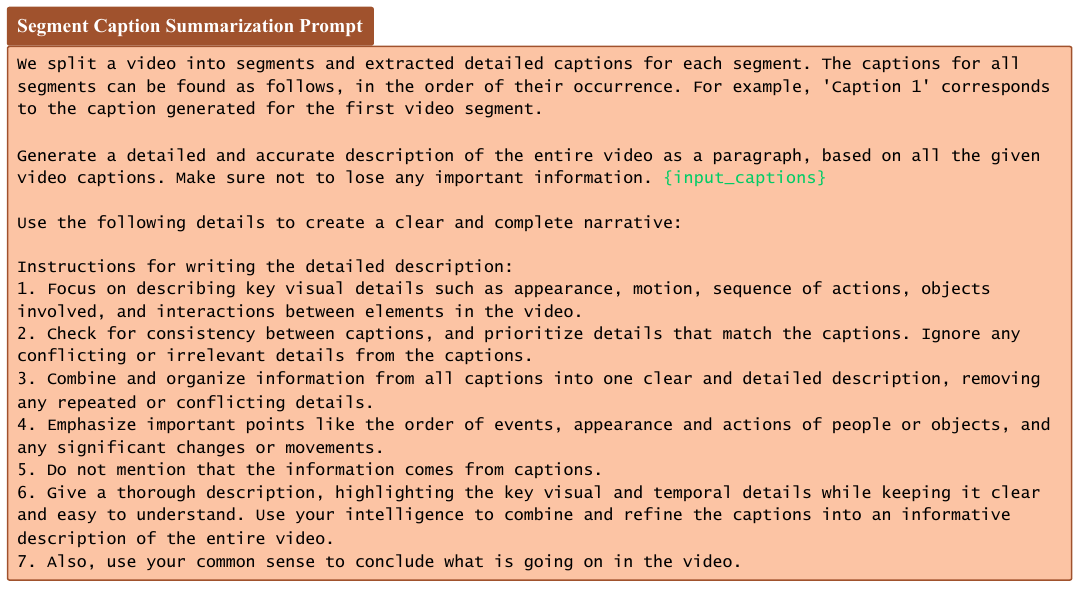}
    \caption{Prompt template for generating cohesive video summaries using Claude 3.5 Sonnet. The Text LLM combines segment-wise captions to create a comprehensive temporal description of the entire video. Refer back to \cref{fig:qa_gen_initial} or \cref{sec:hybrid_approach}.}
    \label{fig:prompts_2}
\end{figure*}

\begin{figure*}[!t]
    \centering
    \includegraphics[width=0.95\textwidth]{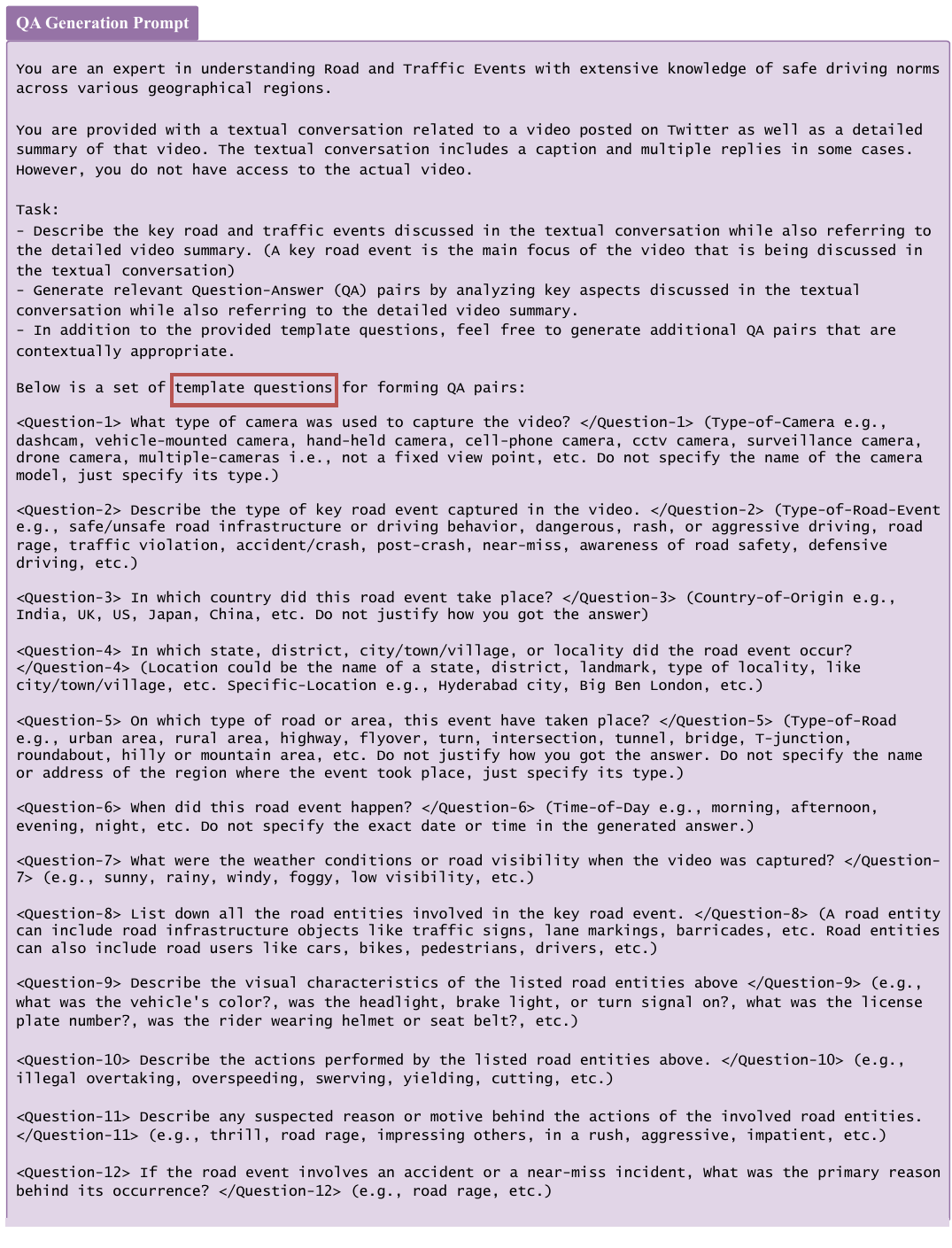}
    \caption*{}  
    \phantomcaption  
    \label{fig:prompt_3_a}
\end{figure*}

\begin{figure*}[!t]
    \centering
    \includegraphics[width=\textwidth]{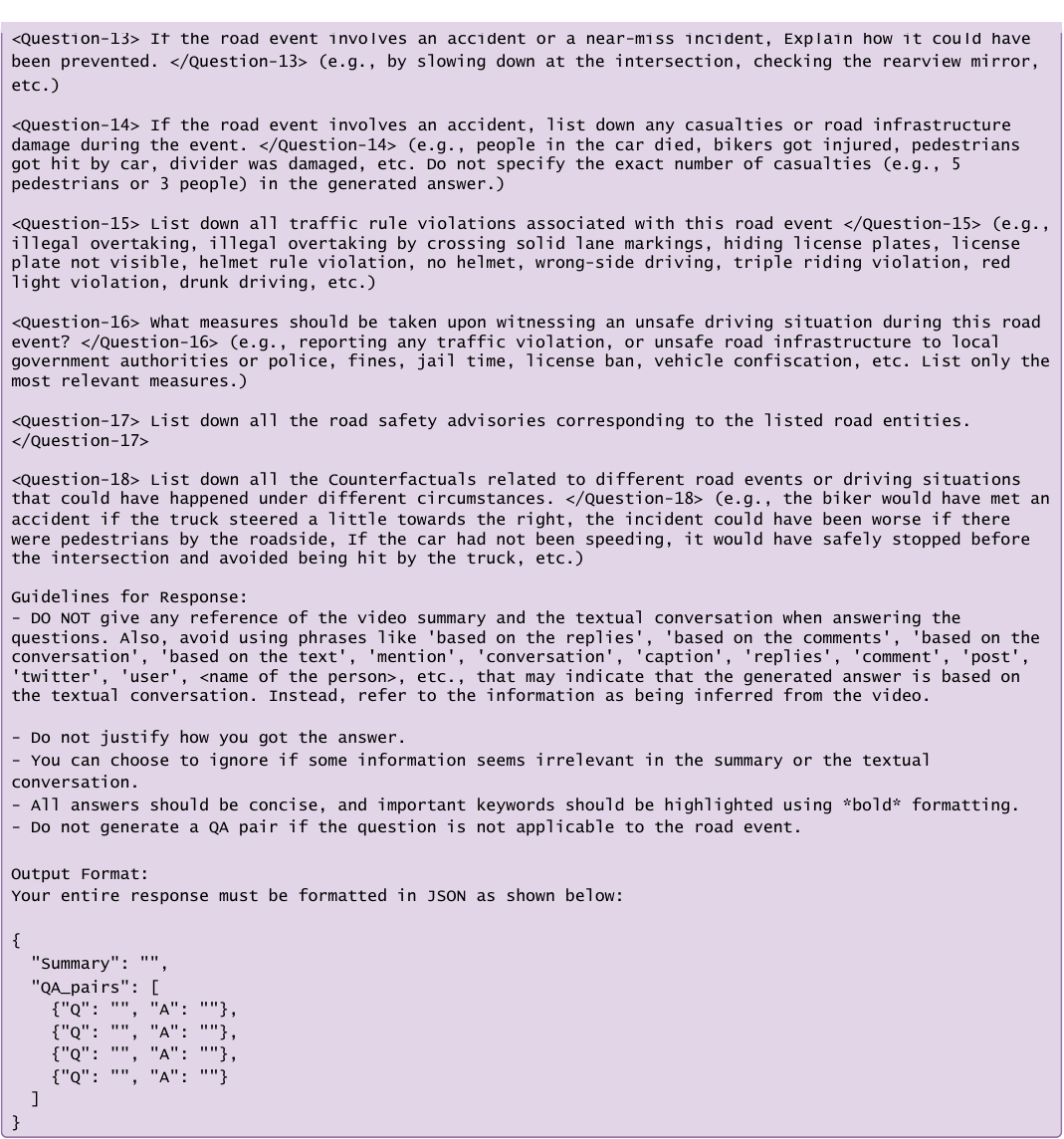}
    \caption{Complete QA generation prompt utilizing both video summary and social media context. Template questions guide Claude 3.5 Sonnet to generate relevant question-answer pairs capturing both visual and social context. Refer back to \cref{fig:qa_gen_initial}.}
    \label{fig:prompt_3_b}
\end{figure*}

\begin{figure*}[!t]
    \centering
    \includegraphics[width=0.95\textwidth]{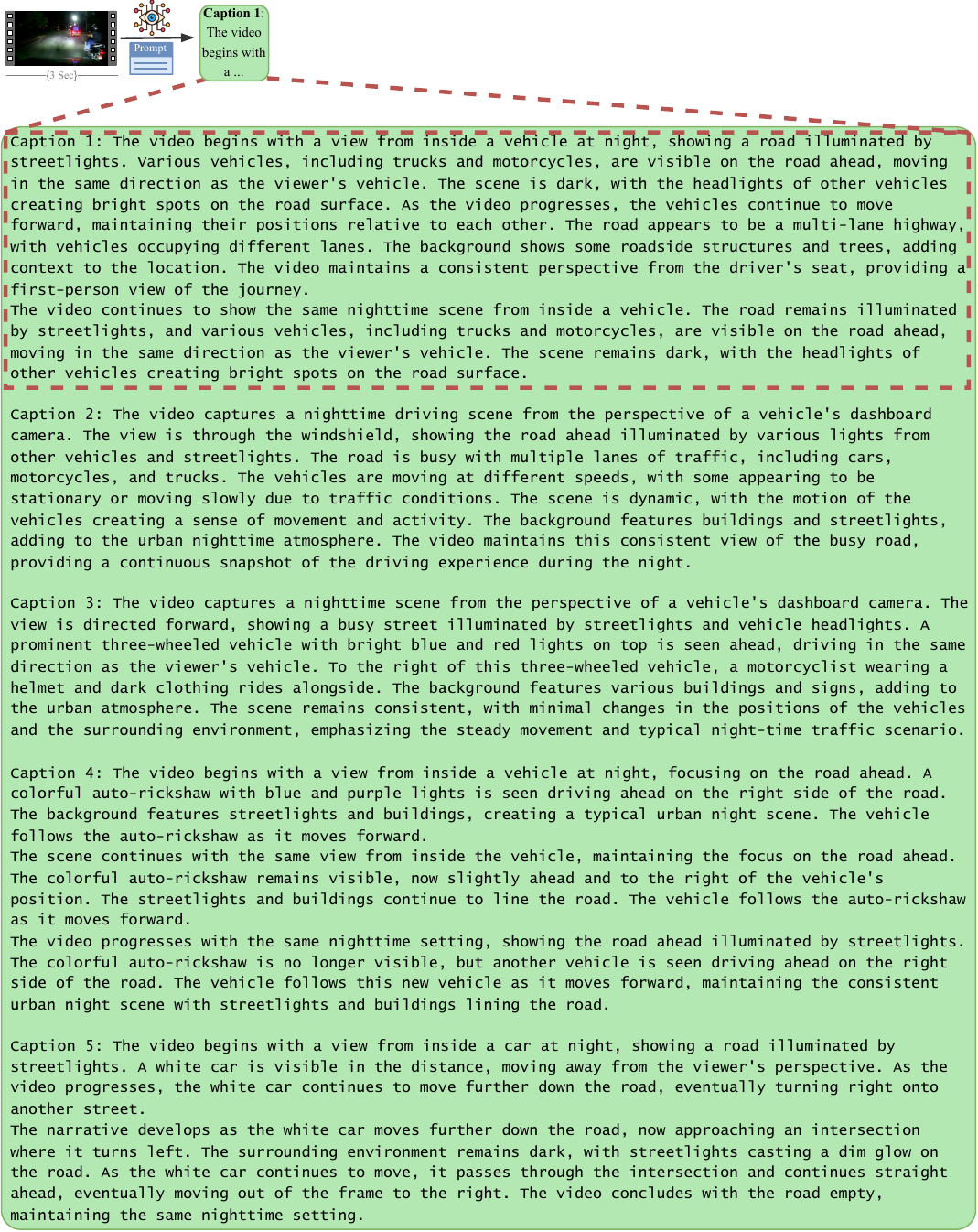}
    \caption*{}  
    \phantomcaption  
    \label{fig:caption_1}
\end{figure*}

\begin{figure*}[!t]
    \centering
    \includegraphics[width=\textwidth]{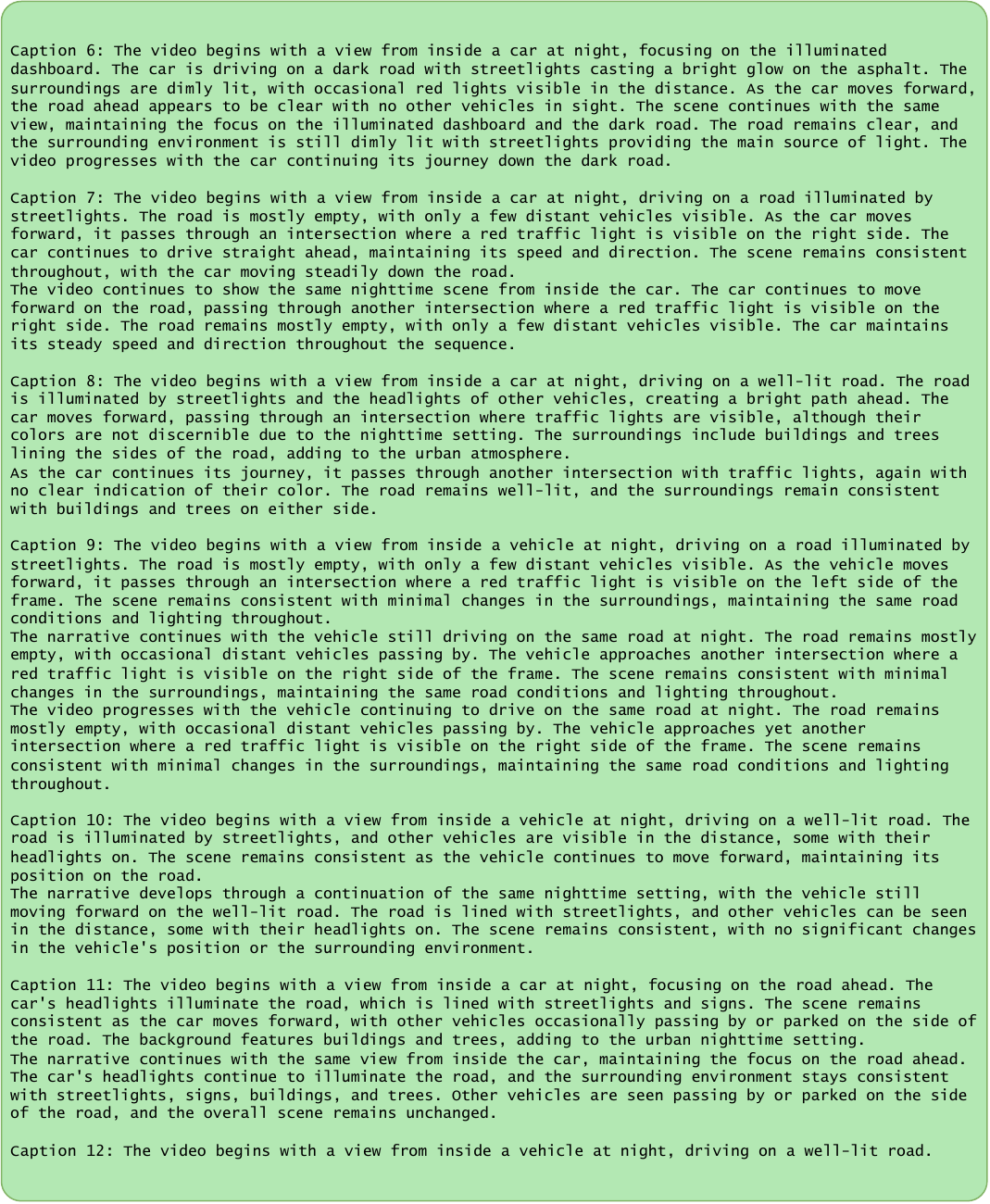}
    \label{fig:caption_2}
\end{figure*}

\begin{figure*}[!t]
    \centering
    \includegraphics[width=\textwidth]{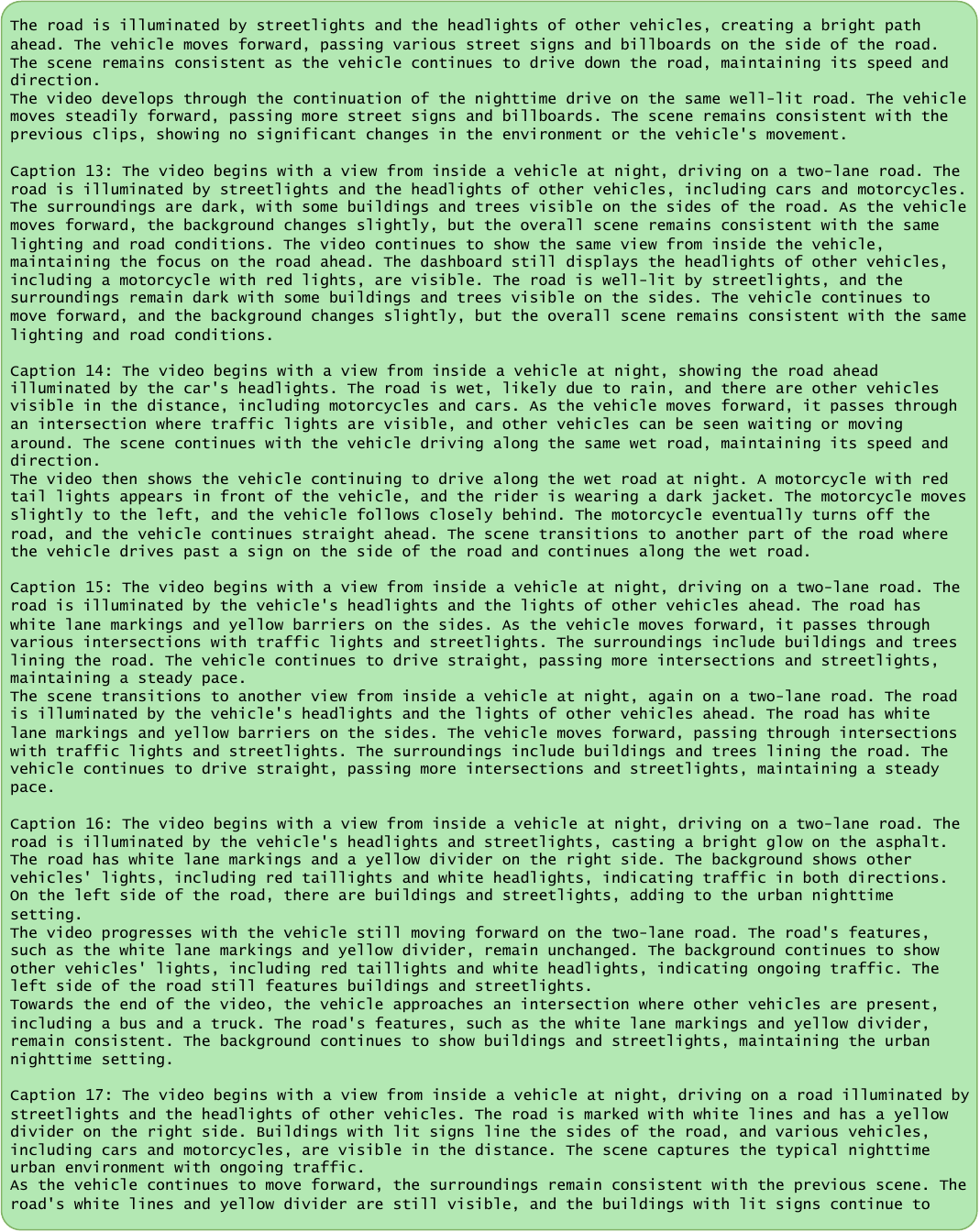}
    \label{fig:caption_3}
\end{figure*}

\begin{figure*}[!t]
    \centering
    \includegraphics[width=\textwidth]{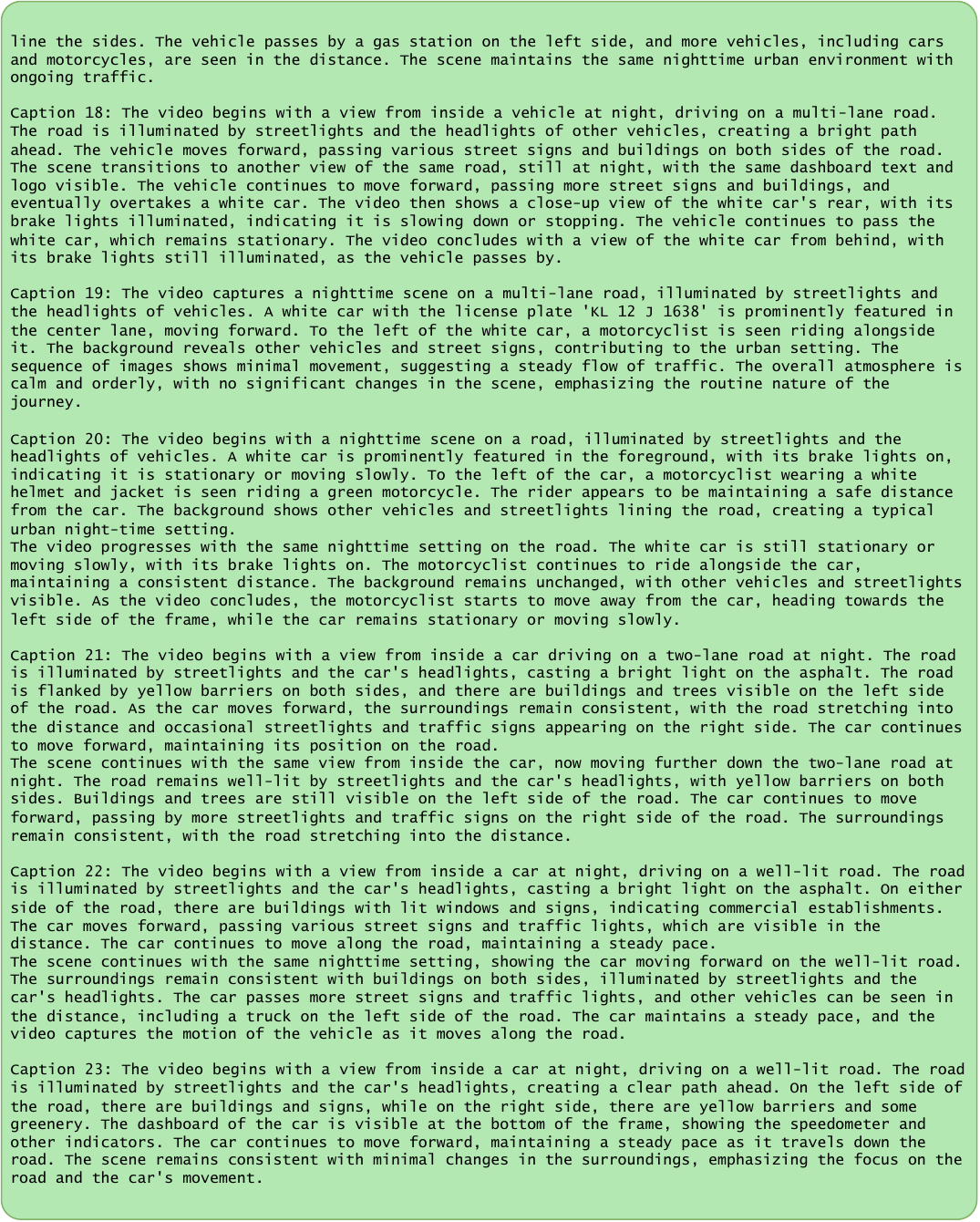}
    \label{fig:caption_4}
\end{figure*}

\begin{figure*}[!t]
    \centering
    \includegraphics[width=\textwidth]{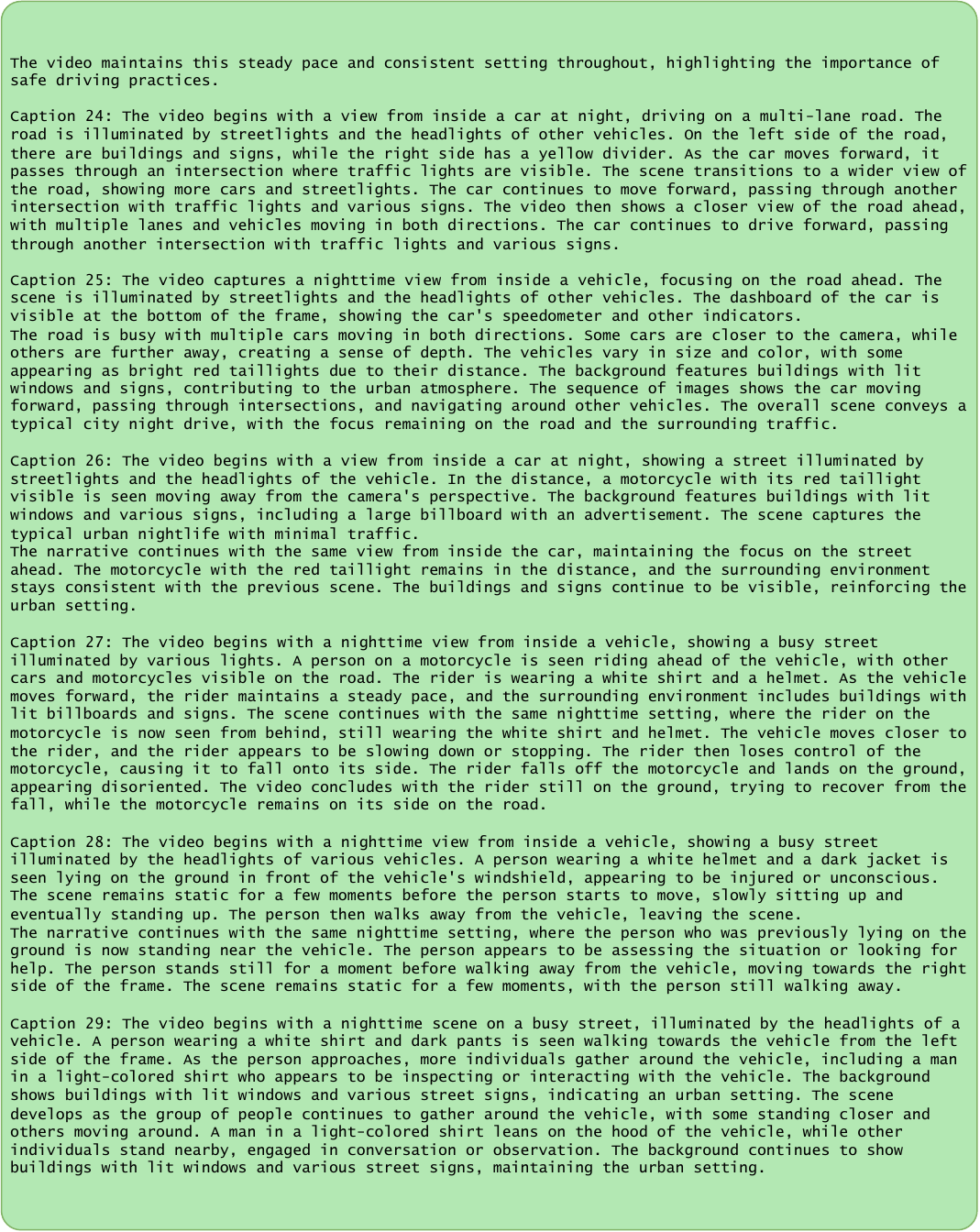}
    \label{fig:caption_5}
\end{figure*}

\begin{figure*}[!t]
    \centering
    \includegraphics[width=\textwidth]{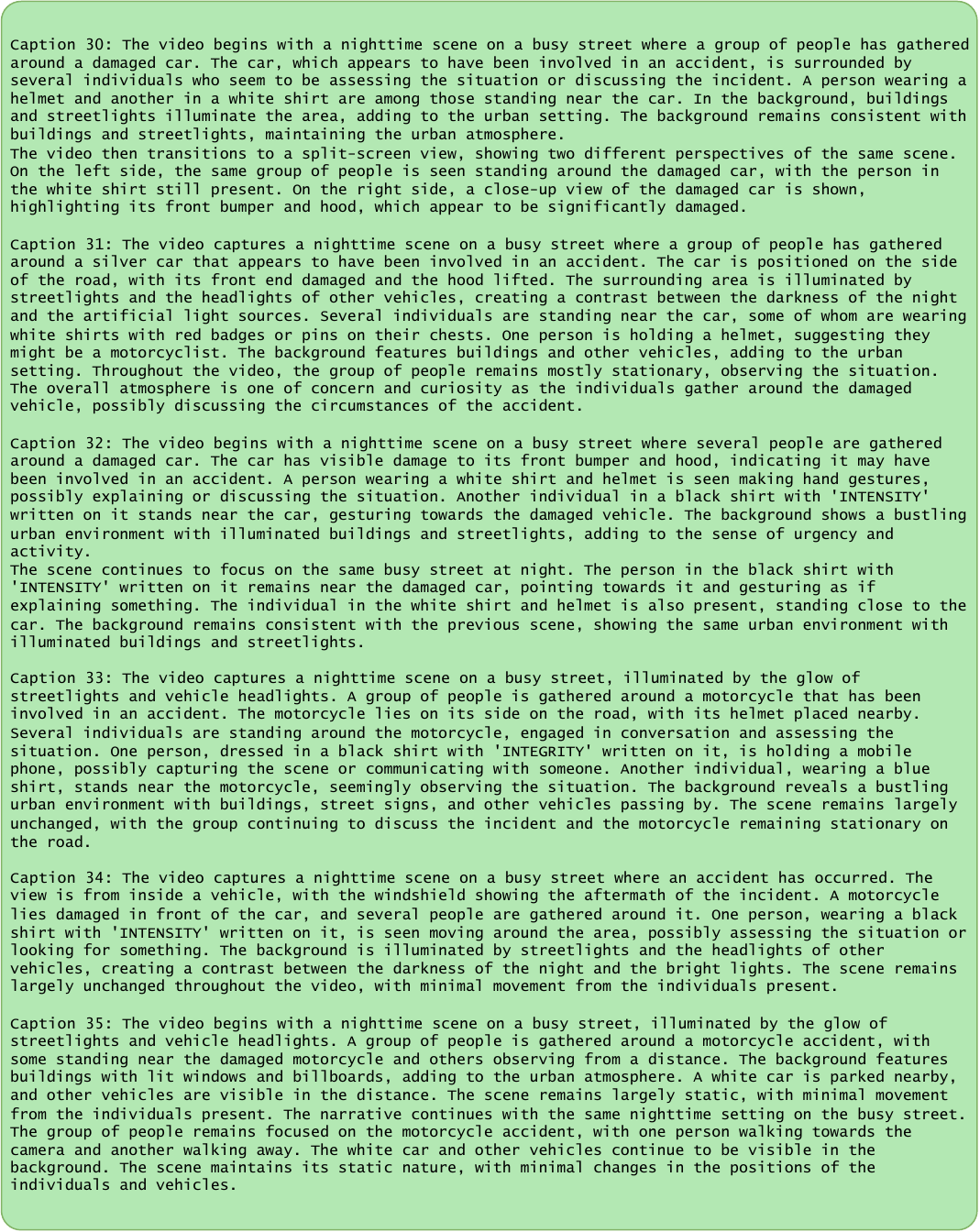}
    \label{fig:caption_6}
\end{figure*}

\begin{figure*}[!t]
    \centering
    \includegraphics[width=\textwidth]{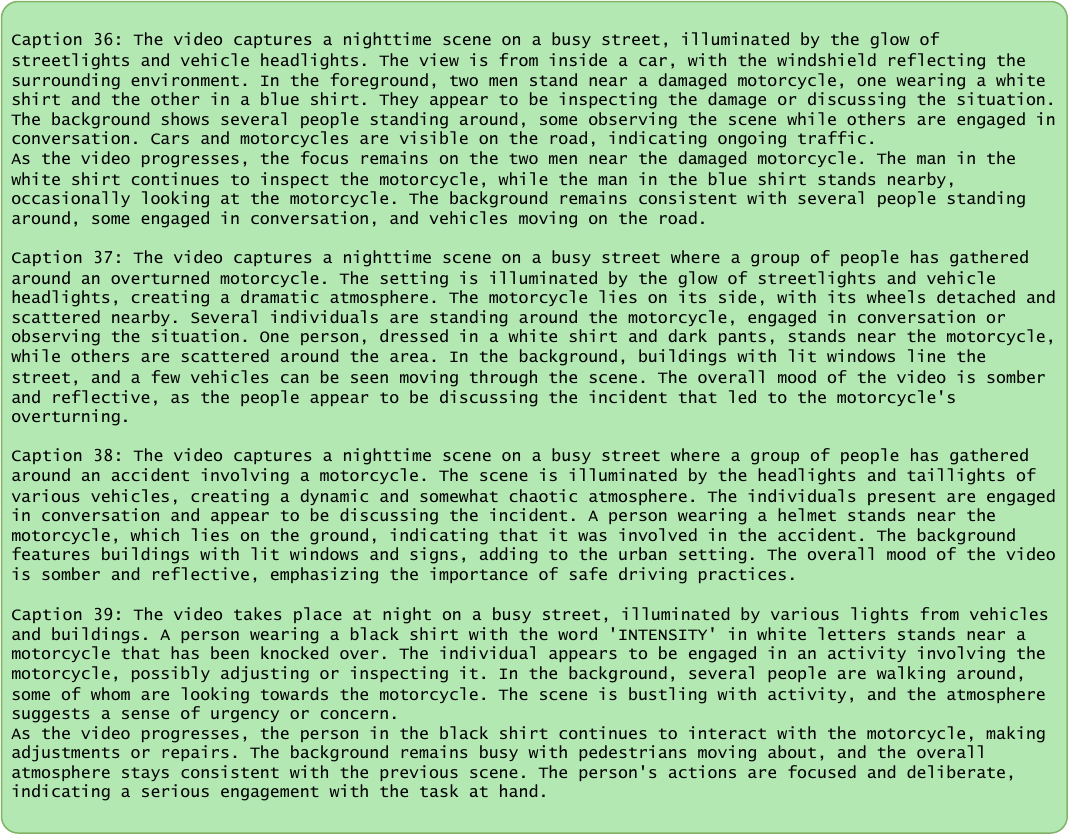}
    \caption{Example of generated captions for each of the three-second segment of the video via the pipeline demonstrated in \cref{fig:qa_gen_initial}}
    \label{fig:caption_7}
\end{figure*}

\begin{figure*}[!t]
    \centering
    \includegraphics[width=\textwidth]{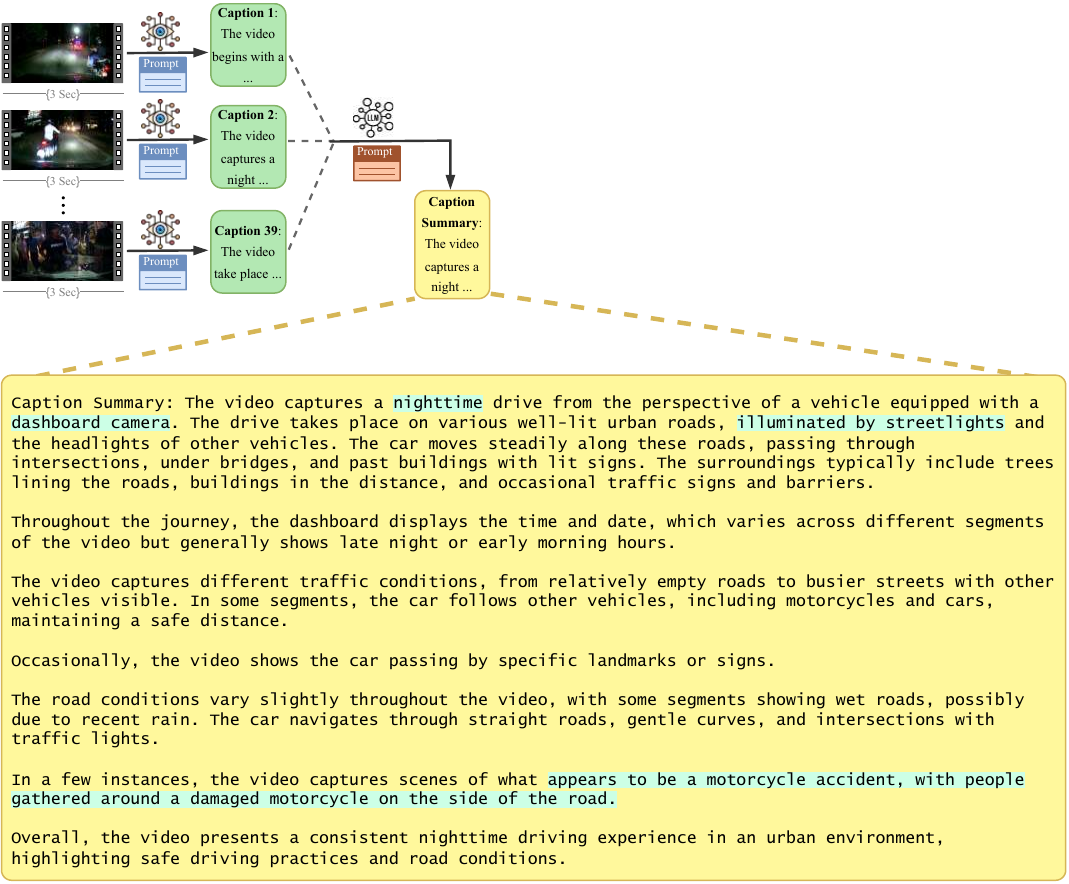}
    \caption{Example of generated video summary from the captions via Text LLM, via the pipeline demonstrated in \cref{fig:qa_gen_initial}.}
    \label{fig:video_summary}
\end{figure*}


\begin{figure*}[!t]
    \centering
    \includegraphics[width=\textwidth]{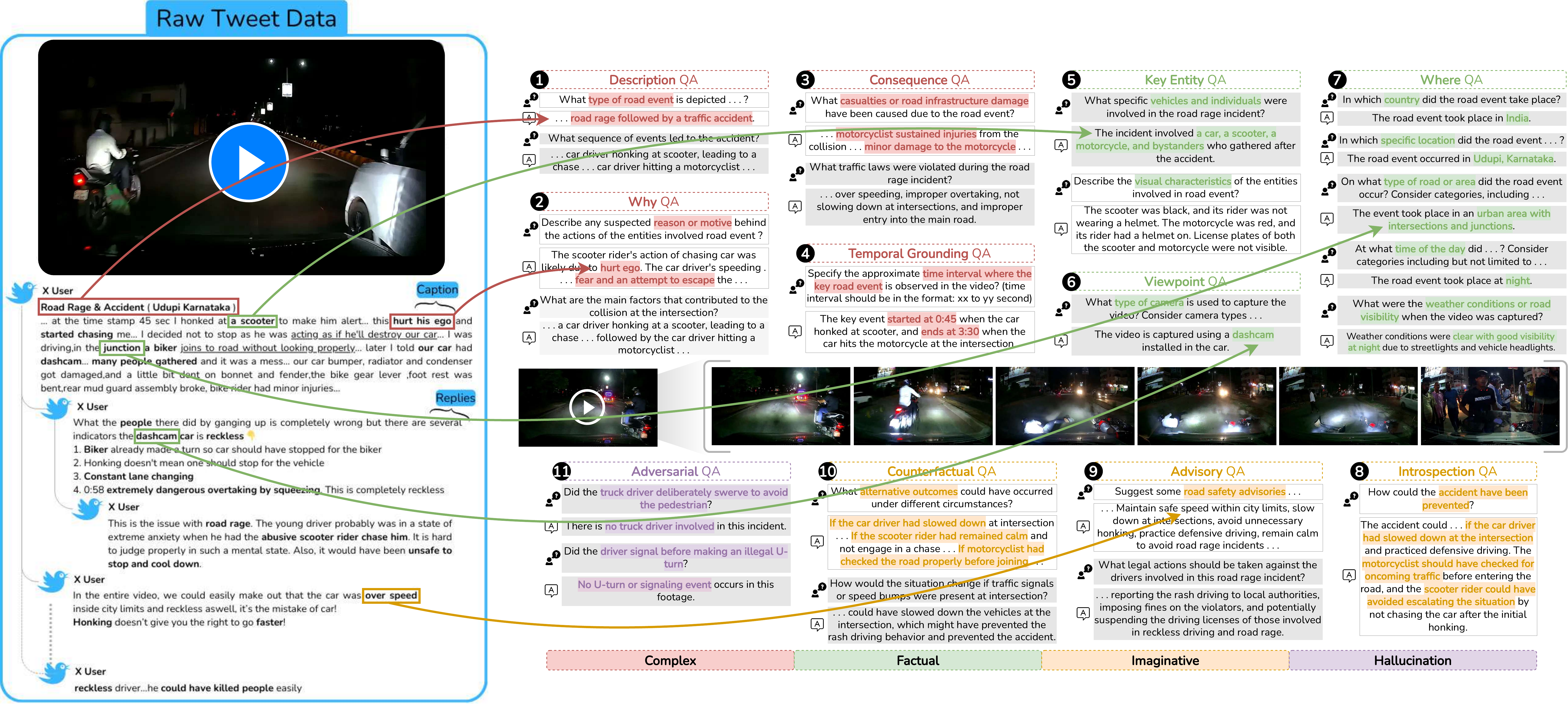}
    \caption{\textbf{Demonstrating the importance of hybrid information sources for QA generation.} While the video summary (\cref{fig:video_summary}) captures basic visual elements (\eg `nighttime', `streetlight illumination'), the tweet conversation provides crucial contextual information (shown by colored boxes) missing from the visual description alone. This example illustrates why our approach combines both video summaries and social media context to generate diverse and socially-informed QA pairs.
    }
    \label{fig:social_context_utility}
\end{figure*}

To generate question-answer pairs for each video, we developed a hybrid approach that leverages both visual content and social media context. Our pipeline, illustrated in \cref{fig:qa_gen_initial}, consists of three main stages that systematically combine video understanding with social context.

First, we extract visual semantics by splitting each video into 3-second segments and employ Qwen2-VL Video LLM~\cite{wang2024qwen2} to generate detailed captions for each segment. The prompt (\raisebox{-3.6pt}{\includegraphics[width=0.025\textwidth, height=0.025\textwidth]{images/prompts_1_icon.jpg}}) to generate caption (\raisebox{-2pt}{\includegraphics[width=0.015\textwidth, height=0.017\textwidth]{images/caption.png}}) for a video-segment is illustrated in \cref{fig:prompts_1}. This temporal segmentation ensures capture of fine-grained details and event progression. Next, these segment-wise captions are processed by Claude 3.5 Sonnet Text LLM~\cite{claude_3_5} to generate a cohesive, visually-rich summary of the entire video. The prompt (\raisebox{-3.6pt}{\includegraphics[width=0.025\textwidth, height=0.025\textwidth]{images/prompts_2_icon.jpg}}) to generate summary of a video from its segment captions is illustrated in \cref{fig:prompts_2}. Finally, we combine this generated summary (\raisebox{-2pt}{\includegraphics[width=0.015\textwidth, height=0.017\textwidth]{images/caption_summary.png}}) with cleaned tweet text (captions \& replies) to create contextually rich QA pairs using template questions through Claude 3.5 Sonnet. The prompt (\raisebox{-3.6pt}{\includegraphics[width=0.025\textwidth, height=0.025\textwidth]{images/prompts_3_icon.jpg}}) that utilizes video summary, clean tweet text and template questions, to generate QA pairs corresponding to a video, is illustrated in \cref{fig:prompt_3_a} - \ref{fig:prompt_3_b}. References for inputs and outputs at each stage of this pipeline are provided in the \cref{fig:qa_gen_initial}. \cref{fig:social_context_utility} demonstrates the utility of social conversation in QA formation.

\subsection{Specific QA Generation}
\label{sec:specific_qa_generation}

To create a comprehensive question set with varying difficulty levels, we developed an approach for generating specific questions from generic template set (\cref{fig:prompt_3_a} - \ref{fig:prompt_3_b}). While generic questions like \texttt{What actions were performed by the road entities involved in the key road event?} require complex temporal reasoning and synthesis of multiple observations, specific questions such as \texttt{How was the truck involved in the accident?} focus on particular entities and events, offering more straightfoward path for answer formulation.

\begin{figure*}[!t]
    \centering
    \includegraphics[width=\textwidth]{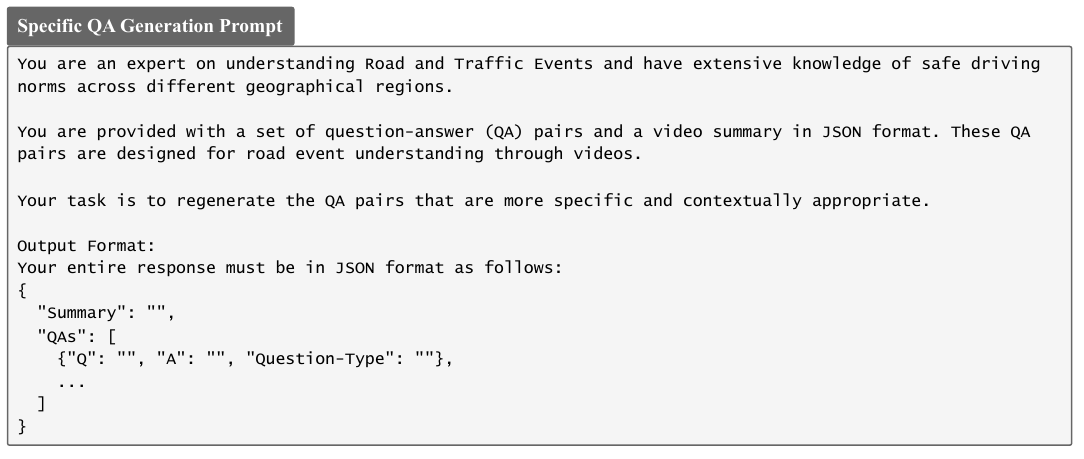}
    \caption{Prompt design for generating specific QA pairs from generic templates. The expert-driven prompt transforms generic questions into contextually specific ones while maintaining alignment with predefined categories (e.g., Camera Device, Road Event Type, Actions). The prompt takes generic QA pairs and video summaries as input and generates specific questions that capture detailed aspects of road events while preserving the taxonomic structure. Refer back to \cref{sec:specific_qa_generation}.}
    \label{fig:prompts_specific}
\end{figure*}

We developed a specialized prompt that instructs the LLM (Claude 3.5 Sonnet) to act as an expert with comprehensive knowledge of driving norms across different geographical regions. The prompt takes two inputs: the generic QA pairs generated from our initial template-based approach (\cref{fig:prompt_3_a} - \ref{fig:prompt_3_b}) and the corresponding video summary (\eg \cref{fig:video_summary}) to generate contextually appropriate specific questions. The prompt (\raisebox{-3.6pt}{\includegraphics[width=0.025\textwidth, height=0.025\textwidth]{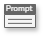}}) to generate specific QA pairs is illustrated in \cref{fig:prompts_specific}.

\subsection{QA Refinement and Categorization}
\label{sec:refinement_and_categorization}

\begin{figure*}[!t]
    \centering
    \includegraphics[width=0.93\textwidth]{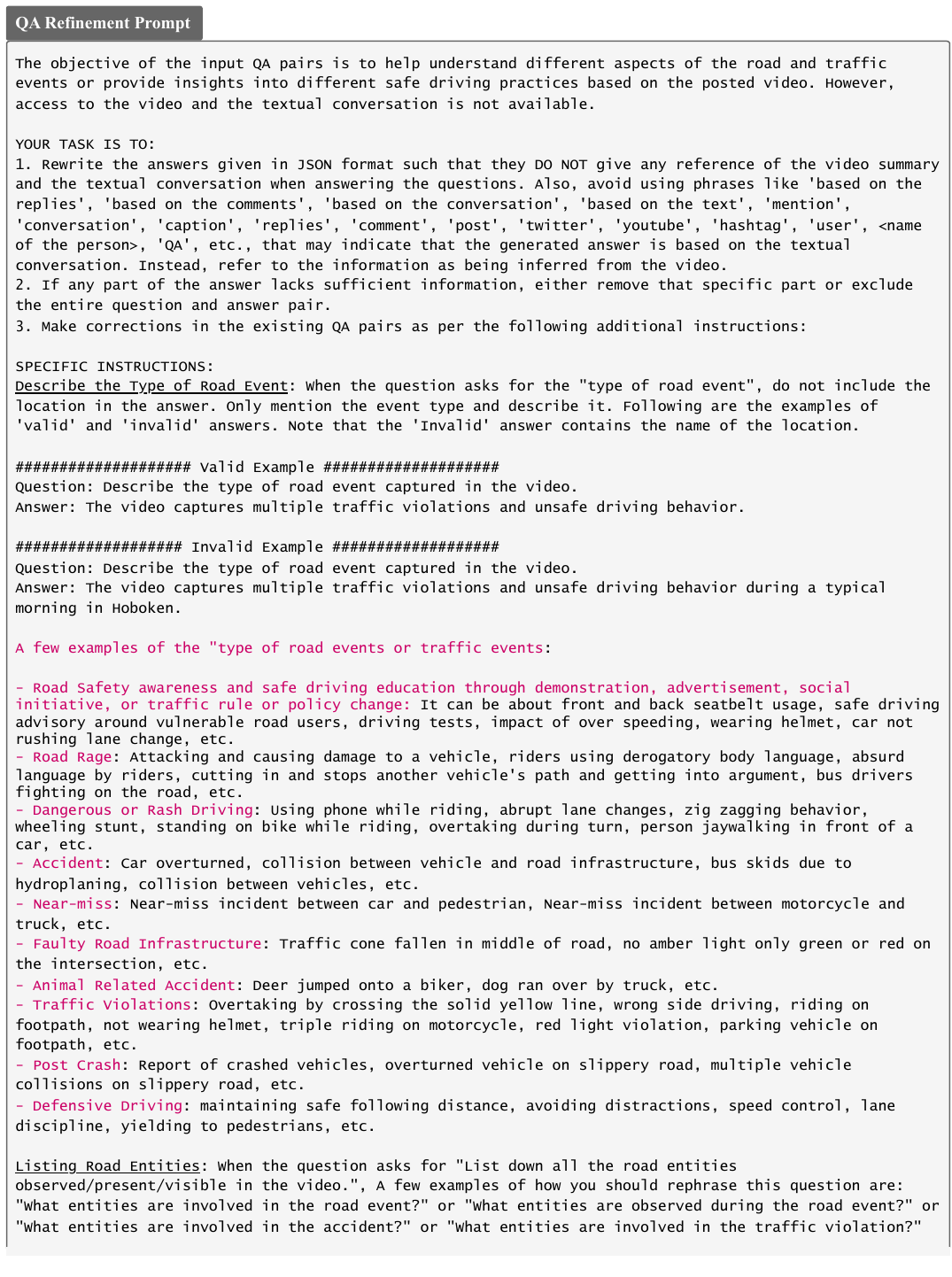}
    \caption*{}  
    \phantomcaption  
    \label{fig:prompts_refine_1}
\end{figure*}

\begin{figure*}[!t]
    \centering
    \includegraphics[width=\textwidth]{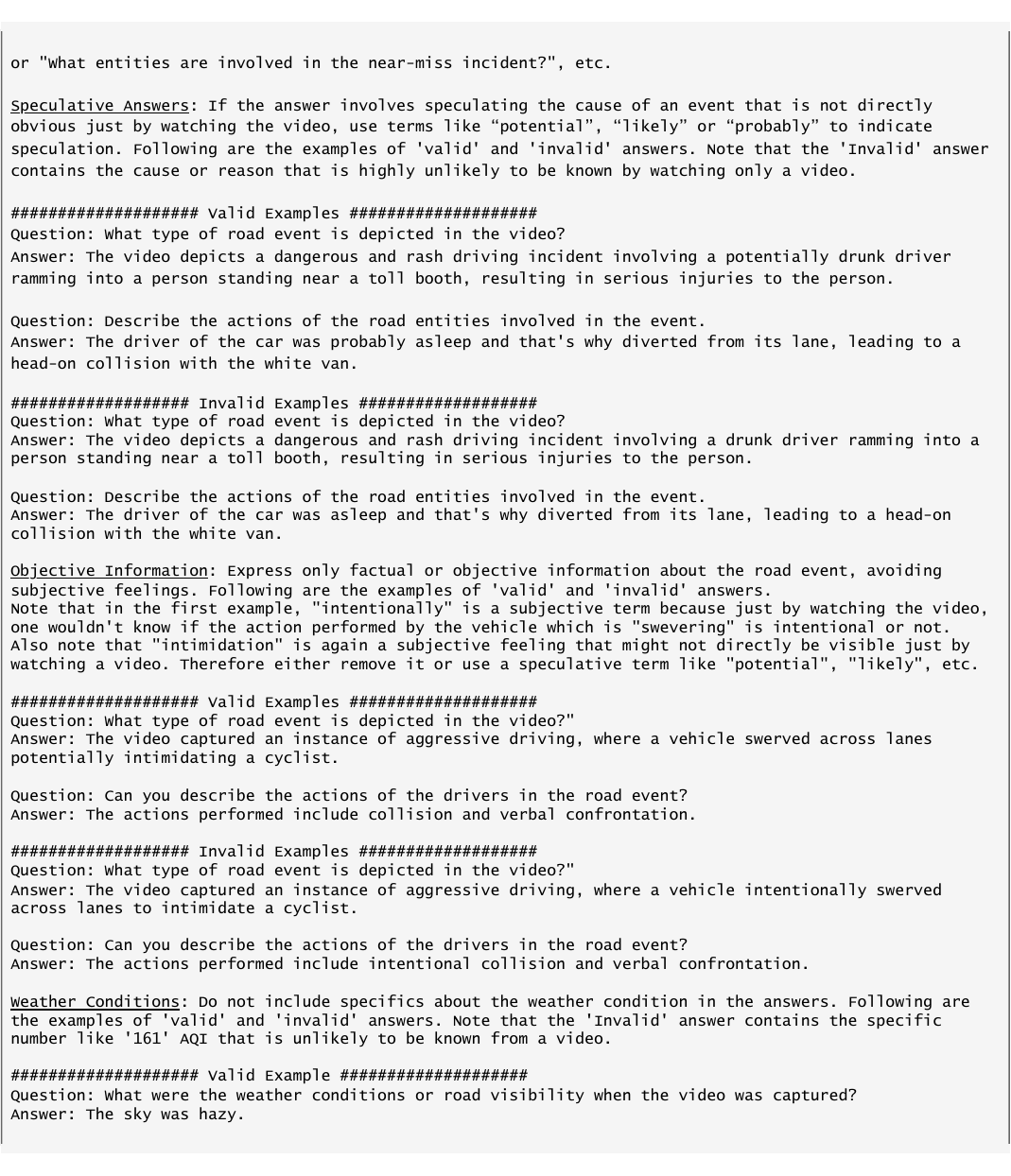}
    \label{prompts_refine_2}
\end{figure*}

\begin{figure*}[!t]
    \centering
    \includegraphics[width=\textwidth]{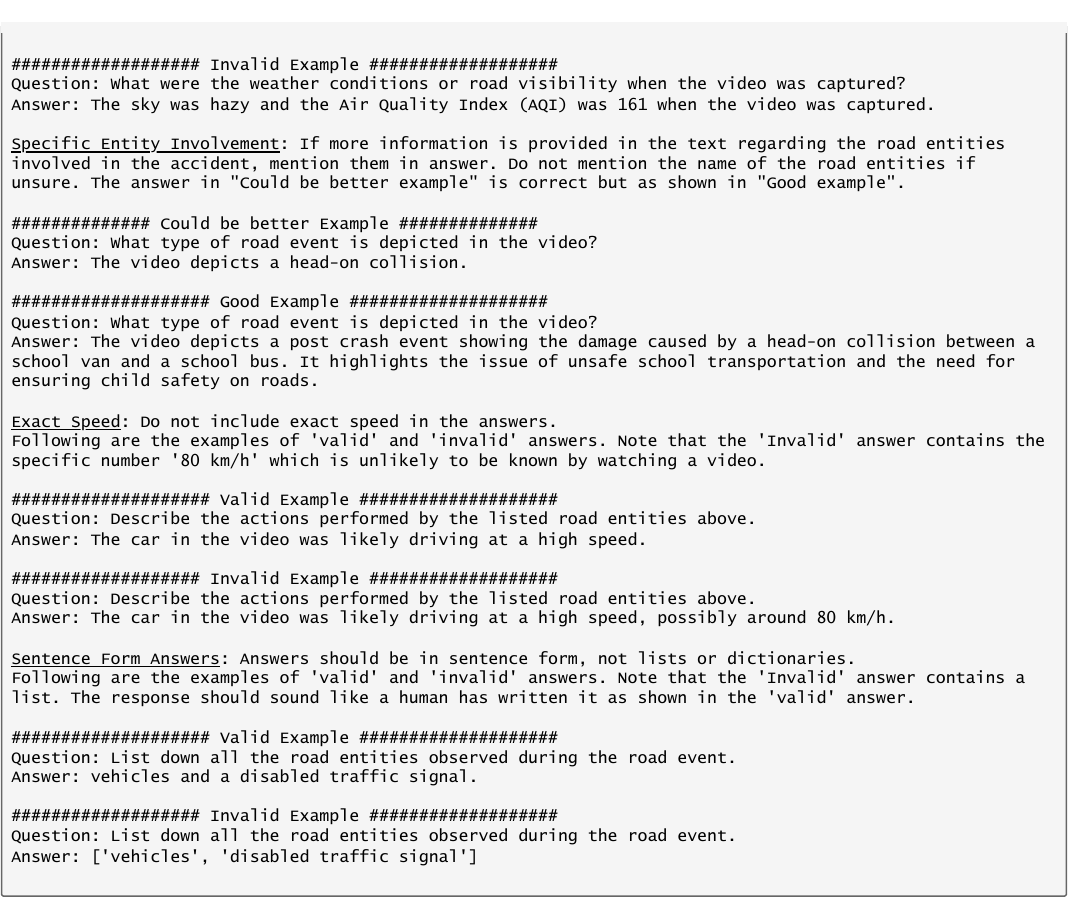}
    \caption{\textbf{QA refinement prompt}: The prompt implements comprehensive guidelines for (1) removing social media references, (2) standardizing temporal information, (3) enforcing speculative language for non-obvious causation, (4) maintaining objective observations, and (5) ensuring human-like sentence-form responses. Refer back to \cref{sec:refinement_and_categorization}}
    \label{fig:prompts_refine_3}
\end{figure*}

\noindent \textbf{QA Refinement}: To ensure our QA pairs are strictly video-centric and maintain high quality, we developed a comprehensive refinement process that addresses the challenges inherent in social media discourse. Social media discussions often contain non-visual information such as personal identifiers, historical references, and specific temporal details that cannot be directly verified through video content alone. To address this challenge, we designed a refinement prompt (\cref{fig:prompts_refine_1} - \ref{fig:prompts_refine_3}) for Claude 3.5 Sonnet Text LLM.

The refinement prompt takes the generic and specific QA pairs generated in previous step as input and applies multiple filtering criteria to ensure video-centricity. The process eliminates references to social media context (\eg `based on replies', `as mentioned in comments') while preserving essential information about entities and events. It standardizes temporal references, converting specific dates and timestamps to general indicators (\eg `morning', `night'). For non-obvious causation, it enforces the use of speculative language (\eg `potential', `likely') while maintaining factual observations for directly visible events. Additionally, the process mandates human-like sentence-form responses and removes precise measurements such as exact speeds or weather metrics that cannot be reliably inferred from video content.

The refinement prompt (\cref{fig:prompts_refine_1} - \ref{fig:prompts_refine_3}) implements specific guidelines for different types of questions and answers. For instance, when describing road events, location information is stripped to focus solely on event characteristics. Entity descriptions maintain specificity when clearly visible (\eg `school bus' vs. generic `vehicle') while avoiding unverifiable details. This structured approach ensures that final QA pairs remain answerable solely through video content while retaining the ability to describe complex road events through observable facts and reasonable inferences. For example, while we remove specific speed measurements like `80 km/h', we retain qualitative assessments like `high speed'. Similarly, instead of stating \texttt{driver was angry}, we describe observable behaviors like \texttt{vehicle swerved across lanes potentially intimidating a cyclist}.

\noindent \textbf{Eliminating potentially harmful and biased tweet content via multi-stage filtering:}
\textit{Prompt-level:} Strict road-event focused prompting and curated template questions within the prompt (Supp, Fig. 8). 
\textit{LLM-level:} Built-in guardrails of LLMs used in our data generation pipeline eliminate harmful and inappropriate content to some extent.
\textit{Post QA generation level:} QA pairs with overly subjective or speculative content are removed. (Supp, Fig. 14-15, Sec: 2.5).
\textit{Human verification level:} Independent annotators review each QA pair against predefined criteria. They reject pairs that violate our guidelines, \eg including unverifiable details (Supp, Fig. 14-15). Final acceptance is determined through majority voting, minimizing individual bias.
\textit{VLM Training:} The measures described above have ensured VLMs trained on our data do not inherit any biases. 
In future, it might be possible to additionally mitigate social media bias by leveraging  world knowledge from external sources.

\noindent \textbf{QA Categorization}: 
We developed a categorization framework for our refined QA pairs to match each question from the refined QA pairs against the same set of 18 template questions that were used in our hybrid QA generation approach (see \cref{sec:hybrid_approach}). These template questions span a range of complexity levels, from basic observational queries (e.g., camera type, weather conditions) to complex analytical questions addressing causation, prevention, and counterfactual scenarios.
 
\begin{figure*}[!t]
    \centering
    \includegraphics[width=0.93\textwidth]{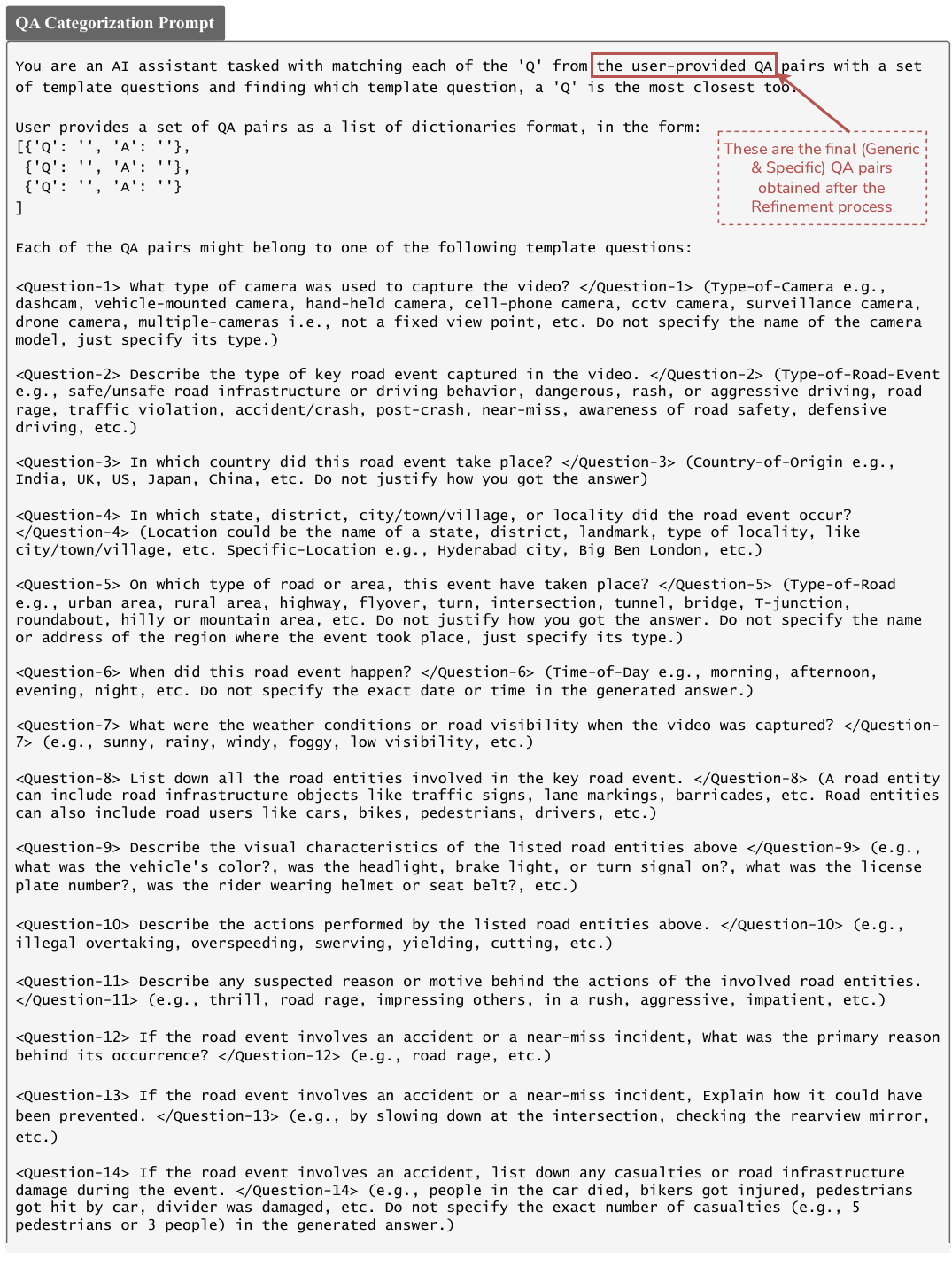}
    \caption*{}  
    \phantomcaption  
    \label{fig:prompts_matching_1}
\end{figure*}

\begin{figure*}[!t]
    \centering
    \includegraphics[width=\textwidth]{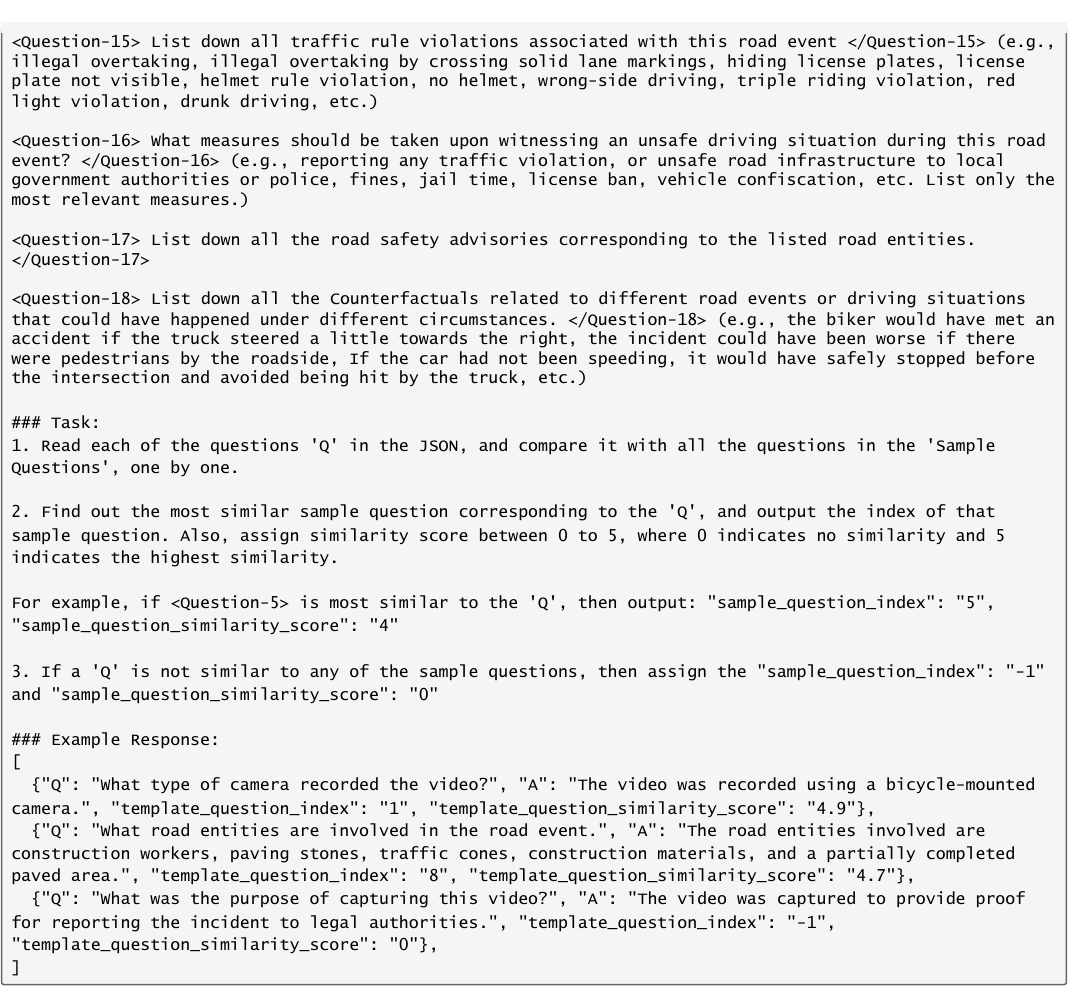}
    \caption{QA categorization prompt design. The prompt (1) matches each refined question against 18 predefined template questions used in QA generation, (2) assigns similarity scores (0-5), and (3) provides examples demonstrating proper template matching for various question types. Refer back to \cref{sec:refinement_and_categorization}
    }
    \label{fig:prompts_matching_2}
\end{figure*}

In this matching process, we prompt Claude 3.5 Sonnet Text LLM to assign a similarity score from 0 to 5 to each QA pair, where 5 indicates perfect alignment with a template question and 0 indicates no meaningful similarity. For example, a question like \texttt{What recording device was used?} would receive a high similarity score with the template \texttt{What type of camera was used to capture the video?}, while a question about video purpose would receive a score of 0 as it doesn't align with any template. 
To ensure categorization quality, QA pairs receiving low similarity scores undergo expert review for potential refinement or removal. This human-in-the-loop validation helps maintain the integrity of our categorization while ensuring comprehensive coverage across all aspects of road event analysis. The complete categorization process and scoring mechanism is implemented through carefully designed prompt (\raisebox{-3.6pt}{\includegraphics[width=0.025\textwidth, height=0.025\textwidth]{images/refine_and_categorize_prompt_icon.png}}) shown in \cref{fig:prompts_matching_1} - \ref{fig:prompts_matching_2}.

\subsection{Incompatible QA Generation}
\label{sec:incompatible_qa_gen}

To evaluate Video LLMs' resilience to hallucination and their ability to discriminate between road and non-road events, we developed an approach for generating incompatible QA pairs. The process involves a three steps: video classification, summarization, and QA generation.

\begin{figure*}[!t]
    \centering
    \includegraphics[width=\textwidth]{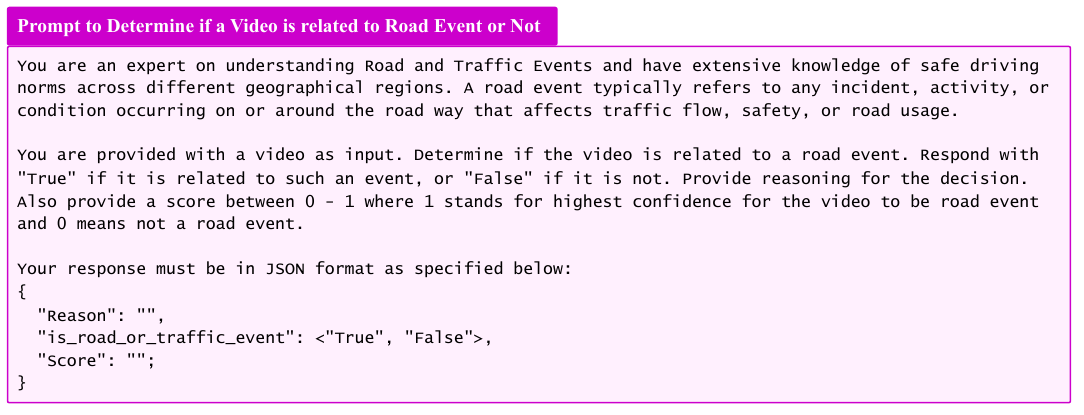}
    \caption{Prompt design for road event classification. The prompt implements binary classification (road/non-road) with confidence scoring (0-1) and reasoning requirements for video content. Refer back to \cref{sec:incompatible_qa_gen}.}
    \label{fig:prompts_incompatible_1}
\end{figure*}

\noindent \textit{Identifying non-road event videos}: First, we employ a specialized prompt (\raisebox{-3.6pt}{\includegraphics[width=0.025\textwidth, height=0.025\textwidth]{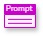}}) to classify videos as road or non-road events. The prompt (\cref{fig:prompts_incompatible_1}) defines road events as `any incident, activity, or condition occurring on or around the roadway that affects traffic flow, safety, or road usage' and assigns a confidence score between 0 and 1, providing detailed reasoning for the classification.

\begin{figure*}[!t]
    \centering
    \includegraphics[width=\textwidth]{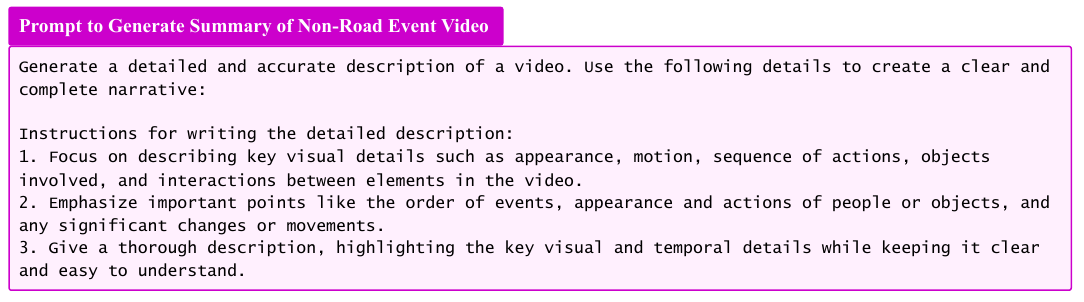}
    \caption{Video summarization prompt for non-road events. The prompt ensures structured description of visual content focusing on key details, temporal sequences, and object interactions. Refer back to \cref{sec:incompatible_qa_gen}.}
    \label{fig:prompts_incompatible_2}
\end{figure*}

\noindent \textit{Non-road event video summarization}: Videos confirmed as non-road events then undergo detailed visual summarization using a structured prompt (\cref{fig:prompts_incompatible_2}) that captures key visual details, temporal sequences, and object interactions.

\noindent \textit{Generating mismatched QA pairs}: For generating incompatible QA pairs, we sample questions generated for our road event videos and apply them to these confirmed non-road videos. We modify our hybrid QA generation approach by incorporating additional prompting constraints. The modified prompt explicitly acknowledges the video's non-road nature and requires the model to articulate the incompatibility between road-event questions and the video content based on the non-road even summary generated in the previous step. This approach generates responses that highlight the fundamental mismatch between the question's assumptions and the video's actual content.

\begin{figure*}[!t]
    \centering
    \includegraphics[width=\textwidth]{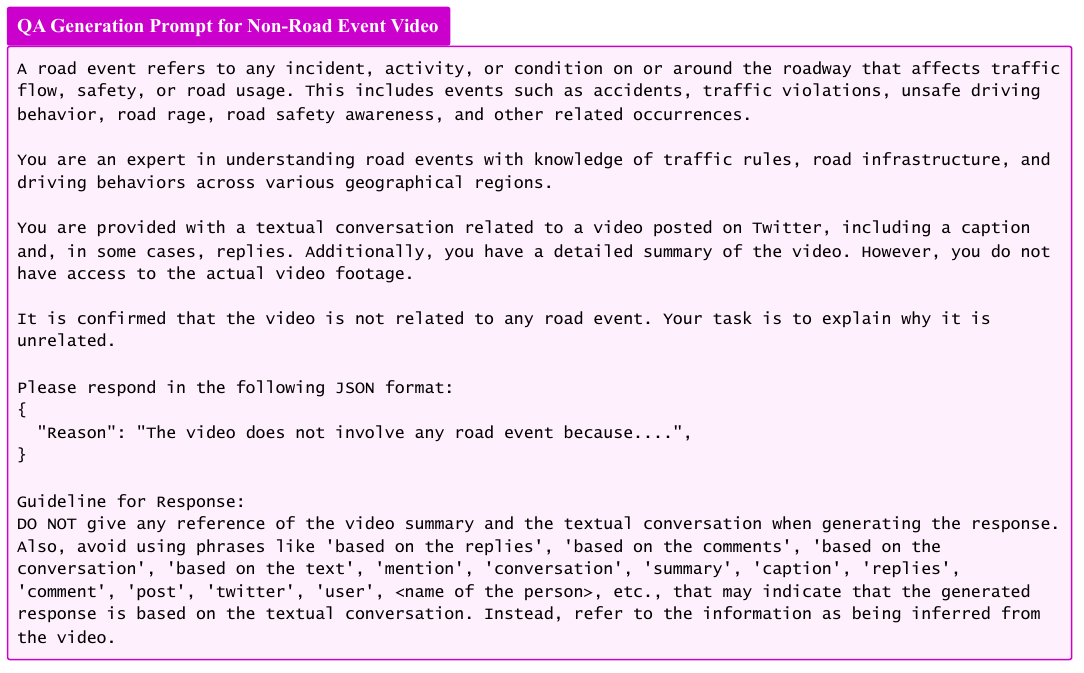}
    \caption{Incompatible QA generation prompt. The prompt generates explanations for why road event questions are incompatible with non-road video content while maintaining established response formats. Refer back to \cref{sec:incompatible_qa_gen}.}
    \label{fig:prompts_incompatible_3}
\end{figure*}

This methodology serves multiple purposes: testing QA generation pipeline robustness, evaluating Video LLMs' ability to recognize and reject inappropriate questions, and generating training data for improving model discrimination. \cref{fig:prompts_incompatible_1}, \cref{fig:prompts_incompatible_2} and \cref{fig:prompts_incompatible_3} illustrates the prompts utilized for generating Incompatible QA pairs for non-road event videos.

\subsection{Adversarial QA Generation}
\label{sec:adversarial_qa_gen}

To evaluate Video LLMs' ability to recognize and reject misleading assumptions, we developed an approach for generating adversarial QA pairs. These QA pairs specifically test models' capabilities in identifying non-occurring road events and avoiding hallucination by introducing questions about events, objects, or actions that are not present in the video.

The generation process employs Claude 3.5 Sonnet Text LLM to analyze the generated QA pairs (from \cref{sec:refinement_and_categorization}) associated with a video and create new questions that maintain the road safety context while introducing irrelevant elements. For example, given a video showing a simple traffic violation, an adversarial question might ask about non-existent casualties or emergency responses. The answers are carefully crafted to explicitly state the absence of these elements while maintaining a video-centric perspective (\eg \texttt{The video shows no emergency vehicles or medical response as there were no casualties in this traffic violation incident}).

\begin{figure*}[!t]
    \centering
    \includegraphics[width=\textwidth]{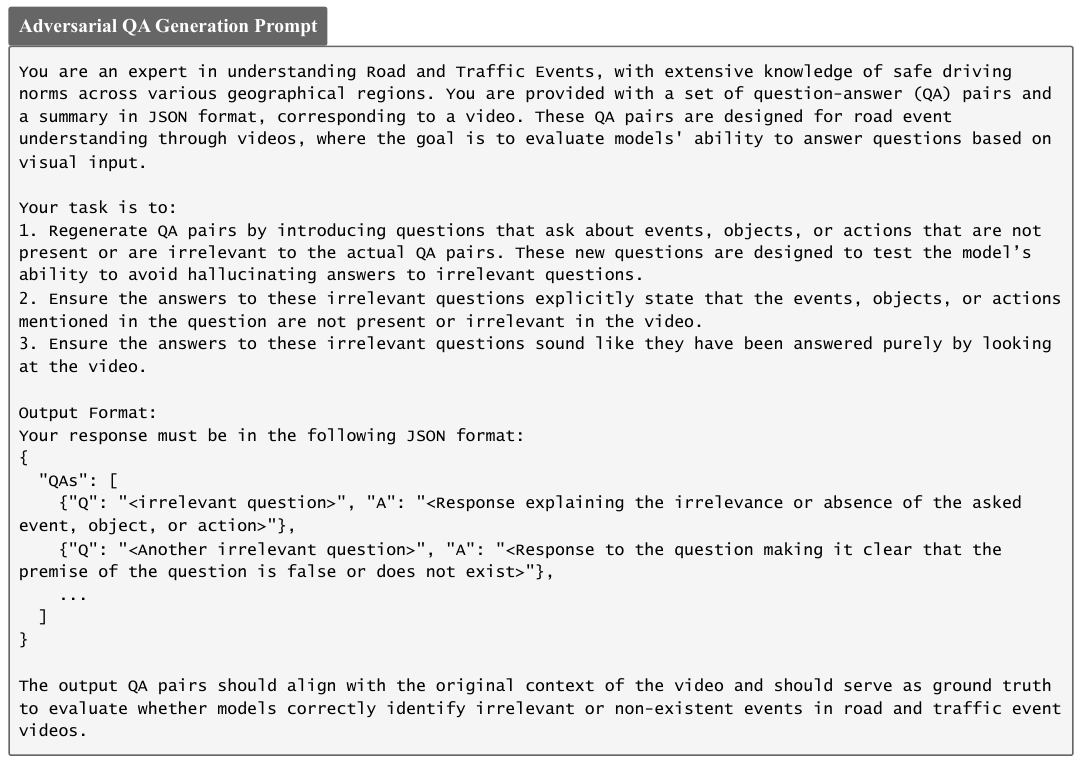}
    \caption{Adversarial QA generation prompt. The prompt instructs the generation of questions about non-occurring events while maintaining road context, with examples demonstrating (1) proper introduction of irrelevant elements, (2) explicit negation in answers, and (3) preservation of video-centric response format. Refer back to \cref{sec:adversarial_qa_gen}.}
    \label{fig:adversarial_prompt}
\end{figure*}

This approach differs from Incompatible QA generation as it maintains the road event context while testing for fine-grained discrimination. While Incompatible QAs evaluate model robustness on completely unrelated videos, Adversarial QAs test the model's ability to reject false premises within relevant road scenarios. Representative examples of adversarial QA pairs are shown in \cref{fig:template_id_mapping_fig3}, \ref{fig:task_qa_examples_advertisement_blindspot} and \ref{fig:task_qa_examples_handheld_hydroplaning}. \cref{fig:adversarial_prompt} demonstrates prompt (\raisebox{-3.6pt}{\includegraphics[width=0.025\textwidth, height=0.025\textwidth]{images/refine_and_categorize_prompt_icon.png}}) to generate adversarial QA pairs from existing QA pairs.

\subsection{Temporal Grounding QA generation} 
The answers in ``Description QA", auto generated by our annotation pipeline, provides details of key video events (\eg `car hitting a biker'). Annotators manually mark the temporal extent (start, end) of these key events.   

\subsection{Final QA Task Taxonomy}
\label{sec:qa_taxonomy}

\begin{figure*}[!t]
    \centering
    \includegraphics[width=\textwidth]{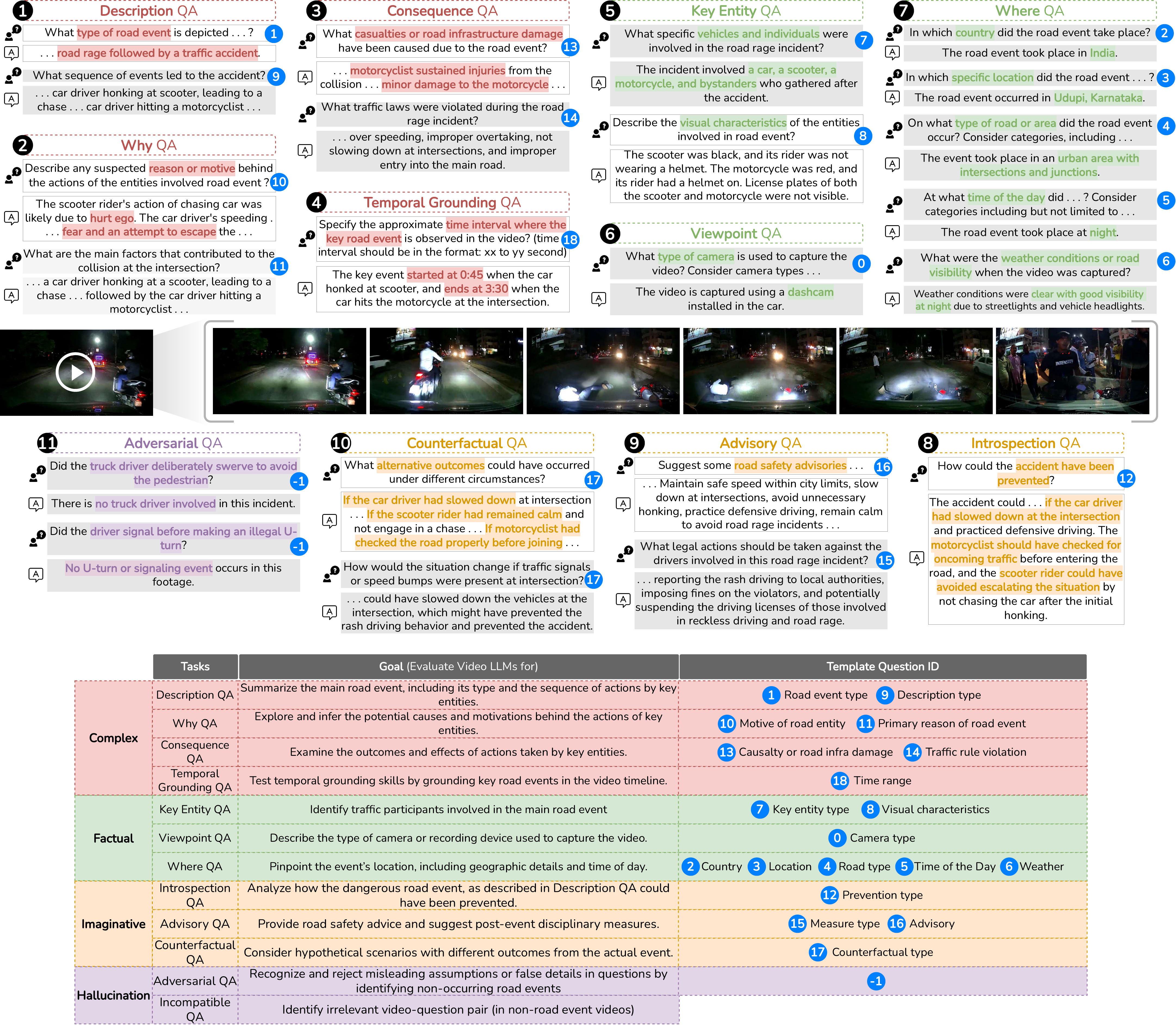}
    \caption{\textbf{QA Task Taxonomy and Template Question ID Mapping for Video LLM Evaluation}: The taxonomy consists of 12 QA tasks organized into four reasoning categories: Complex (red), Factual (green), Imaginative (orange), and Hallucination (purple). The 19 template question IDs (blue circles) map to QA tasks designed for evaluating road event understanding. For Incompatible QAs, which evaluate model robustness on non-road event videos, we employ a separate generation pipeline (\cref{sec:incompatible_qa_gen}) without template ID mapping. Refer back to \cref{sec:qa_taxonomy}.}
    \label{fig:template_id_mapping_fig3}
\end{figure*}

\begin{figure*}[!t]
    \centering
    \includegraphics[width=\textwidth]{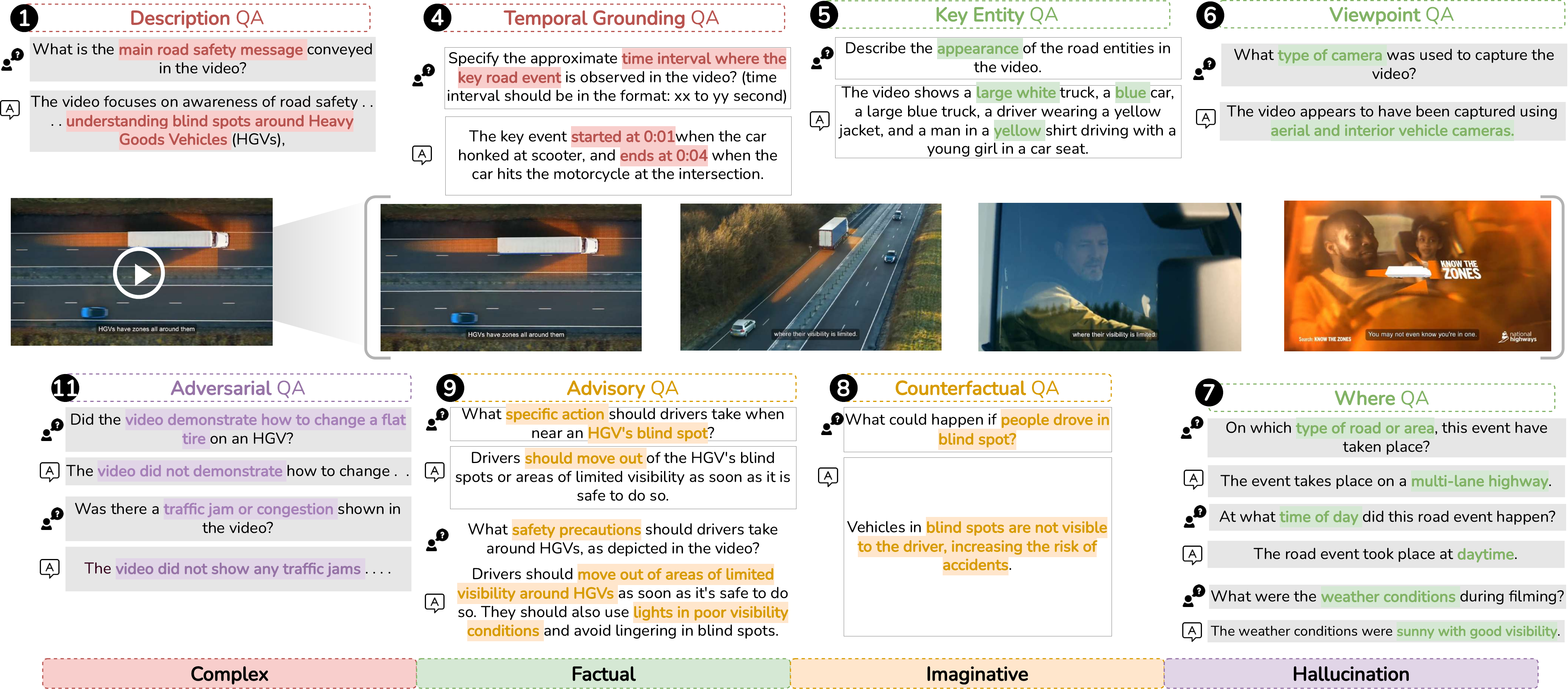}
    \caption{Examples of QA Pairs grouped by tasks and color-coded by task category (for an advertisement video captured via multiple viewpoints). Gray fill shading indicates specific questions while the non-shaded QAs are generic. Highlighted text indicates key information. Refer back to \cref{sec:qa_taxonomy}.}
    \label{fig:task_qa_examples_advertisement_blindspot}
\end{figure*}

\begin{figure*}[!t]
    \centering
    \includegraphics[width=\textwidth]{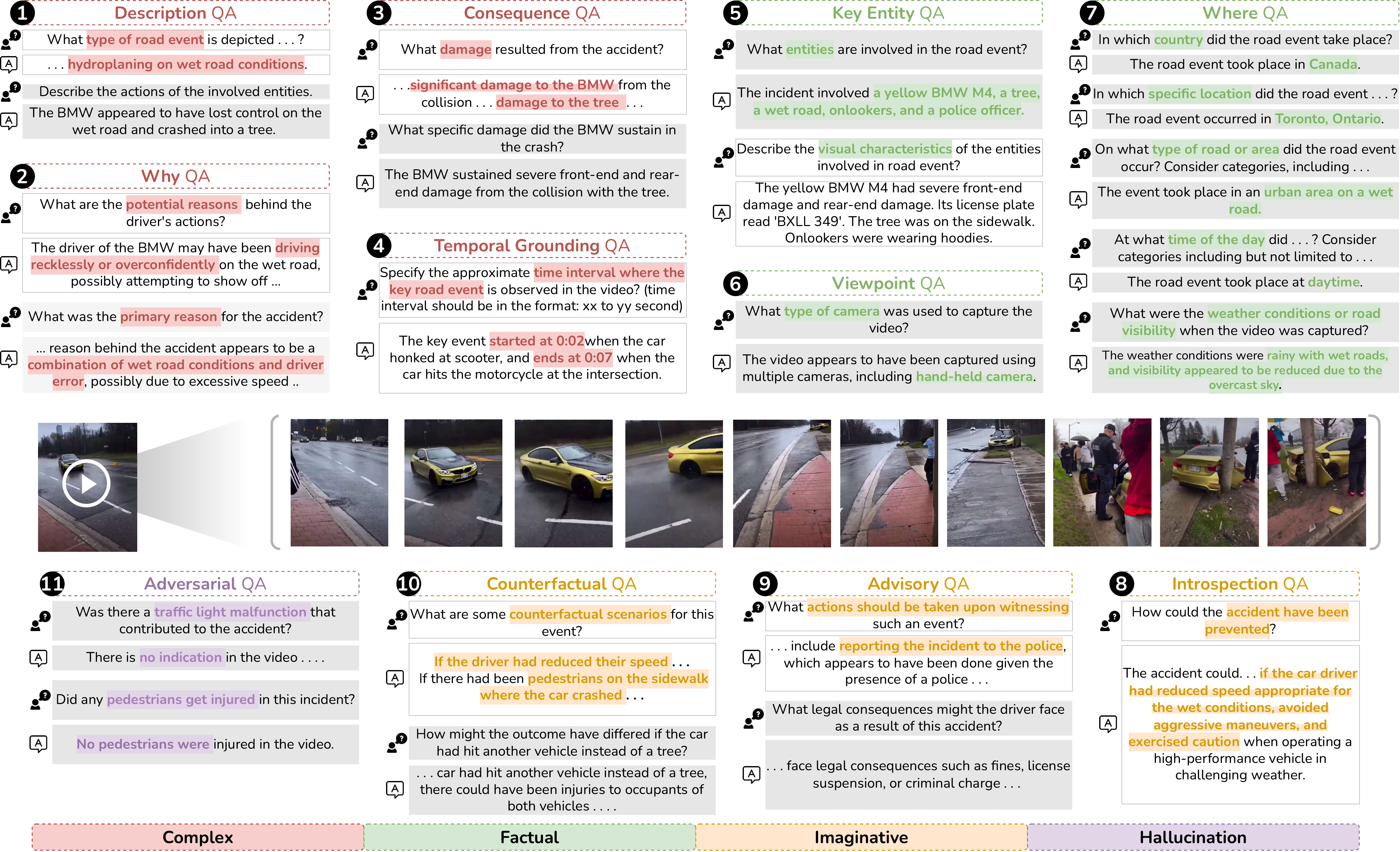}
    \caption{Examples of QA Pairs grouped by tasks and color-coded by task category (for an hydroplaning incident captured via handheld camera). Gray fill shading indicates specific questions while the non-shaded QAs are generic. Highlighted text indicates key information. Refer back to \cref{sec:qa_taxonomy}.}
    \label{fig:task_qa_examples_handheld_hydroplaning}
\end{figure*}

We developed a structured taxonomy to evaluate Video Large Language Models' (Video LLMs) capabilities across different reasoning categories. Our taxonomy organizes question-answer (QA) pairs into four primary reasoning categories: Complex, Factual, Imaginative and Hallucination reasoning. These categories are further divided into 12 distinct QA tasks designed to assess specific aspects of video understanding.
Through our QA generation pipeline (discussed till now), each QA pair is assigned one of 19 template question IDs. We then map these template IDs to specific QA tasks within our taxonomy, as illustrated in \cref{fig:template_id_mapping_fig3}. This systematic mapping enables structured evaluation of Video LLMs across different reasoning capabilities while ensuring comprehensive coverage of our taxonomy's tasks.

\begin{figure*}[!t]
    \centering
    \includegraphics[width=\textwidth]{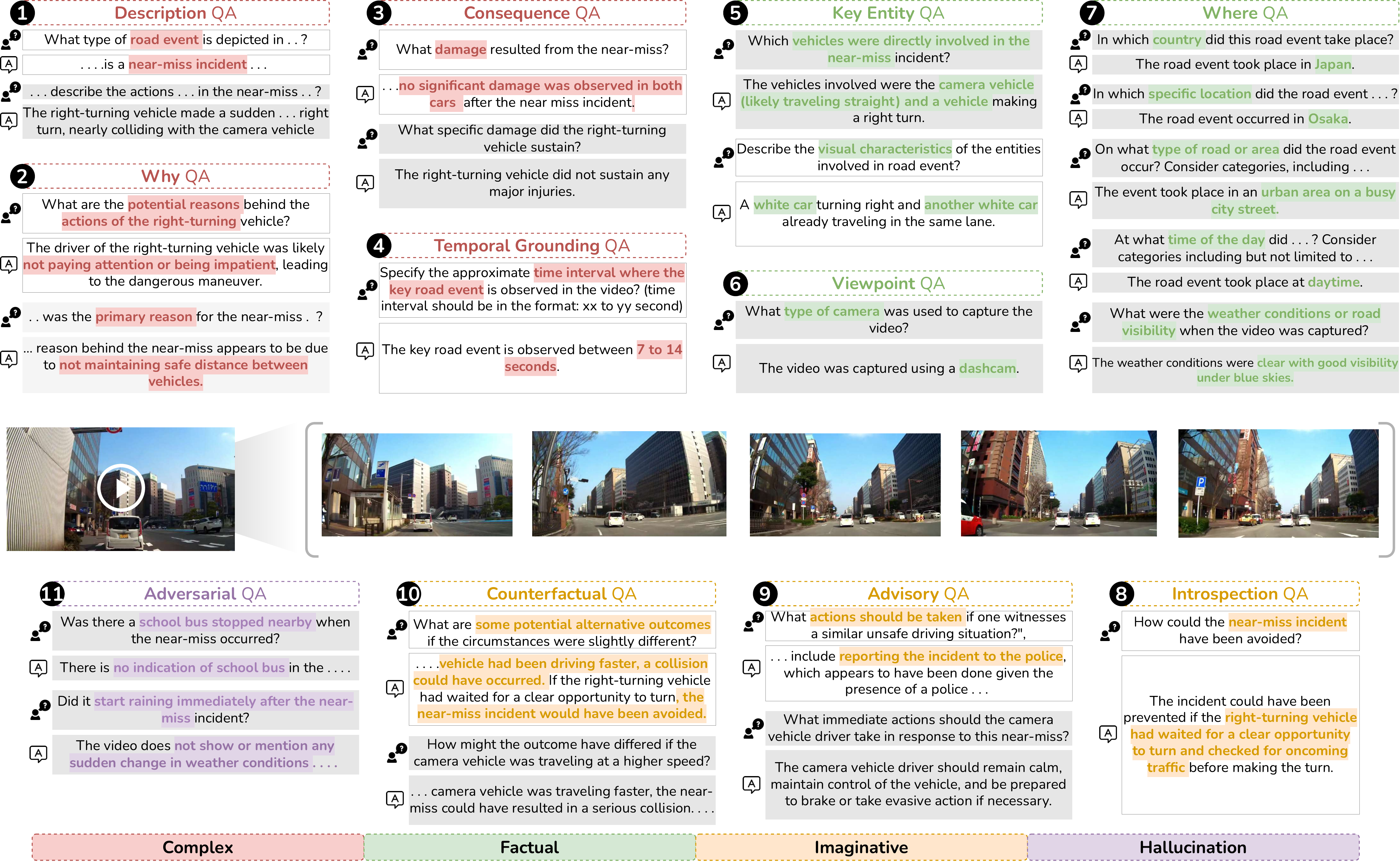}
    \caption{Examples of QA Pairs grouped by tasks and color-coded by task category (for a near-miss incident captured in Japan). Gray fill shading indicates specific questions while the non-shaded QAs are generic. Highlighted text indicates key information. Refer back to \cref{sec:qa_taxonomy}.}
    \label{fig:task_qa_examples_japanese_nearmiss}
\end{figure*}

\begin{figure*}[!t]
    \centering
    \includegraphics[width=\textwidth]{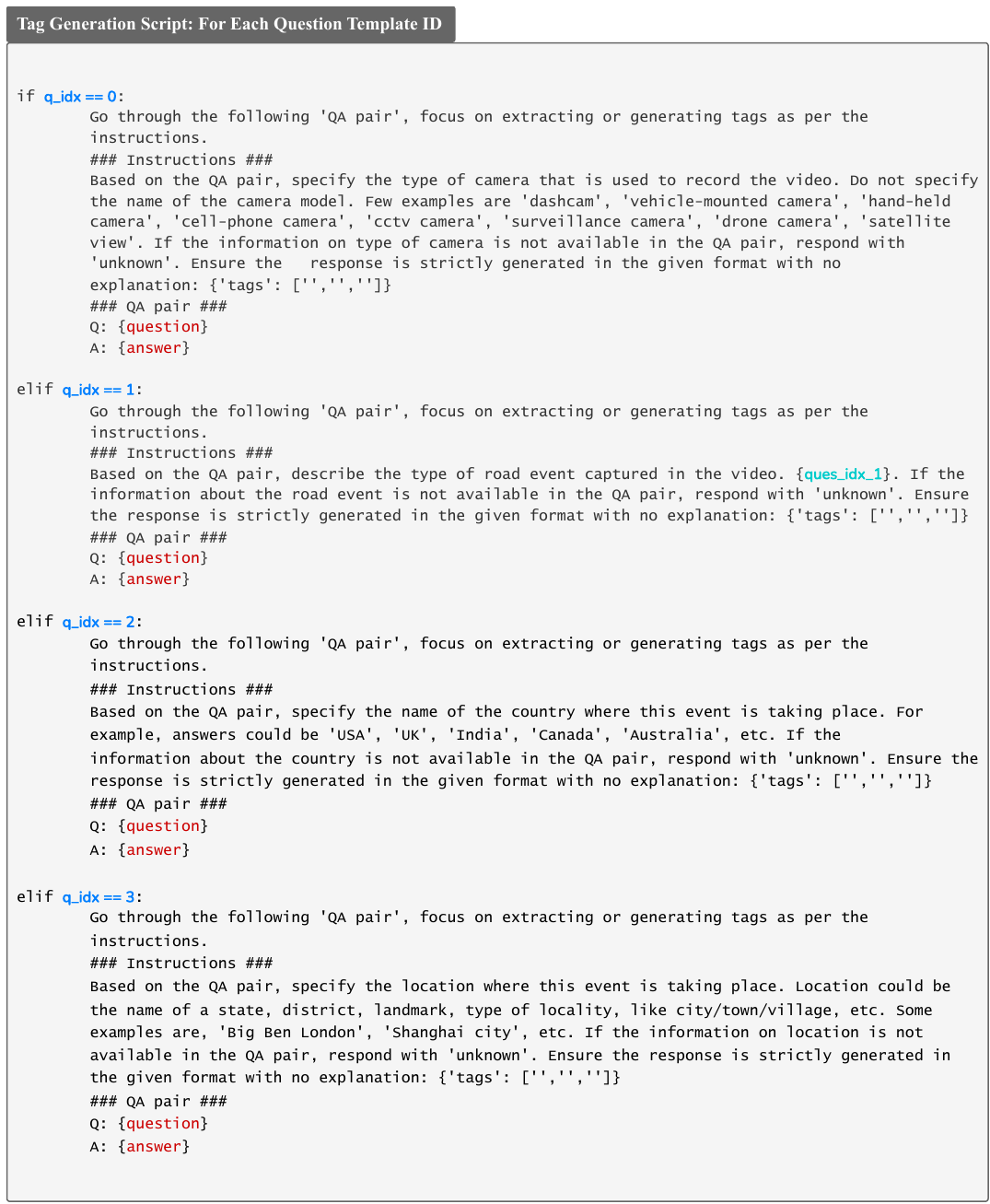}
    \caption*{}  
    \phantomcaption  
    \label{fig:tag_gen_prompts_1}
\end{figure*}

\begin{figure*}[!t]
    \centering
    \includegraphics[width=\textwidth]{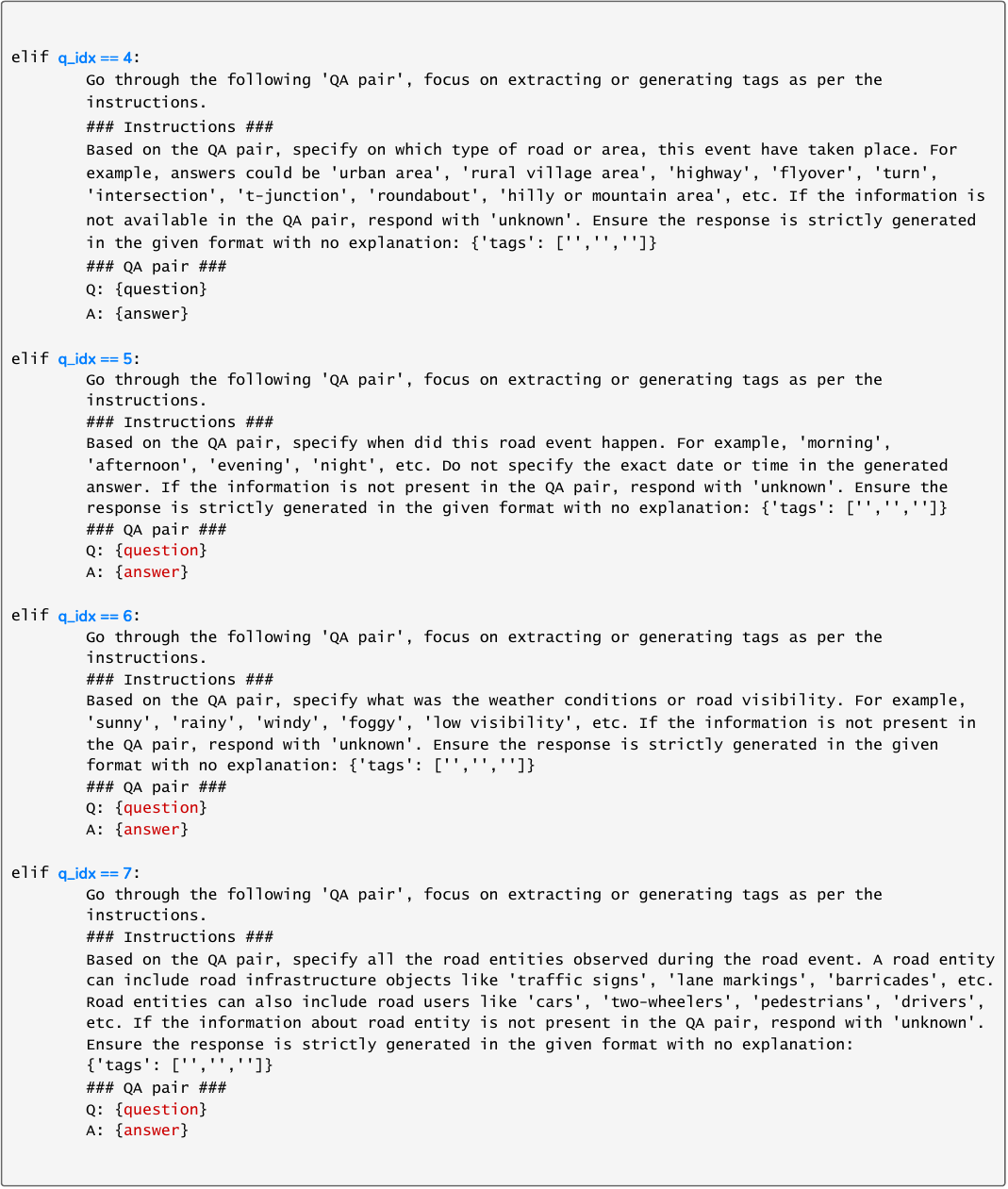}
    \label{tag_gen_prompts_2}
\end{figure*}

\begin{figure*}[!t]
    \centering
    \includegraphics[width=\textwidth]{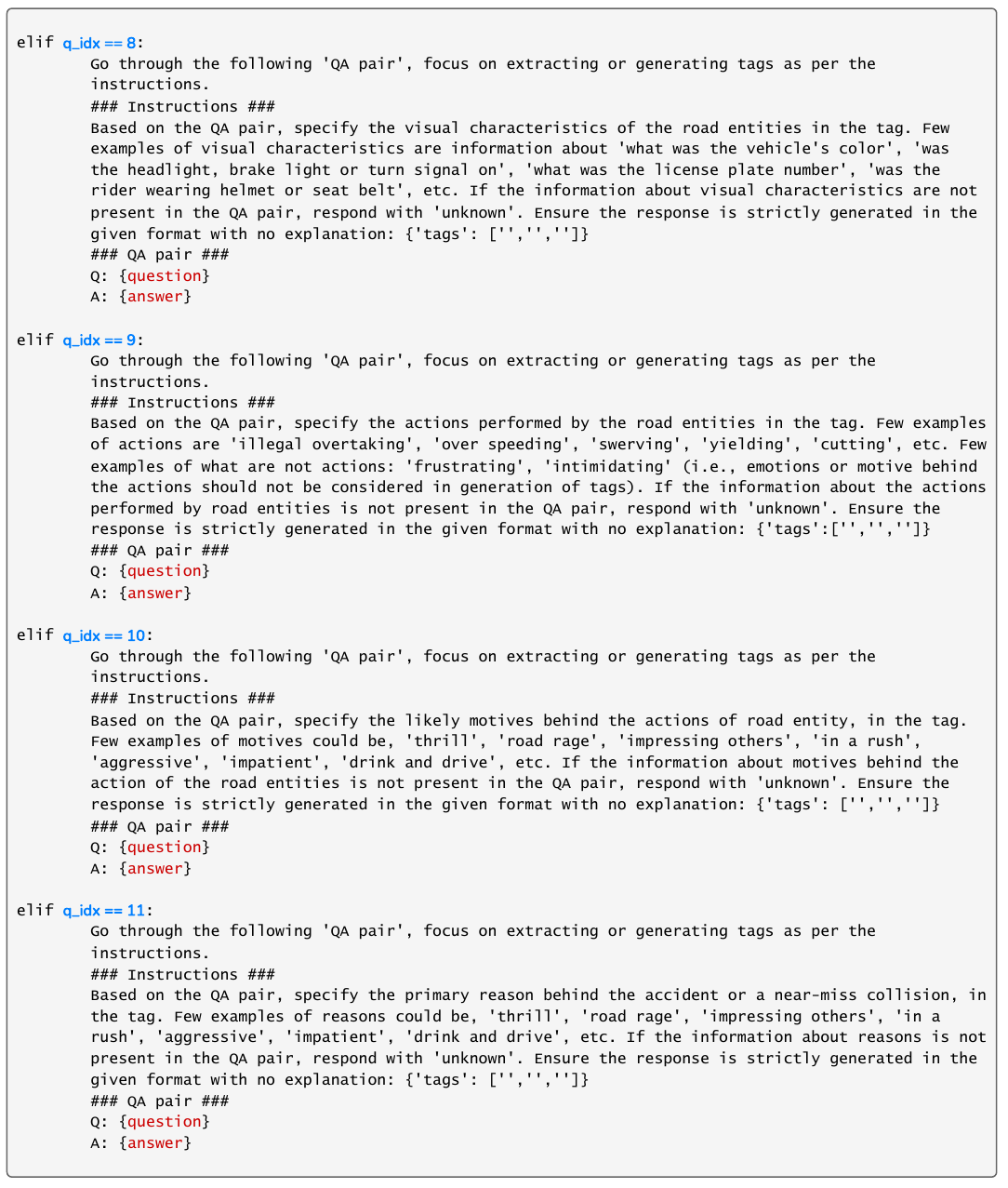}
    \label{tag_gen_prompts_3}
\end{figure*}

\begin{figure*}[!t]
    \centering
    \includegraphics[width=\textwidth]{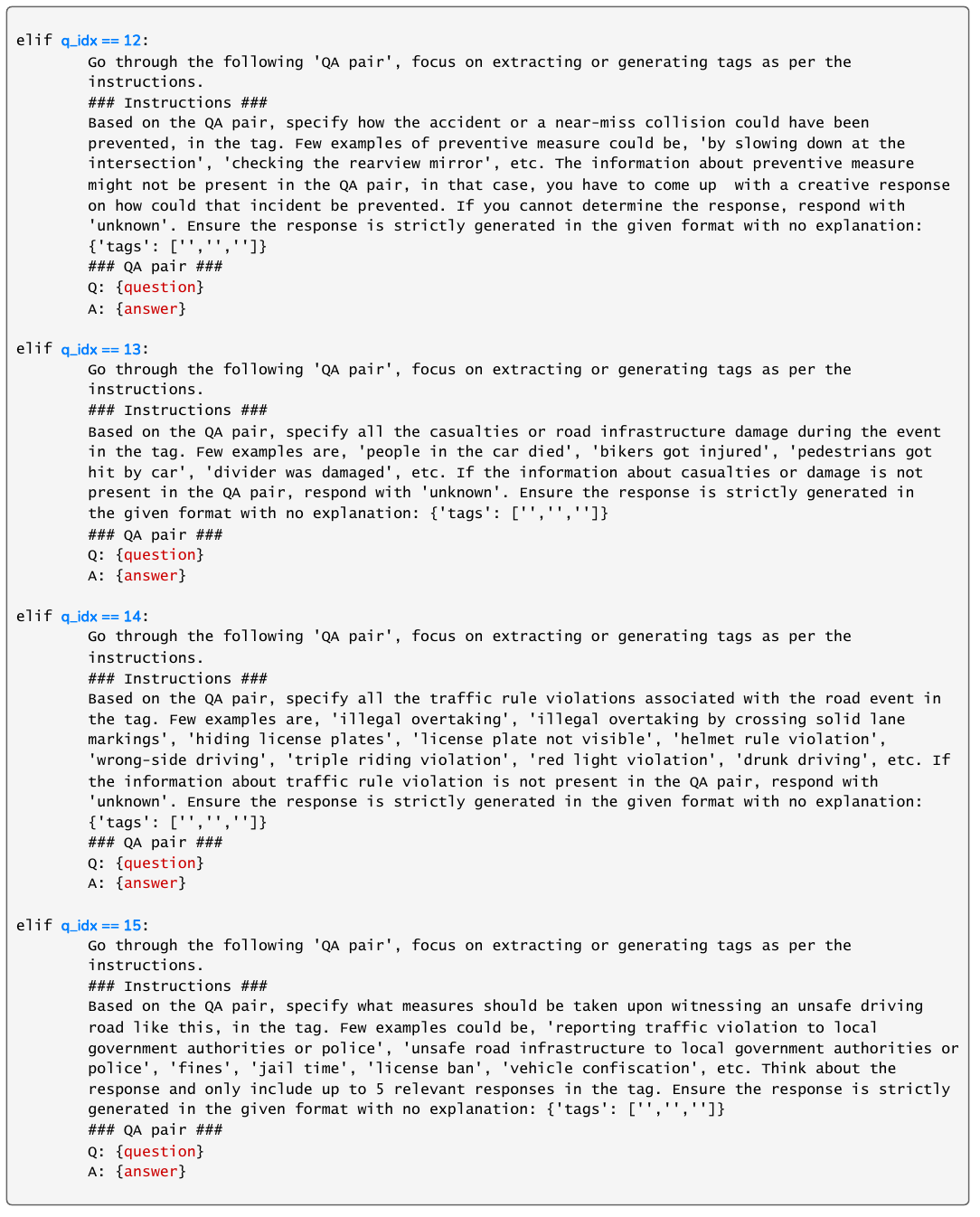}
    \label{tag_gen_prompts_4}
\end{figure*}

\begin{figure*}[!t]
    \centering
    \includegraphics[width=\textwidth]{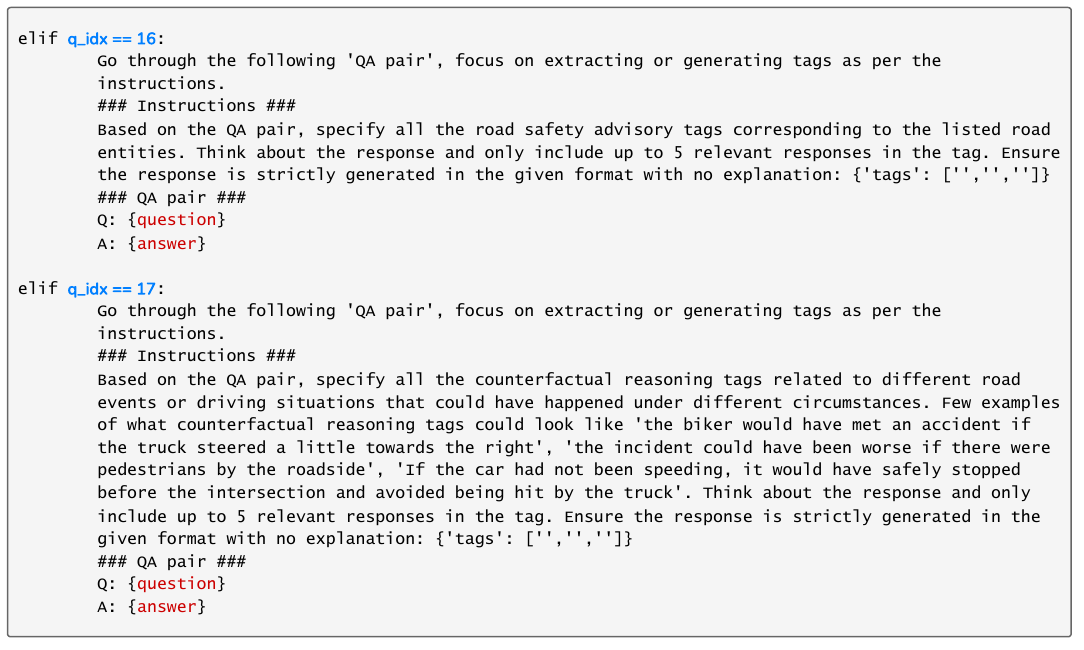}
    \caption{Tag extraction prompt design for different template question IDs. The prompt employs conditional logic based on question IDs to generate appropriate tags: camera type (q\_idx=0), road event type (q\_idx=1), country (q\_idx=2), and specific location (q\_idx=3). Each condition includes specific instructions and examples for tag generation, ensuring standardized output format {'tags': ['','','']}. Note: \textbf{\textcolor{cyan}{ques\_idx\_1}} is a command providing tag generation instructions for q\_id=1. This command can be found in \cref{fig:tag_gen_prompts_1_part}. Refer back to \cref{sec:video_tag_gen}.}
    \label{fig:tag_gen_prompts_5}
\end{figure*}

\begin{figure*}[!t]
    \centering
    \includegraphics[width=0.93\textwidth]{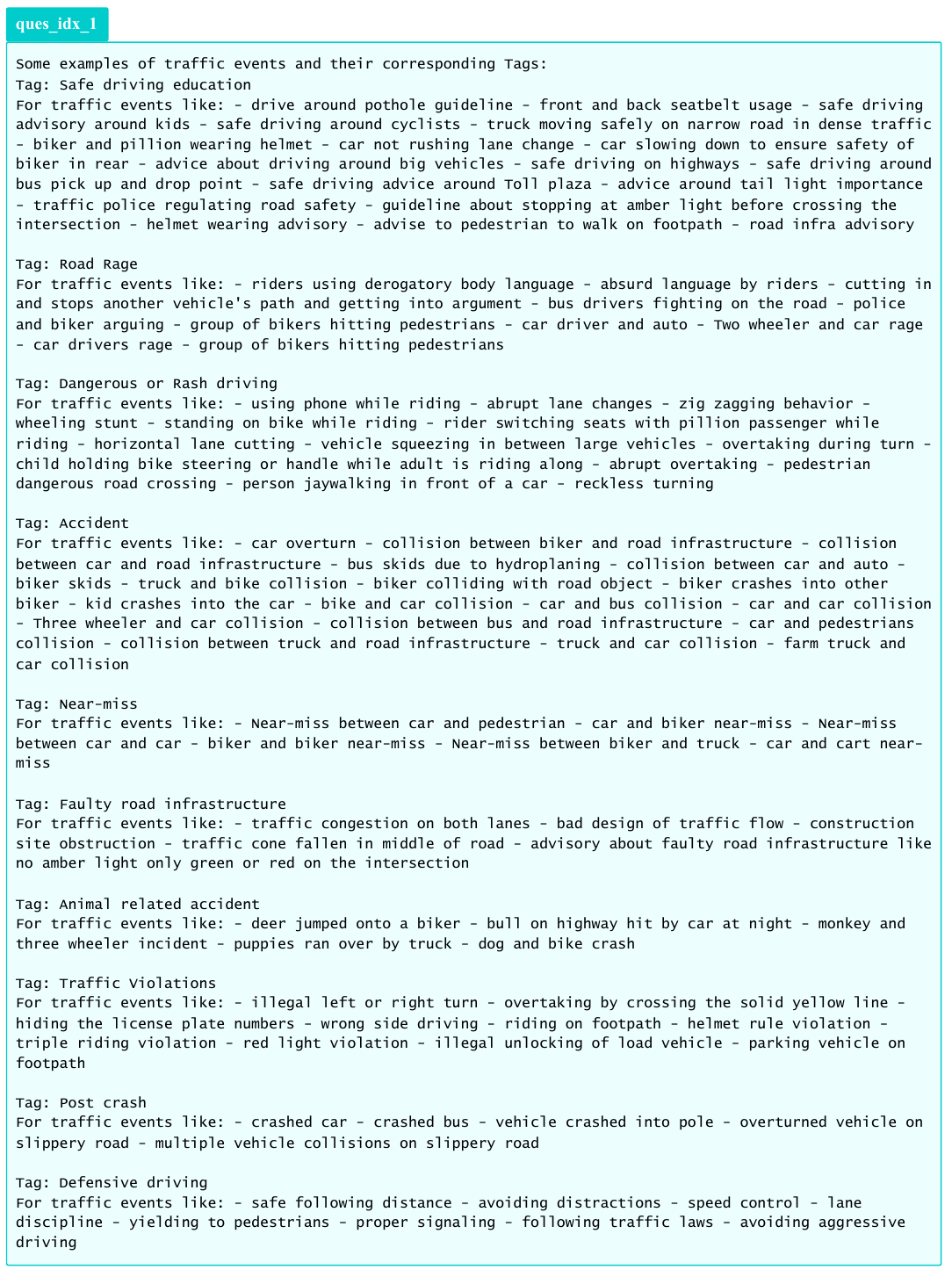}
    \caption{Refer to \cref{fig:tag_gen_prompts_5} for details.}
    \label{fig:tag_gen_prompts_1_part}
\end{figure*}



\begin{figure*}[!t]
    \centering
    \includegraphics[width=\textwidth]{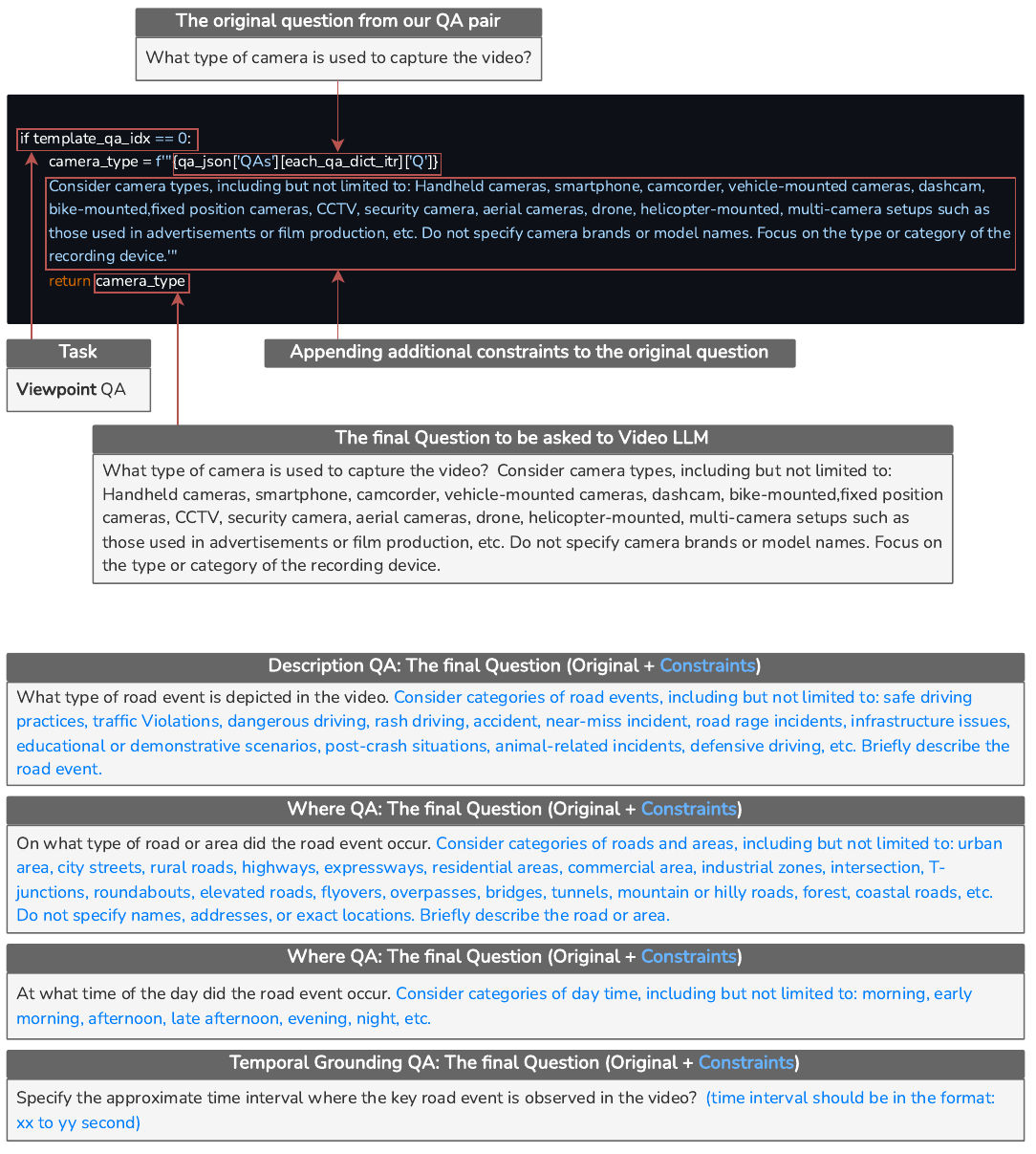}
    \caption{The diagram shows an example of task-specific prompt utilized for the evaluation of Video LLMs. The code snippet at the top demonstrates how this is done. First, for a specific question, we find its QA type via its template ID, then for that template ID, if we have a predefined constraint, we append that to the original question. Original question $+$ constraint examples for Description QA, Why QA and Temporal Grounding QA tasks is shown. Rest of the tasks have only original questions and no predefined constraints. Refer back to \cref{sec:data_setup}.}
    \label{fig:prompt_strategy}
\end{figure*}

\begin{figure*}[!t]
    \centering
    \includegraphics[width=\textwidth]{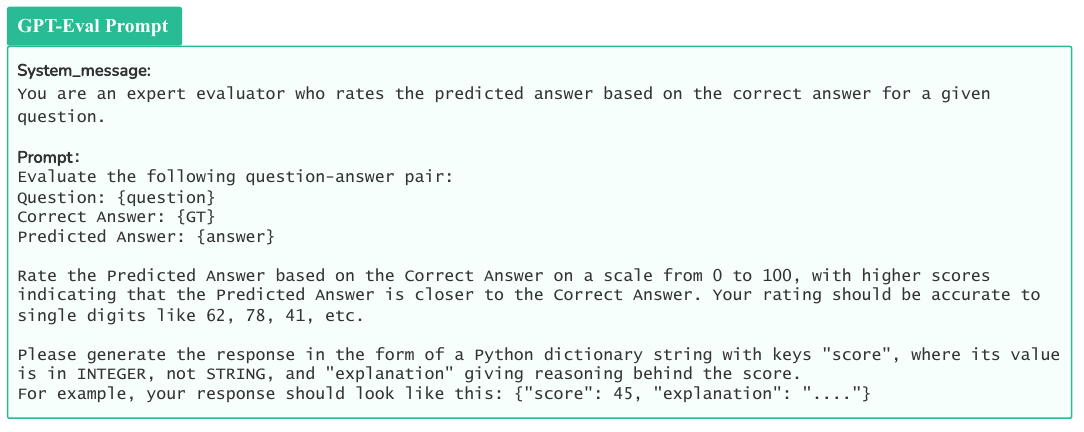}
    \caption{Evaluation prompt for assessing model-generated answers. The prompt implements (1) structured comparison between predicted and ground-truth answers, (2) fine-grained scoring on a 0-100 scale, and (3) requirement for explanatory justification. The output format ensures programmatic processing while maintaining evaluation transparency. Refer back to \cref{sec:model_setup_supp}.}
    \label{fig:prompts-GPT-Eval}
\end{figure*}

\begin{figure*}[!ht]
    \centering
    \includegraphics[width=\textwidth]{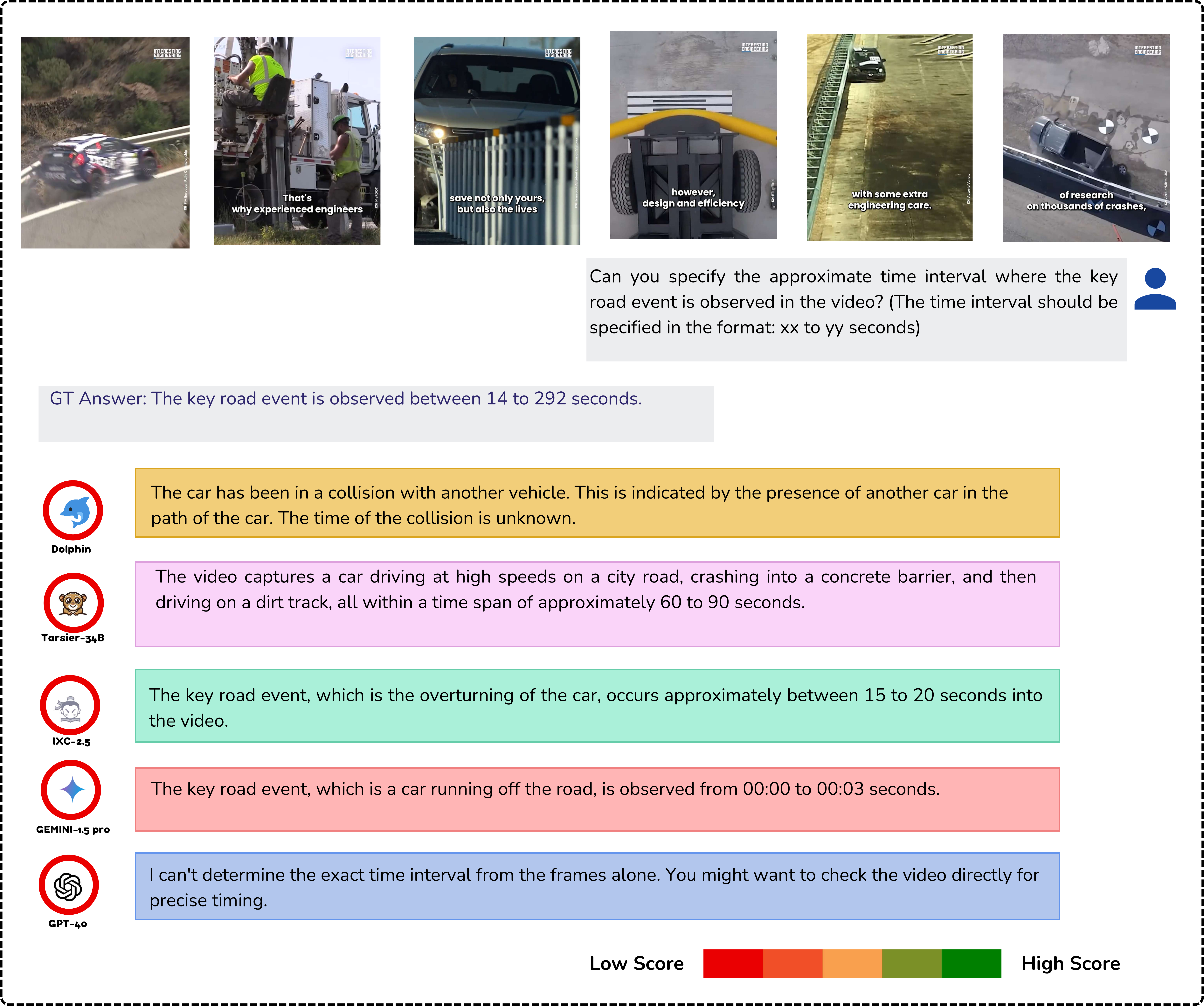}
    \caption{\textbf{Model performance comparison on Temporal Grounding task}: Top: Frames from a video showing a car accident sequence. Middle: Models are asked to specify the temporal interval of the key road event. Ground truth (in gray) indicates the event spans 14-292 seconds. Bottom: Model responses (colored boxes) demonstrate varying approaches: while some attempt to provide specific intervals (e.g., 15-20 seconds, 0-3 seconds), others offer vague temporal descriptions. Red circles around model icons indicate that despite different response styles, all models fail to accurately identify the correct time interval. This example illustrates the significant challenge Video LLMs face in precise temporal localization of road events. Refer back to \cref{sec:qualitative}.}
    \label{fig:QA_time_1}
\end{figure*}

\begin{figure*}[!ht]
    \centering
    \includegraphics[width=\textwidth]{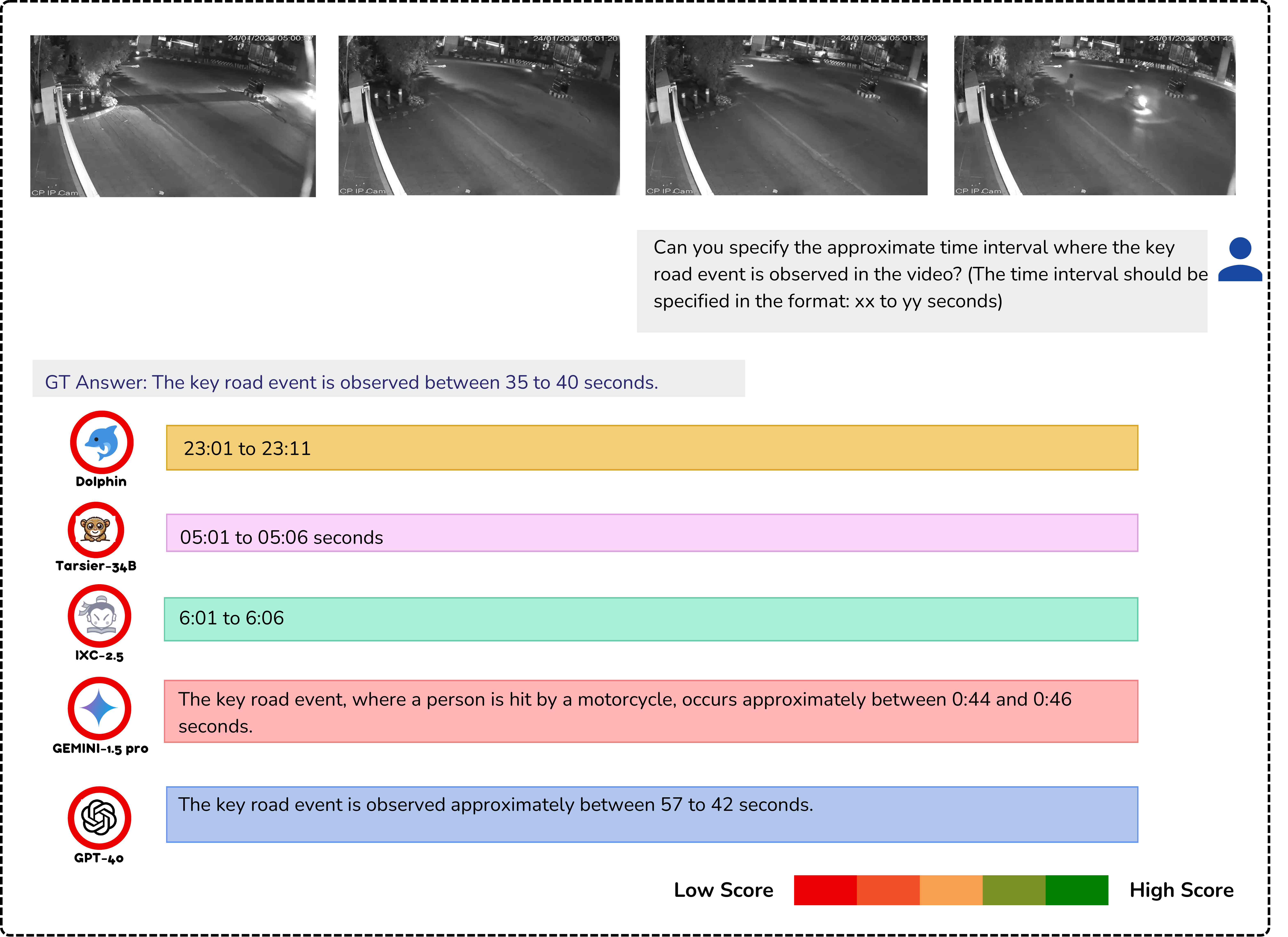}
    \caption{\textbf{Model performance comparison on Temporal Grounding task}: Top: Sequential frames from a CCTV video showing a nighttime road scene. Middle: Models are asked to specify the temporal interval of the key road event, with ground truth spanning 35-40 seconds (gray box). Bottom: Model responses (colored boxes) demonstrate varying approaches: most provide specific time intervals (e.g., 23:01-23:11, 05:01-05:06) while Gemini additionally describes the event type ('person hit by motorcycle'). Red circles around model icons indicate that despite different response styles, all models fail to provide the correct interval. GPT-4o's response (57 to 42 seconds) even shows incorrect temporal ordering. This example highlights Video LLMs' consistent difficulty with precise temporal localization. Refer back to \cref{sec:qualitative}.}
    \label{fig:QA_time_2}
\end{figure*}

\begin{figure*}[!ht]
    \centering
    \includegraphics[width=\textwidth]{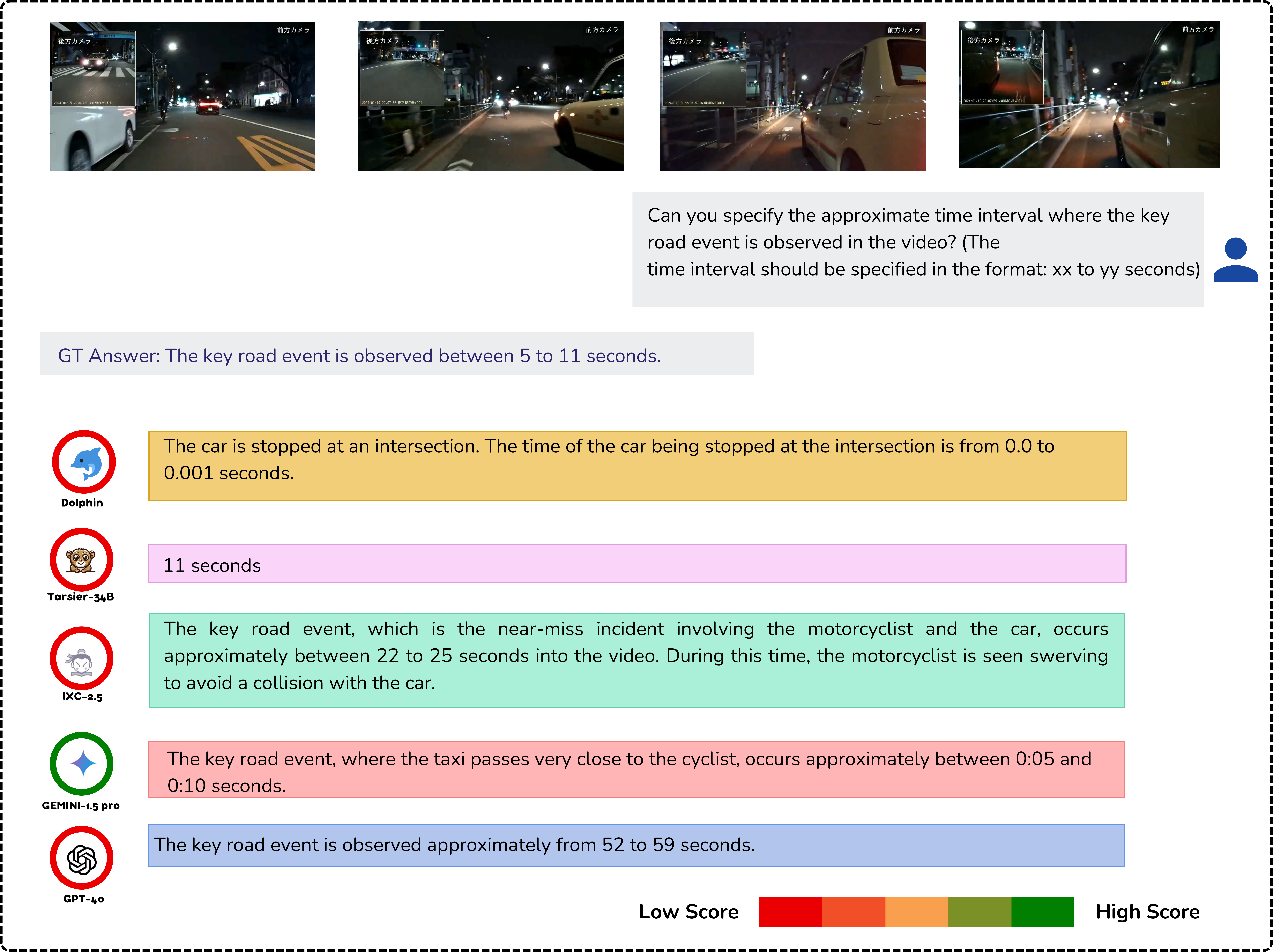}
    \caption{\textbf{Model performance comparison on Temporal Grounding task}: Top: Dashcam footage showing a nighttime near-miss incident between a taxi and cyclist. Middle: Models are asked to specify the temporal interval of the key road event, with ground truth spanning 5-11 seconds (gray box). Bottom: Model responses (colored boxes) show diverse approaches: while Gemini-1.5 Pro (green circle) correctly identifies both the event type and provides a reasonable time estimate (0:05-0:10), other models either give incorrect intervals (IXC: 22-25s, GPT-4o: 52-59s), overly precise timing (Dolphin: 0.0-0.001s), or incomplete responses (Tarsier: '11 seconds'). This example demonstrates that even when models accurately describe the event (taxi passing close to cyclist), precise temporal localization remains challenging, with only one model achieving high accuracy. Refer back to \cref{sec:qualitative}.}
    \label{fig:QA_time_3}
\end{figure*}


\begin{figure*}[!ht]
    \centering
    \includegraphics[width=0.95\textwidth]{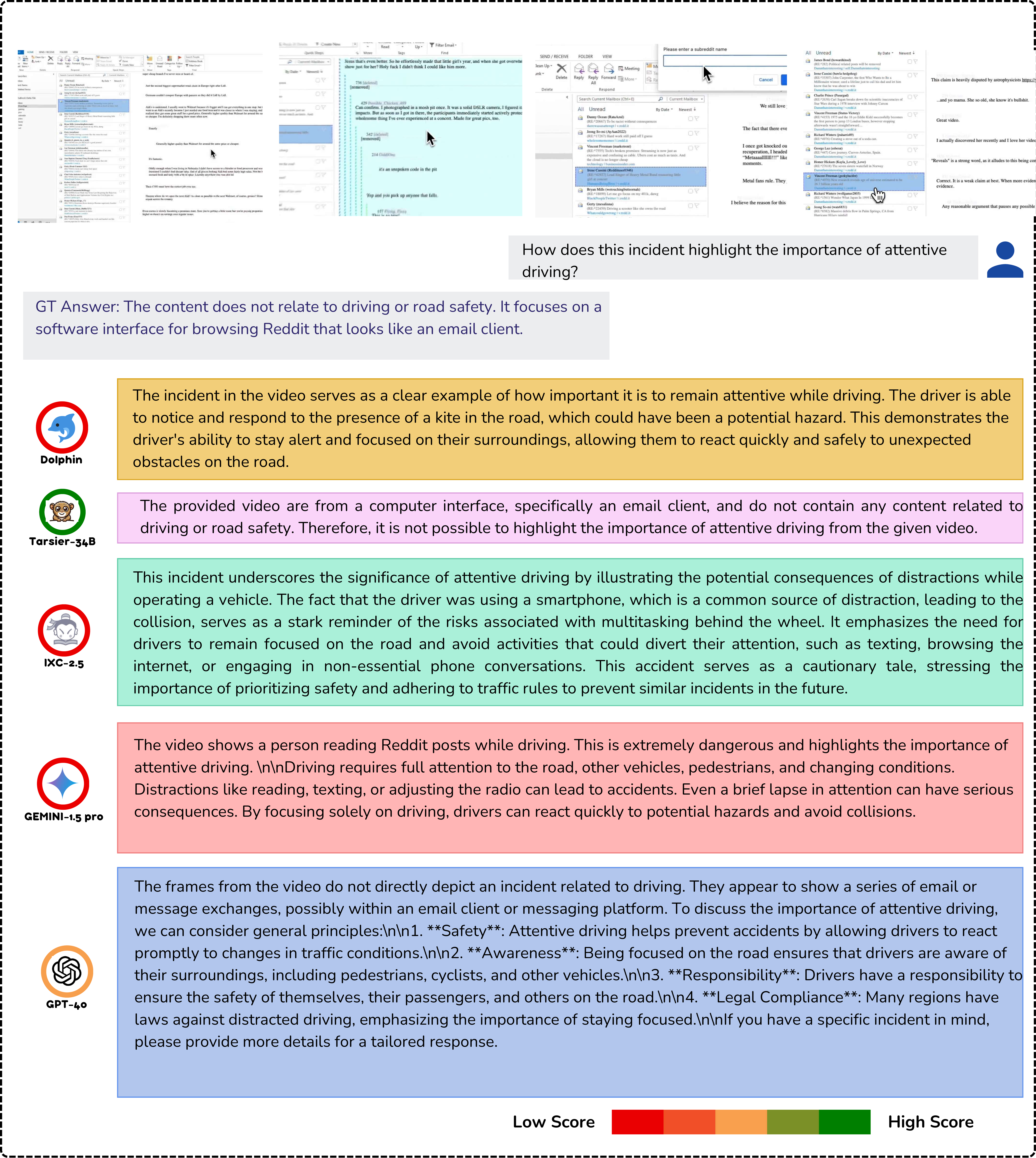}
    \caption{\textbf{Model performance comparison on Incompatible QA task}: Top: Video of a Reddit interface (non-road-event content). Middle: Models are asked about attentive driving implications, while the ground truth (gray box) correctly mentions that the content is unrelated to driving safety. Bottom: Model responses showcase varying levels of hallucination: most models (red circles) fabricate driving scenarios and safety implications despite the irrelevant content, while Tarsier (green) correctly identifies that the video is not related to road event. Although, GPT-4o (orange) correctly identifies the computer interface, it still attempts to relate it to driving. This example highlights a critical challenge in Video LLM robustness - the tendency to hallucinate road safety contexts even when presented with completely unrelated visual content. Refer back to \cref{sec:qualitative}.}
    \label{fig:QA_incompatible_2}
\end{figure*}

\begin{figure*}[!ht]
    \centering
    \includegraphics[width=0.95\textwidth]{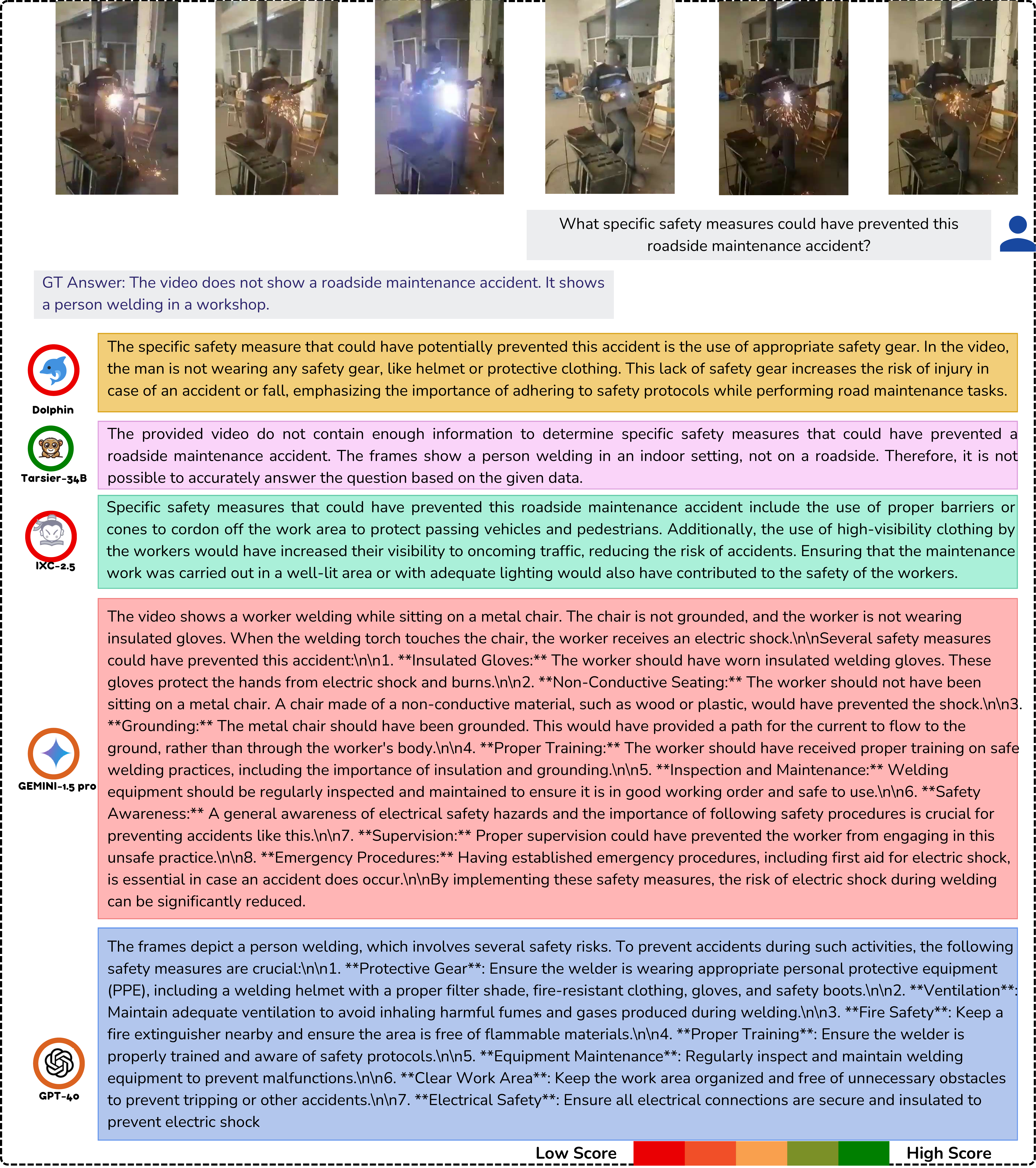}
    \caption{\textbf{Model performance comparison on Incompatible QA task}: Top: Frames showing indoor welding activity in a workshop. Middle: Models are asked about roadside maintenance accident prevention, while the ground truth (gray box) correctly indicates that the content shows indoor welding, not roadside maintenance. Bottom: Model responses (colored boxes) demonstrate varying degrees of hallucination: while Tarsier (green) correctly acknowledges insufficient information to discuss roadside maintenance, Dolphin and IXC (red circles) fabricate elaborate safety measures despite the obvious indoor setting. Gemini and GPT-4o's (dark orange) detailed response about welding safety, while technically accurate, still fails to address the fundamental context mismatch. This example illustrates how models can generate plausible but irrelevant safety recommendations when presented with visually similar but contextually different scenarios. Refer back to \cref{sec:qualitative}.}
    \label{fig:QA_incompatible_3}
\end{figure*}

\begin{figure*}[!ht]
    \centering
    \includegraphics[width=0.95\textwidth]{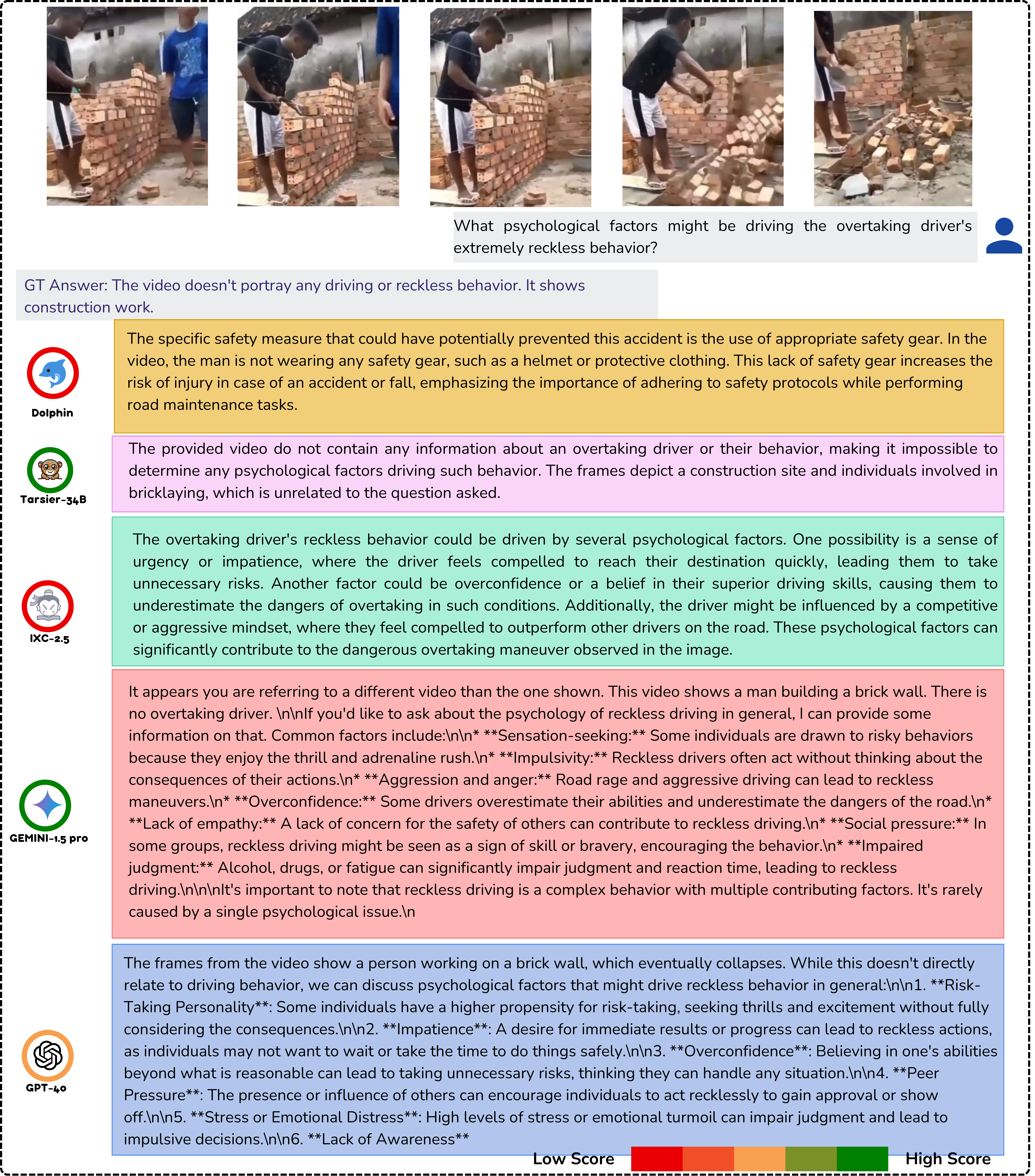}
    \caption{\textbf{Model performance comparison on Incompatible QA task}: Top: Frames showing a person building and subsequently demolishing a brick wall. Middle: Models are asked about psychological factors behind overtaking behavior, while the ground truth (gray box) correctly mentions this as unrelated to overtaking. Bottom: Model responses show varying levels of hallucination and context confusion: Dolphin and IXC (red circles) completely ignore the brick wall context and fabricate scenarios about road safety, while Tarsier and Gemini (green) correctly acknowledges the construction setting and clearly states the content mismatch. This example demonstrates how models can struggle with maintaining contextual accuracy, with some generating elaborate but irrelevant psychological analyses despite clearly unrelated visual content. Refer back to \cref{sec:qualitative}.}
    \label{fig:QA_incompatible_4}
\end{figure*}

\begin{figure*}[!ht]
    \centering
    \includegraphics[width=0.9\textwidth]{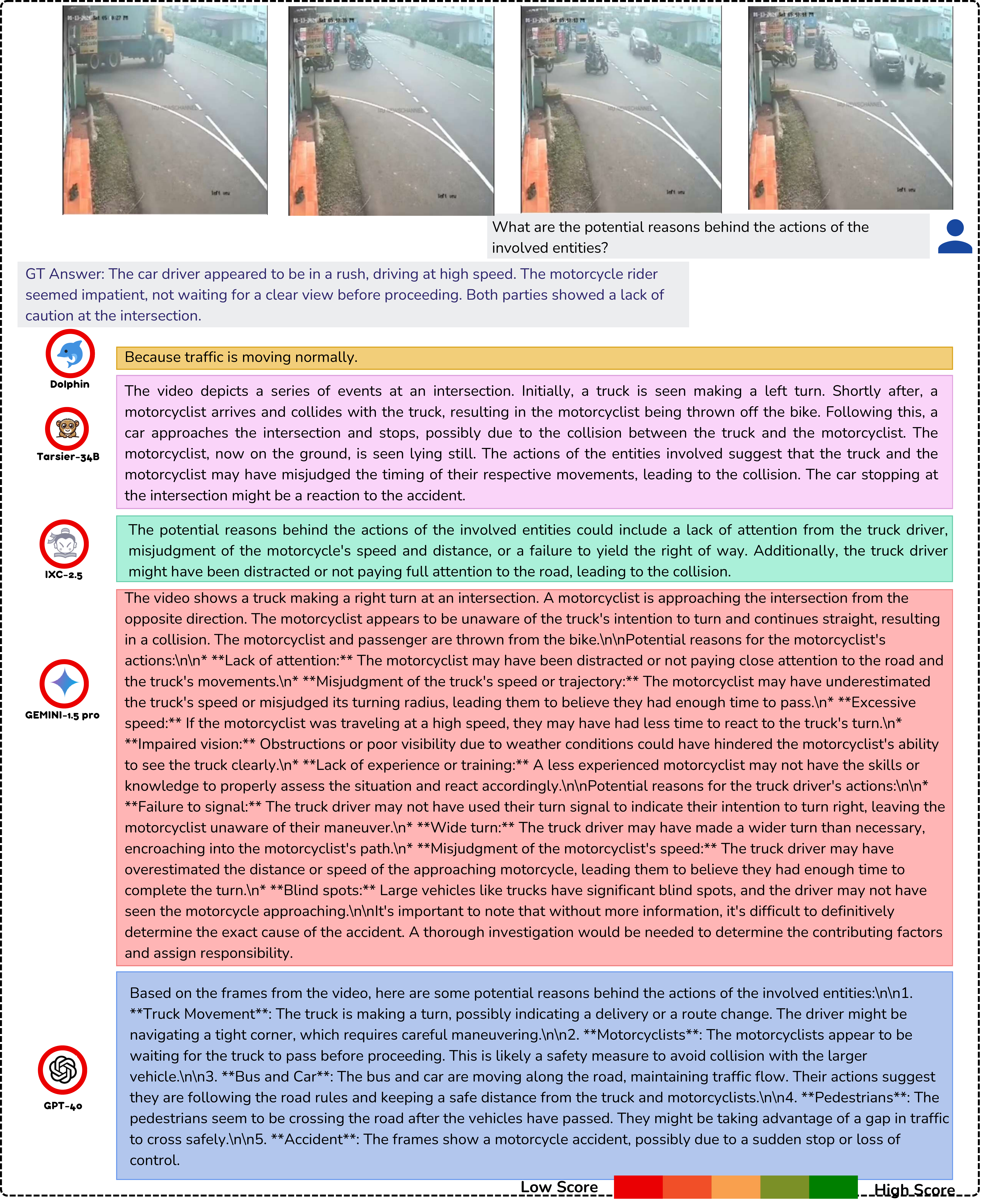}
    \caption{\textbf{Model performance comparison on Why QA task}: Top: CCTV footage showing an intersection incident between a truck and motorcyclist. Middle: Models are asked about potential reasons behind the road entities' actions, with ground truth (gray box) indicating rush and lack of caution as primary factors. Bottom: Model responses demonstrate varying levels of reasoning and detail: While Dolphin (red circle) provides an oversimplified response ('Because traffic moving normally'), other models offer increasingly complex analyses. Gemini generates a comprehensive analysis considering multiple factors (weather conditions, road visibility, driver attention), while GPT-4o provides a structured but possibly over-analyzed response with enumerated factors. This example illustrates the challenge of providing appropriate depth in causal reasoning without over-speculation. Refer back to \cref{sec:qualitative}.}
    \label{fig:QA_why_1}
\end{figure*}

\begin{figure*}[!ht]
    \centering
    \includegraphics[width=\textwidth]{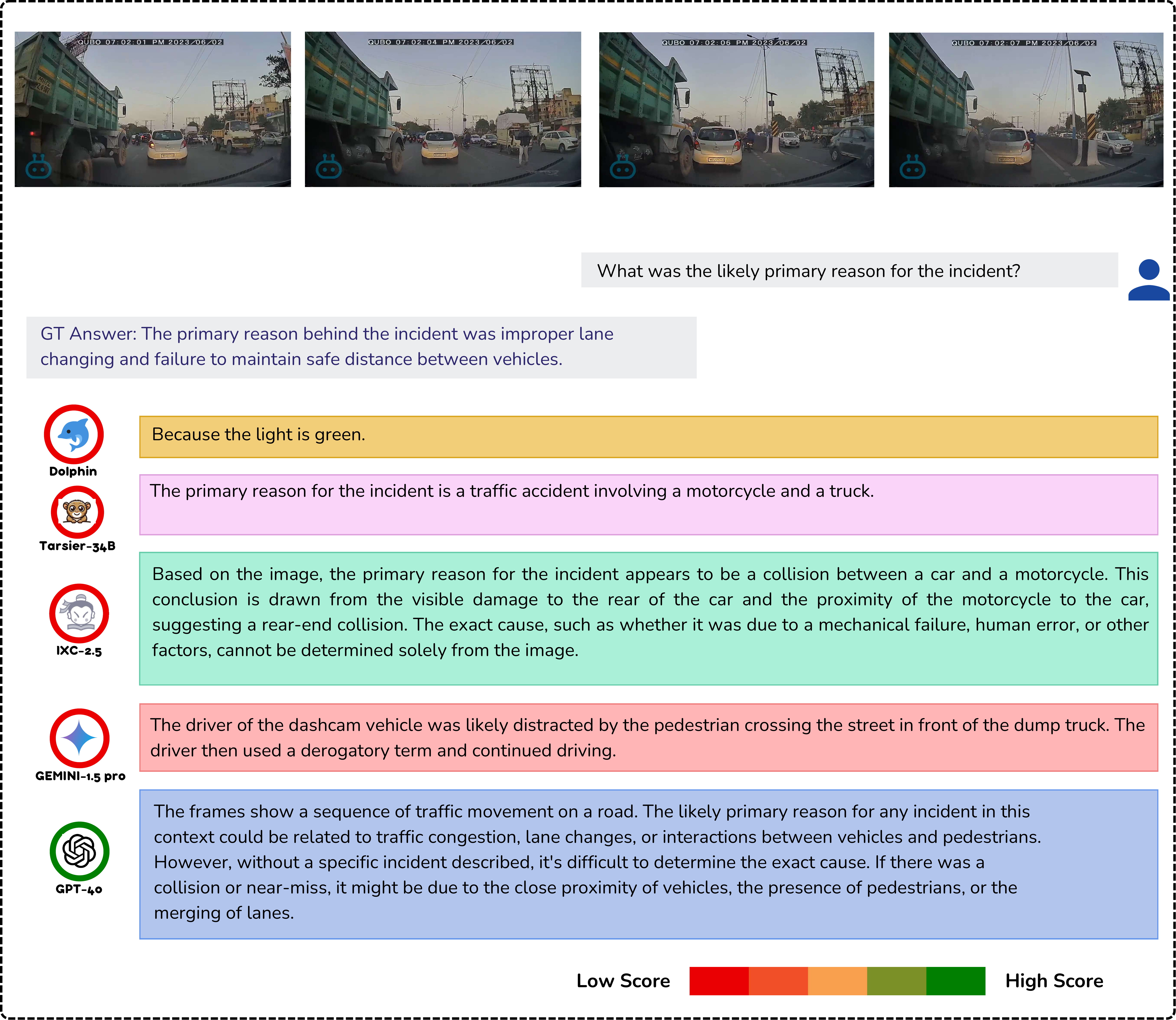}
    \caption{\textbf{Model performance comparison on Why QA task}: Top: Dashcam footage showing a traffic scenario with lane changing incidents. Middle: Models are asked about the primary reason for the incident, with ground truth (gray box) identifying improper lane changing and insufficient vehicle distance as key factors. Bottom: Model responses show varying levels of analytical accuracy and specificity: Dolphin offers an oversimplified and irrelevant response ('Because the light is green'), while Tarsier-34B provides a vague description without specific reasoning. IXC-2.5 attempts causal analysis but misidentifies the vehicles involved, and Gemini-1.5 Pro introduces unobserved elements (pedestrian crossing). GPT-4o demonstrates appropriate caution by acknowledging the difficulty in determining exact causes without clearer incident details. This example highlights the challenges in balancing between definitive causal analysis and appropriate uncertainty when visual evidence is ambiguous. Refer back to \cref{sec:qualitative}.}
    \label{fig:QA_why_2}
\end{figure*}

\begin{figure*}[!ht]
    \centering
    \includegraphics[width=0.85\textwidth]{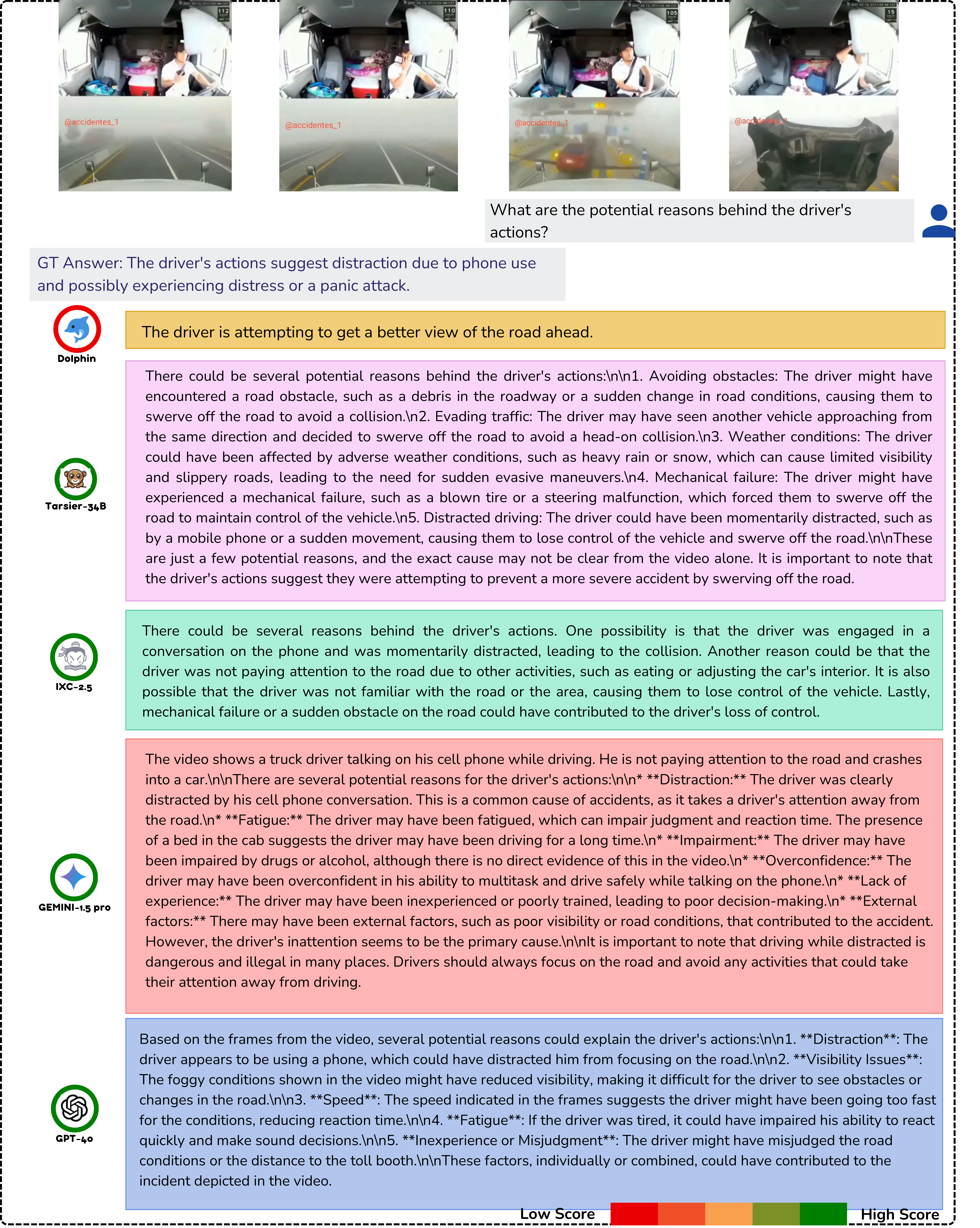}
    \caption{\textbf{Model performance comparison on Why QA task}: Top: Split-screen dashcam footage showing both driver behavior (phone use) and road view leading to an incident. Middle: Models are asked about potential reasons behind the driver's actions, with ground truth (gray box) identifying distraction from phone use and possible panic/distress. Bottom: Model responses demonstrate varying depths of causal analysis: Dolphin provides an oversimplified response about road view, while Gemini-1.5 Pro offers a comprehensive multi-factor analysis incorporating both observed behaviors (phone distraction) and possible underlying causes. IXC-2.5 stays focused on direct observables, while GPT-4V extensively analyzes multiple scenarios but maintains grounding in the visible evidence (phone conversation). This example shows how models balance between observable evidence (phone use) and inferring potential psychological states, with varying success in maintaining relevance to the visual content. Refer back to \cref{sec:qualitative}.}
    \label{fig:QA_why_3}
\end{figure*}

\begin{figure*}[!ht]
    \centering
    \includegraphics[width=\textwidth]{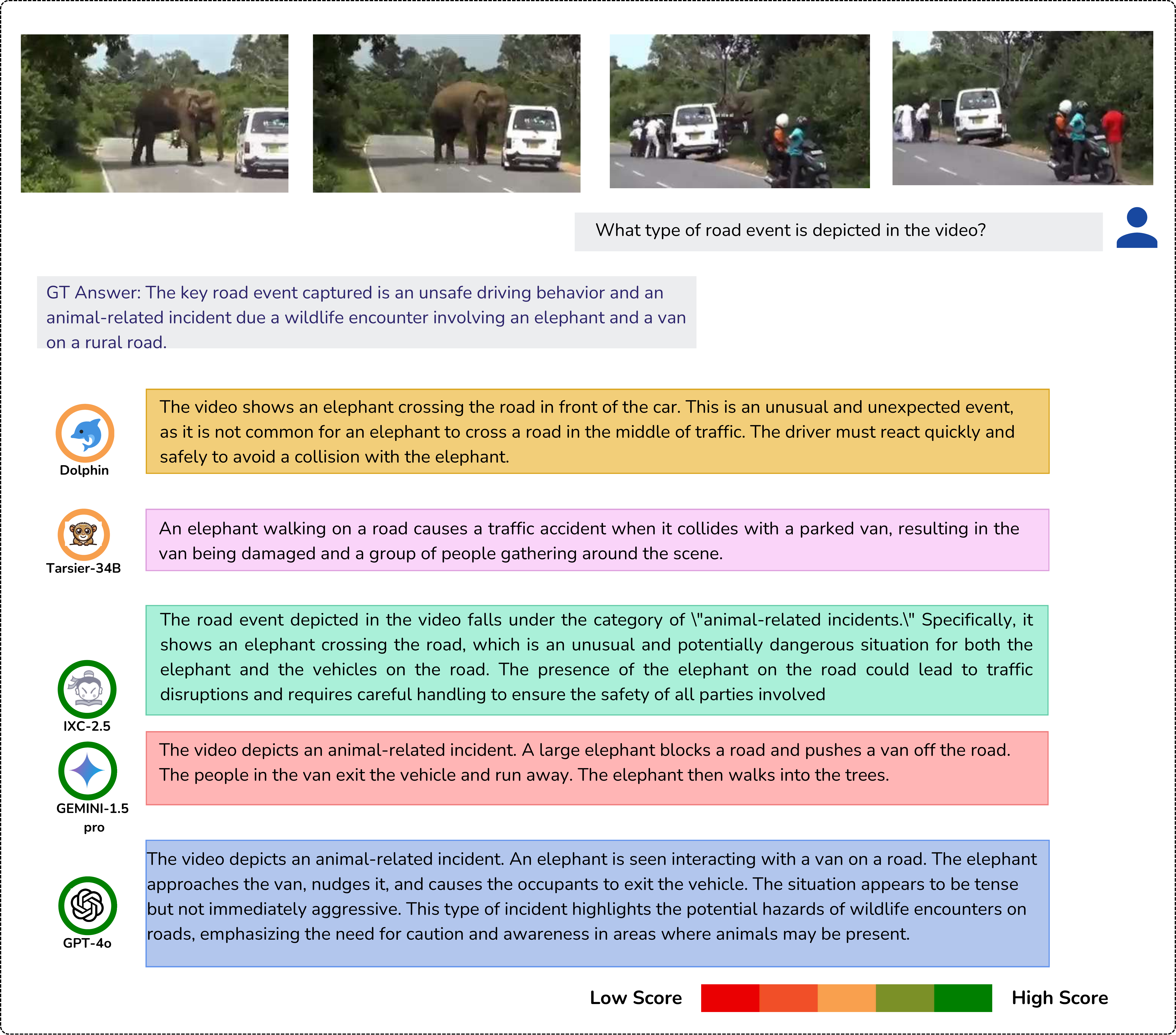}
    \caption{\textbf{Model performance comparison on Description QA task}: Top: Video frames showing an encounter between an elephant and a van on a rural road. Middle: Models are asked to describe the type of road event, with ground truth (gray box) identifying it as both unsafe driving behavior and an animal-related wildlife encounter. Bottom: Model responses show varying accuracy in event categorization and detail: Tarsier-3LB incorrectly describes a collision, while IXC-2.5 (green circle) provides a well-balanced response that correctly categorizes the event as 'animal-related incident' while acknowledging the safety implications for all parties. Gemini-1.5 Pro and GPT-4o offer accurate but differently focused descriptions, with Gemini emphasizing the physical interaction and GPT-4o highlighting the broader safety context. This example demonstrates models' varying abilities to balance between event classification, factual description, and safety implications in unusual road scenarios. Refer back to \cref{sec:qualitative}.}
    \label{fig:QA_descp_1}
\end{figure*}

\begin{figure*}[!ht]
    \centering
    \includegraphics[width=\textwidth]{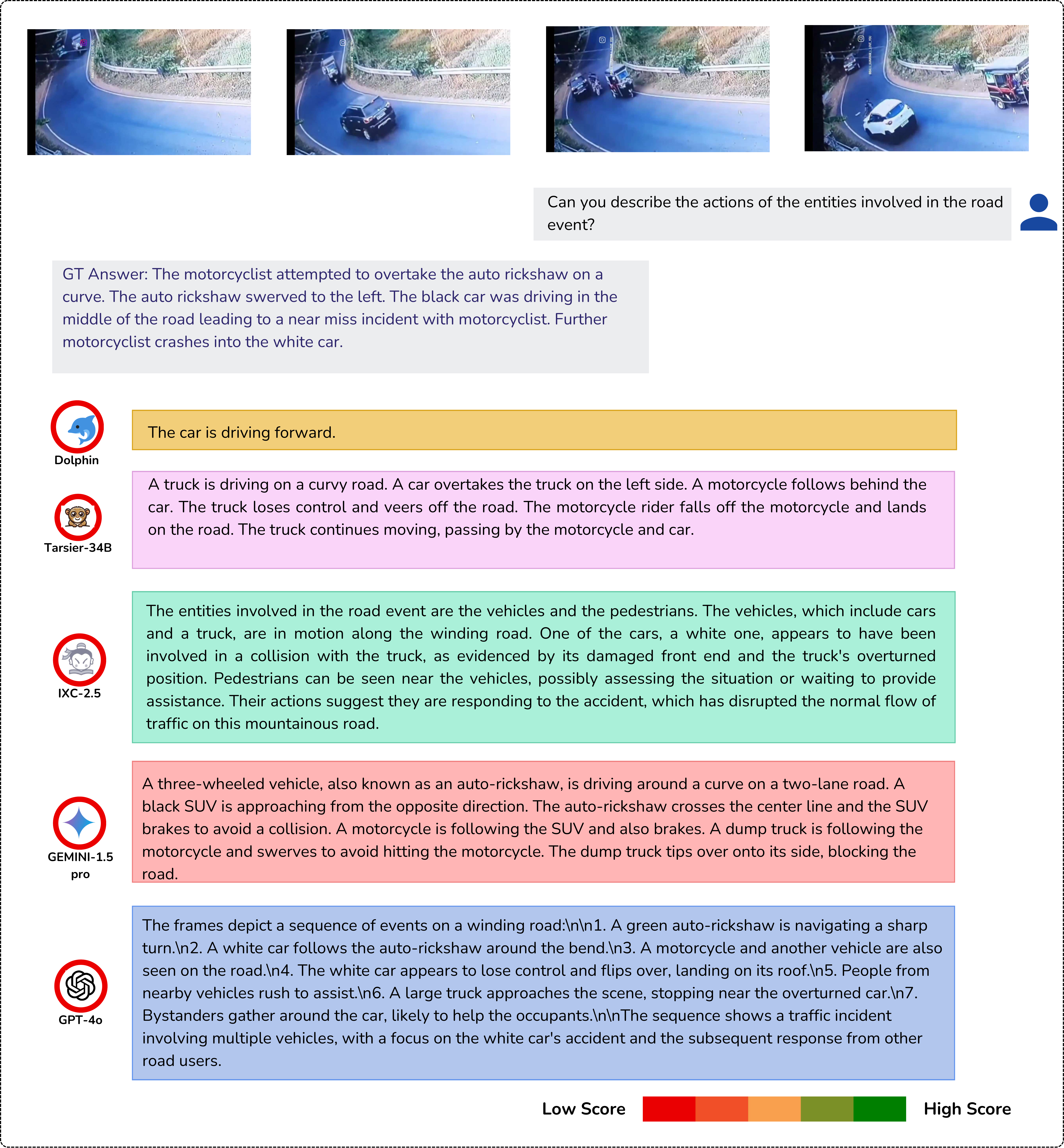}
    \caption{\textbf{Model performance comparison on Description QA task}: 
 Top: Video frames show a collision between a car and a bike on a curvy road. Middle: Models are asked to describe the actions of the entities involved in the road event, with ground truth (gray box) identifying it near-miss incident that further led to the collision. Bottom: Model responses show varying accuracy in event categorization and detail: All the models fail to answer this question due to incorrect identifications. GPT-4o fails to identify the motorcycle that was initially overtaking the auto that crashed a car. Tarsier-34B incorrectly identifies overtaking between the car and the truck. Refer back to \cref{sec:qualitative}.}
    \label{fig:QA_descp_2}
\end{figure*}

\begin{figure*}[!ht]
    \centering
    \includegraphics[width=\textwidth]{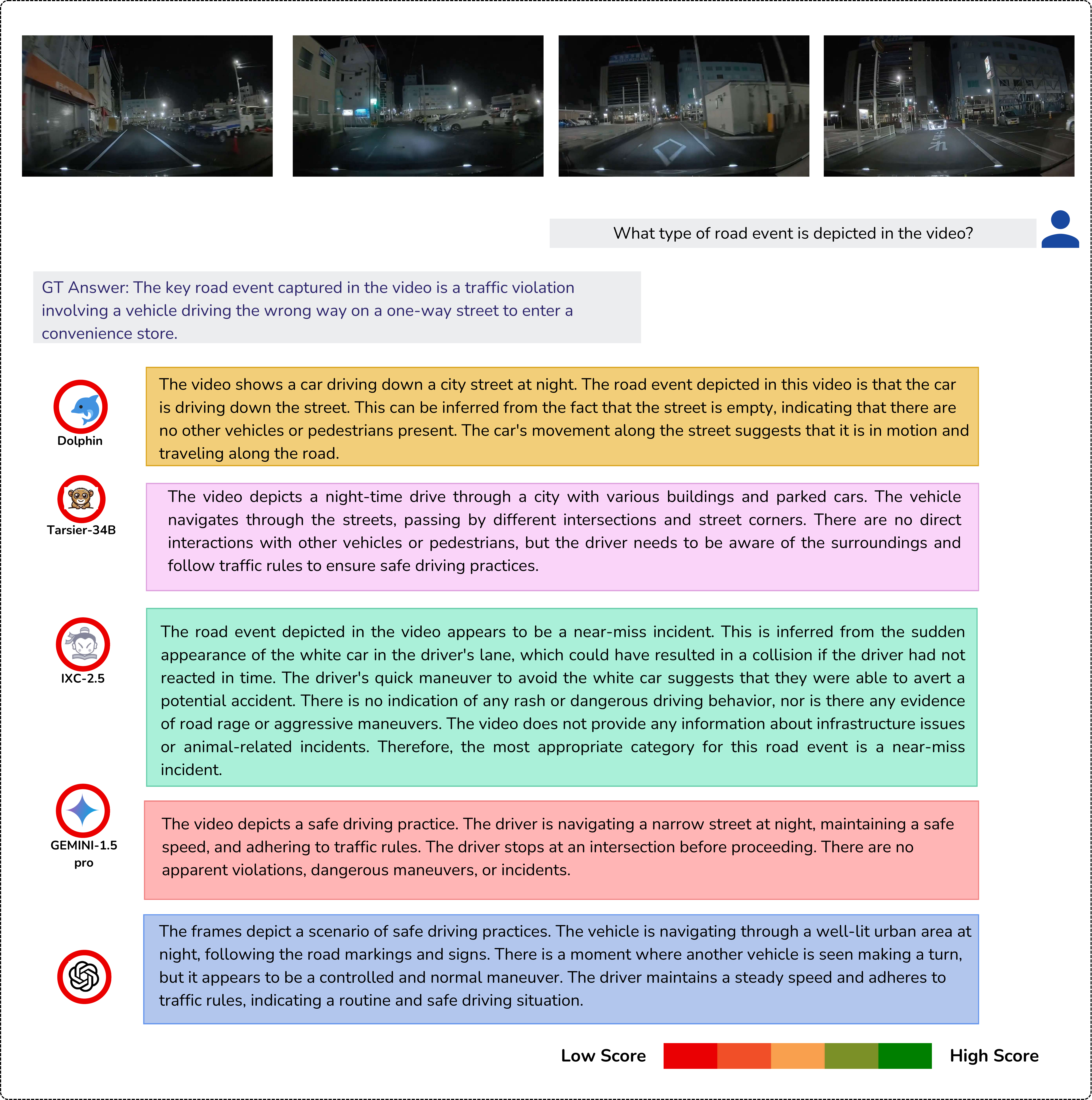}
    \caption{\textbf{Model performance comparison on Description QA task}: 
 Top: Video frames show a traffic violation involving. Middle: Models are asked to describe the type of road event depicted in the video, with ground truth (gray box) identifying it as a vehicle driving the wrong way on a one-way street. Bottom: All models fail to recognize the violation. Refer back to \cref{sec:qualitative}.}
    \label{fig:QA_descp_3}
\end{figure*}

\begin{figure*}[!ht]
    \centering
    \includegraphics[width=\textwidth]{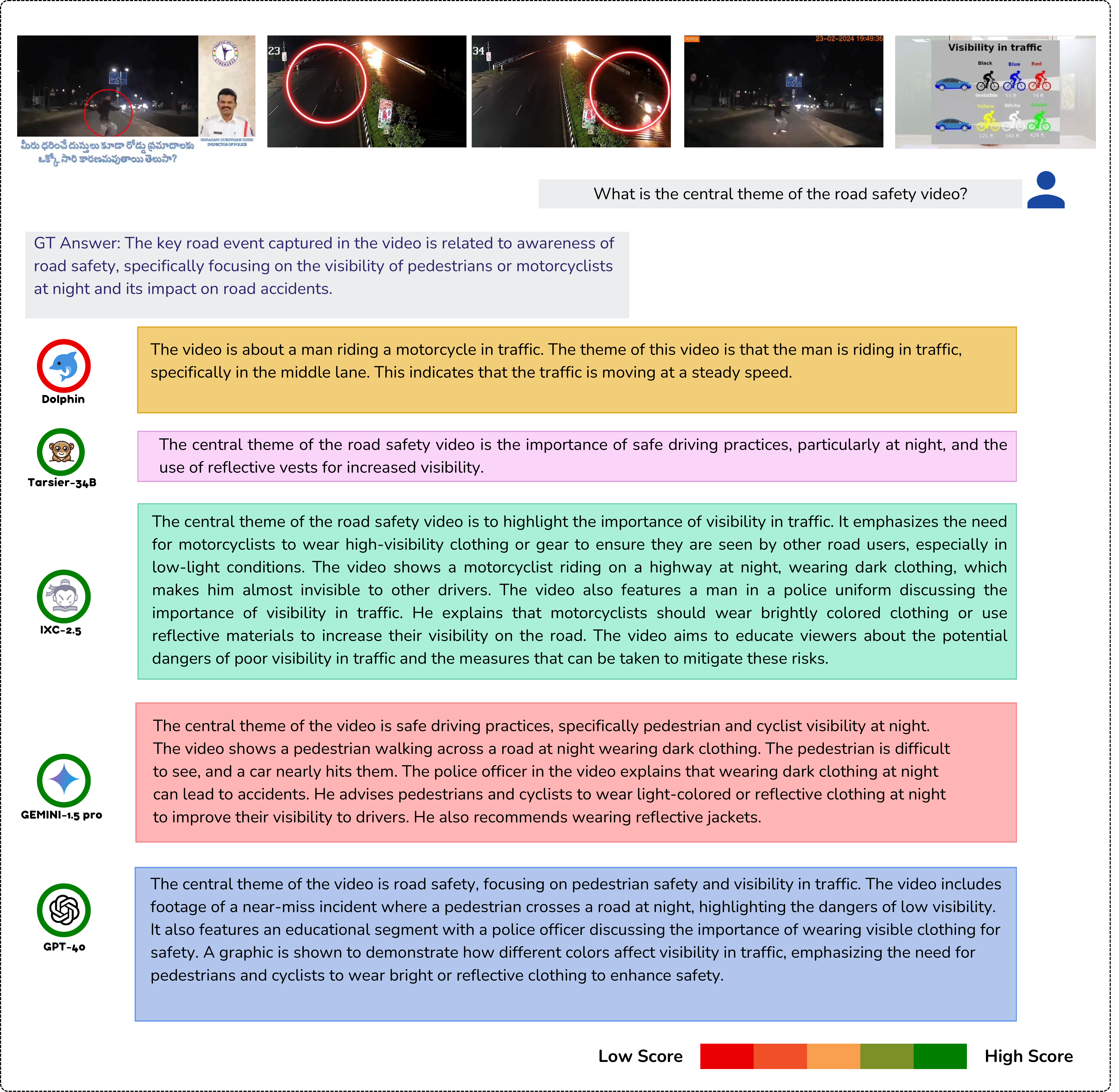}
    \caption{\textbf{Model performance comparison on Description QA task}: Top: Video frames showing a road safety awareness video aimed towards pedestrians or motorcyclists at night. Middle: Models are asked to describe the theme of the video, with ground truth (gray box) indicating that it is a safety awareness video. Bottom: Model responses show varying accuracy in event categorization and detail: All the models except Dolphin successfully capture the global context or theme of the video. Refer back to \cref{sec:qualitative}.}
    \label{fig:QA_descp_4}
\end{figure*}

\begin{figure*}[!ht]
    \centering
    \includegraphics[width=\textwidth]{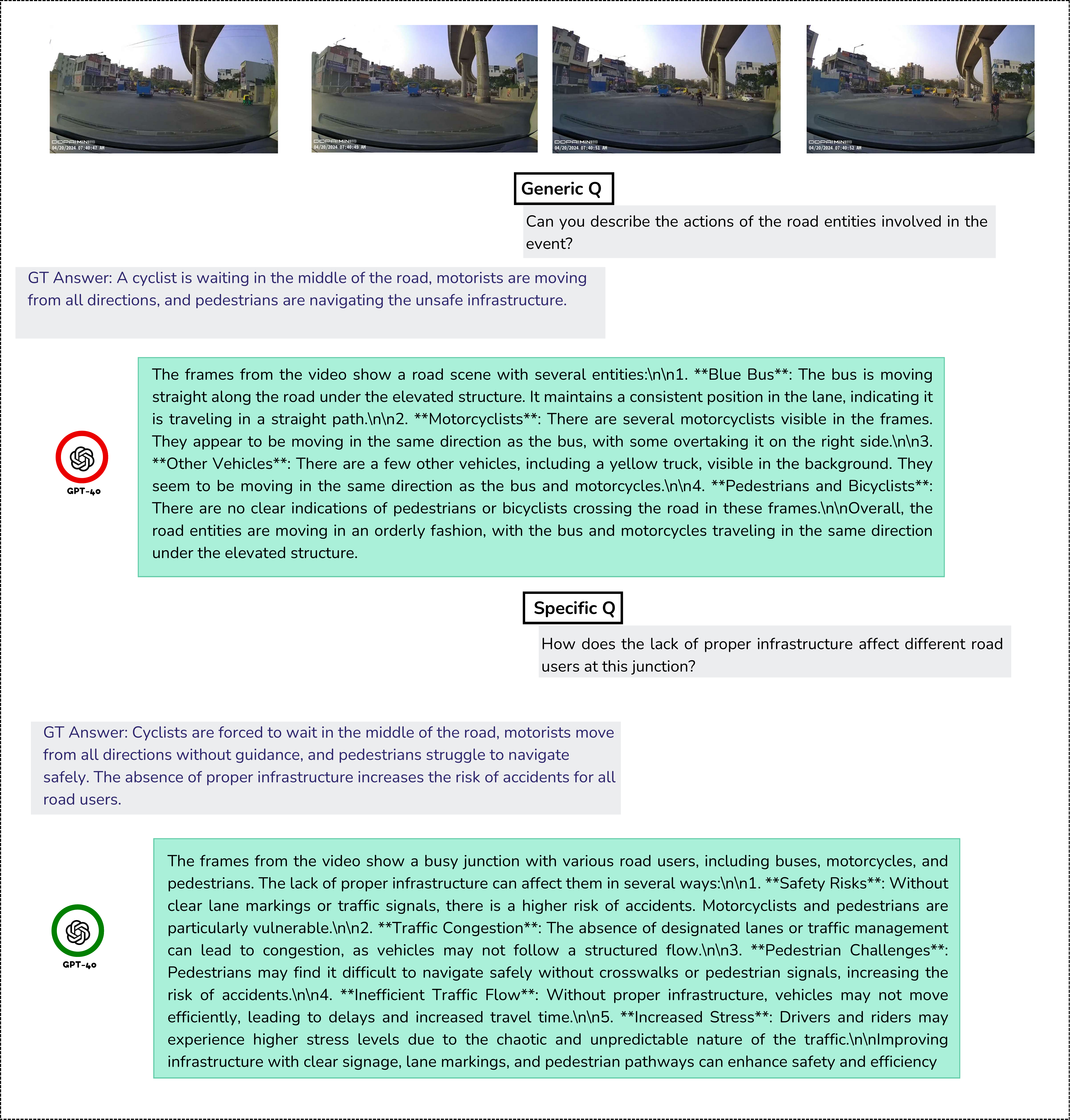}
    \caption{The image shows a qualitative analysis of the performance of GPT-4o Video LLM for two types of questions - a generic question about the actions of the road entities, and a specific question about how the lack of proper infrastructure affects different road users. The ground truth (GT) answers are provided, and the predicted answers by the model are shown using icons - a red circle indicates the model's prediction does not align well with the ground truth, while a green icon indicates the model performs well. GPT-4o seems to be performing well in specific questions than generic one. This performance gap could be because generic questions require the model to infer the context while specific questions directly reference the event and entities, making it easier for models to answer them.}
    \label{fig:SpecificQA1}
\end{figure*}

\begin{figure*}[!ht]
    \centering
    \includegraphics[width=\textwidth]{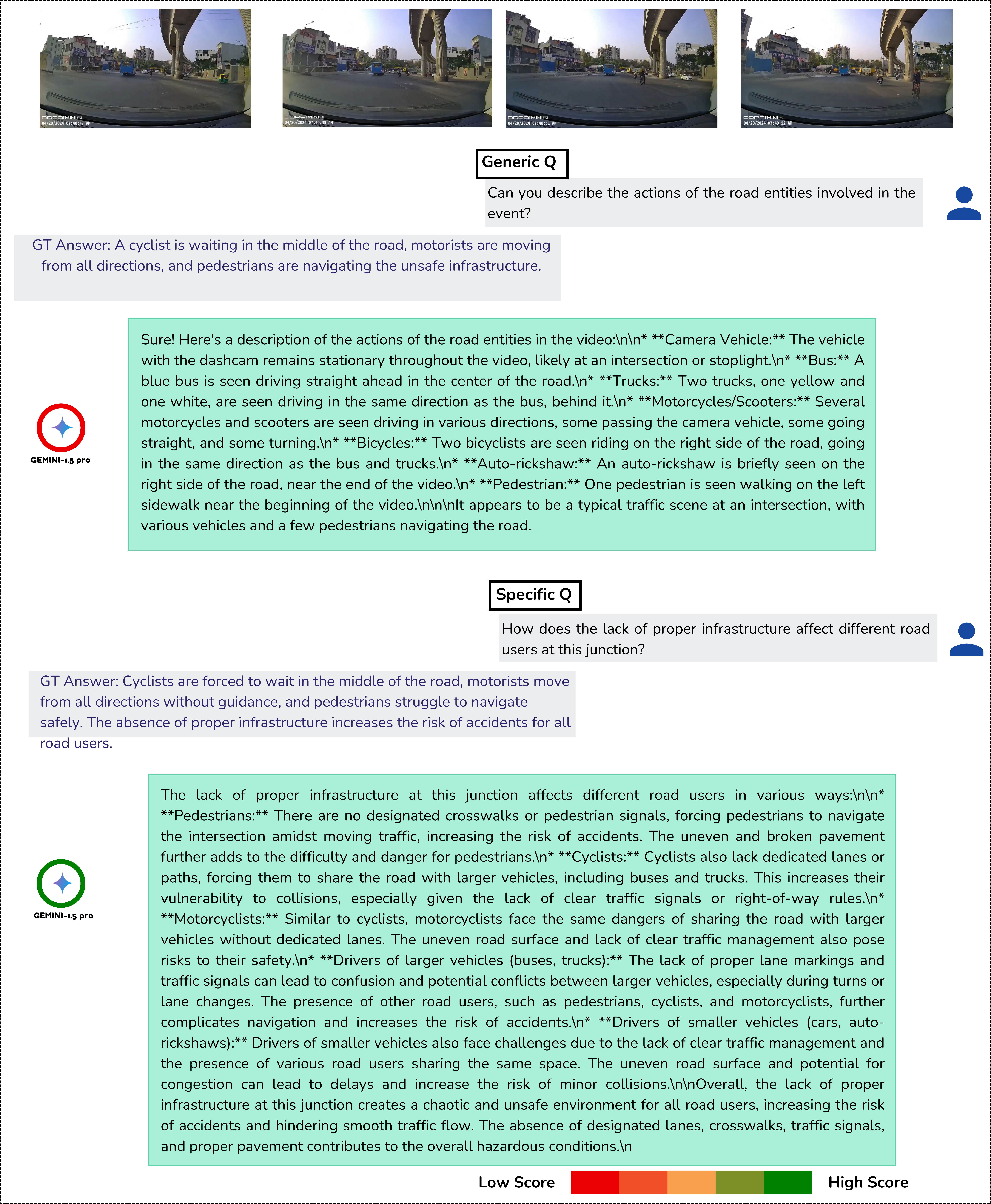}
    \caption{A similar phenomena between the gap between generic and specific QAs is reflected in Gemini, as seen in the previous example.}
    \label{fig:SpecificQA2}
\end{figure*}

\begin{figure*}[!ht]
    \centering
    \includegraphics[width=\textwidth]{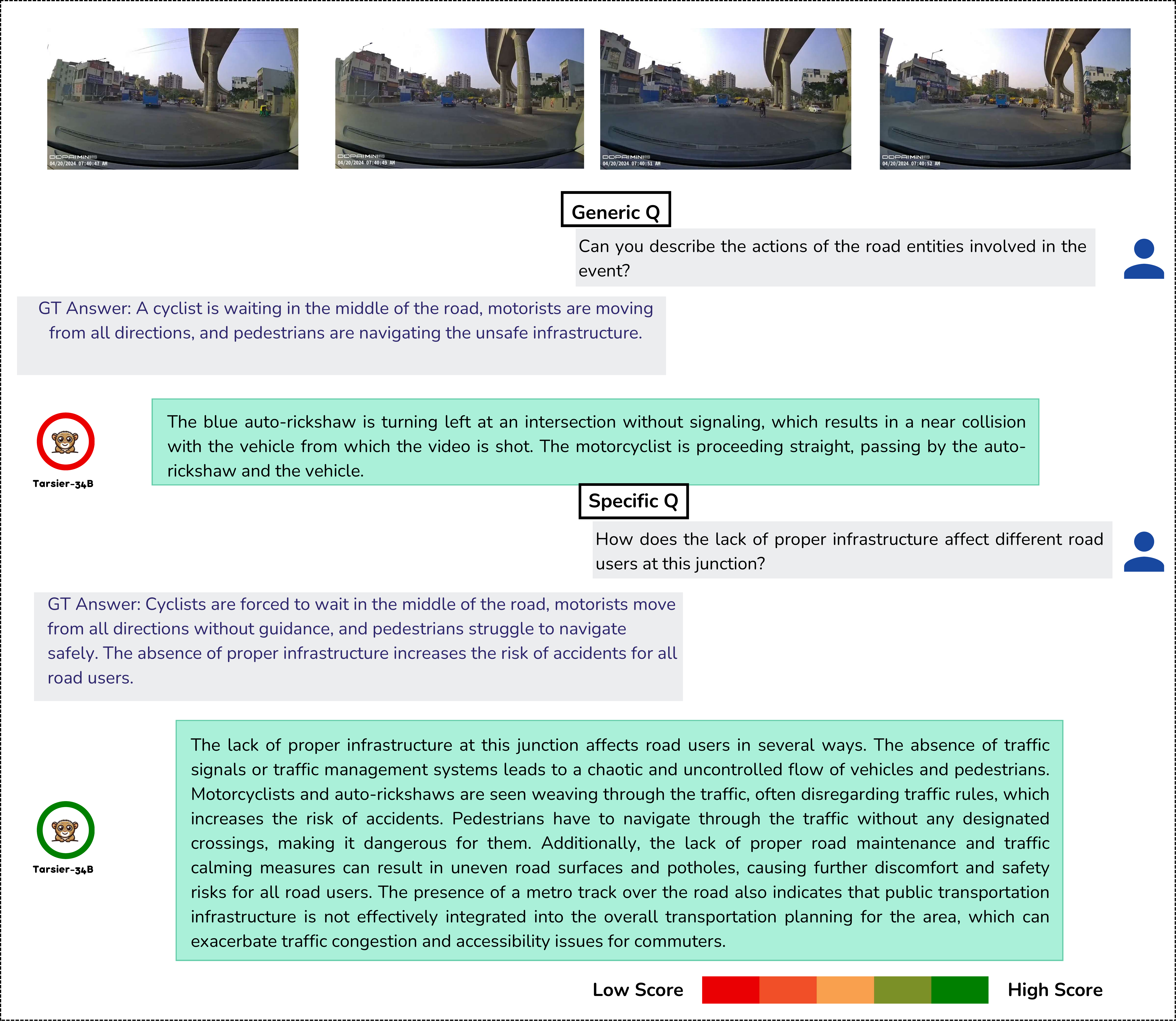}
    \caption{A similar phenomena indicating the gap between generic and specific QAs is reflected in Tarsier, similar to what was observed in the Gemini and GPT-4o in previous examples.}
    \label{fig:SpecificQA3}
\end{figure*}

\begin{figure*}[!ht]
    \centering
    \includegraphics[width=\textwidth]{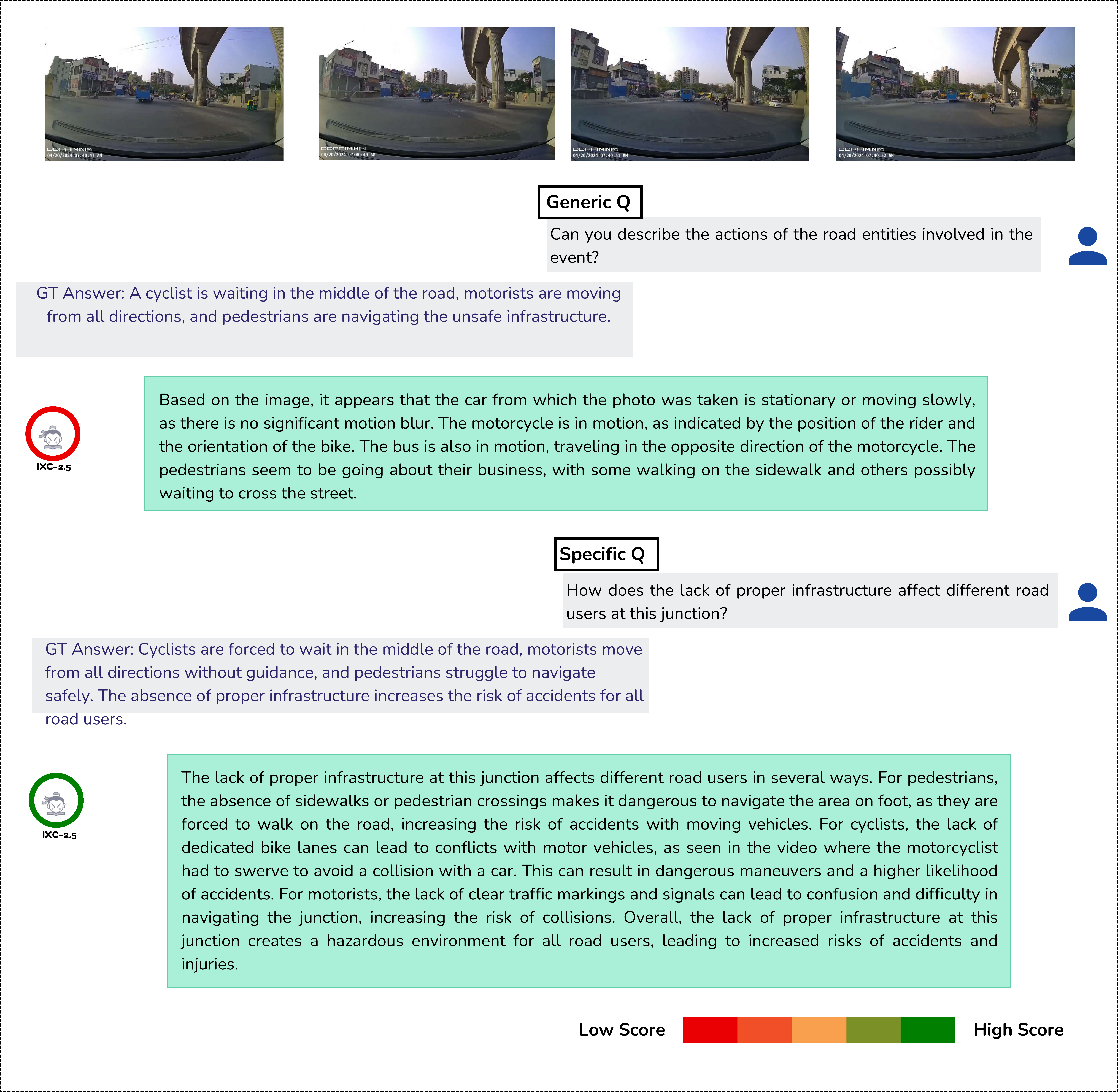}
    \caption{The phenomena of the model performing better in specific QAs than their generic counterparts persist in IXC as well.}
    \label{fig:SpecificQA4}
\end{figure*}

\subsection{Video-level Tag Generation}
\label{sec:video_tag_gen}

To enable efficient retrieval and analysis of videos based on content characteristics, we developed a tag generation system that generates diverse video-level tags from refined QA pairs (\cref{sec:refinement_and_categorization}). Our method employs Claude 3.5 Sonnet Text LLM~\cite{claude_3_5} to analyze QA pairs and generate tags across multiple categories (as shown in Fig 1, Main Paper).

For each QA pair, based on its template question category (\cref{sec:qa_taxonomy}), the LLM generates specific tags following structured guidelines provided in prompt \cref{fig:tag_gen_prompts_1} - \ref{fig:tag_gen_prompts_5}).

\begin{figure}[!t]
    \centering
    \includegraphics[width=\linewidth]{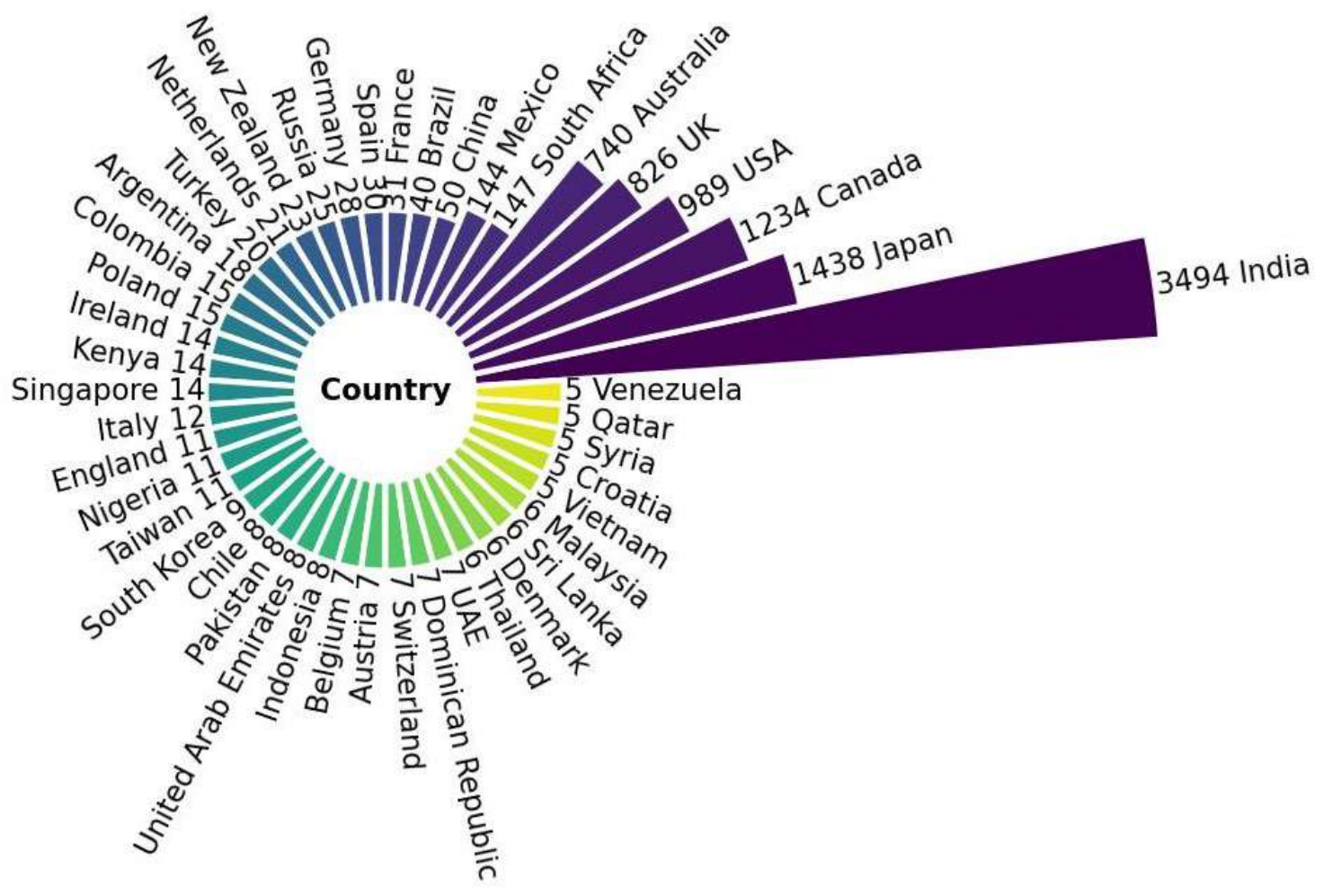}
    \caption{Geographical location (country of origin) distribution of video tags. Tags with fewer than five videos are omitted from the radar plot for clarity and to reduce clutter.}
    \label{fig:country_tags}
\end{figure}

\begin{figure}[!t]
    \centering
    \includegraphics[width=\linewidth]{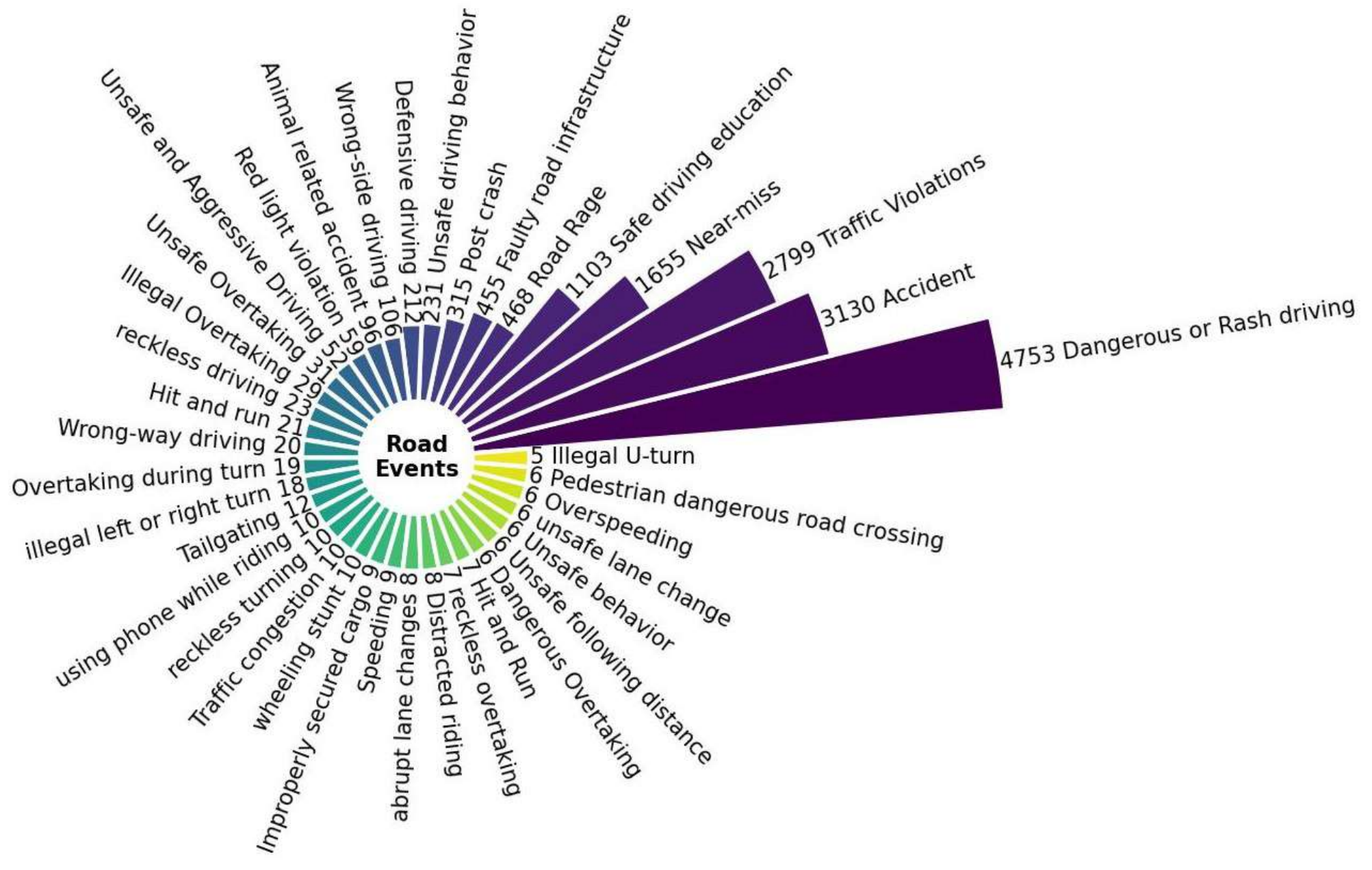}
    \caption{Road Event Video Tags distribution. Tags with fewer than five videos are omitted from the radar plot for clarity and to reduce clutter.}
    \label{fig:roadevent_tags}
\end{figure}

\begin{figure}[!t]
    \centering
    \includegraphics[width=\linewidth]{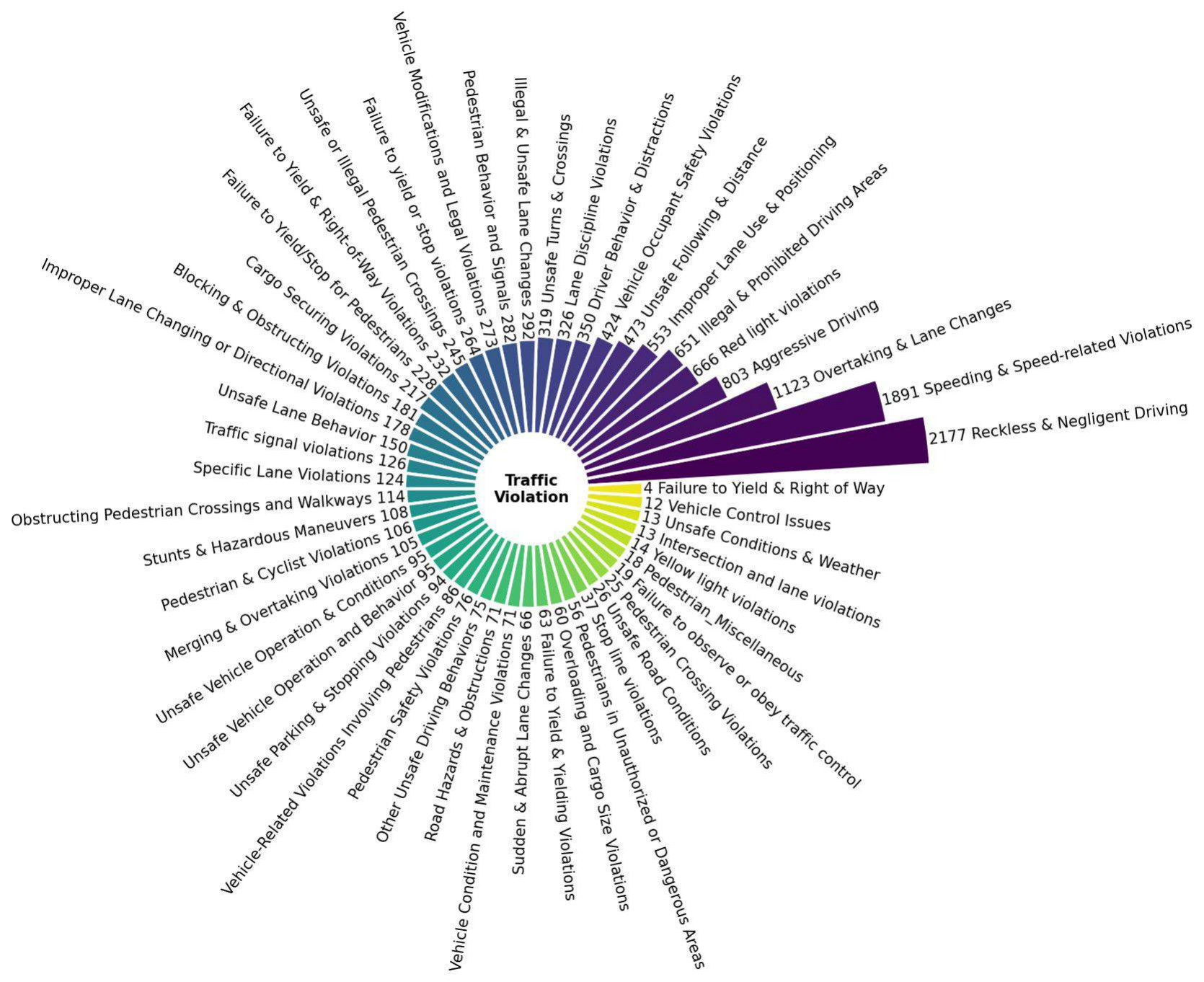}
    \caption{Traffic Violation Video Tags distribution. Tags with fewer than four videos are omitted from the radar plot for clarity and to reduce clutter.}
    \label{fig:Traffic_Violation_tags}
\end{figure}

\begin{figure}[!t]
    \centering
    \includegraphics[width=\linewidth]{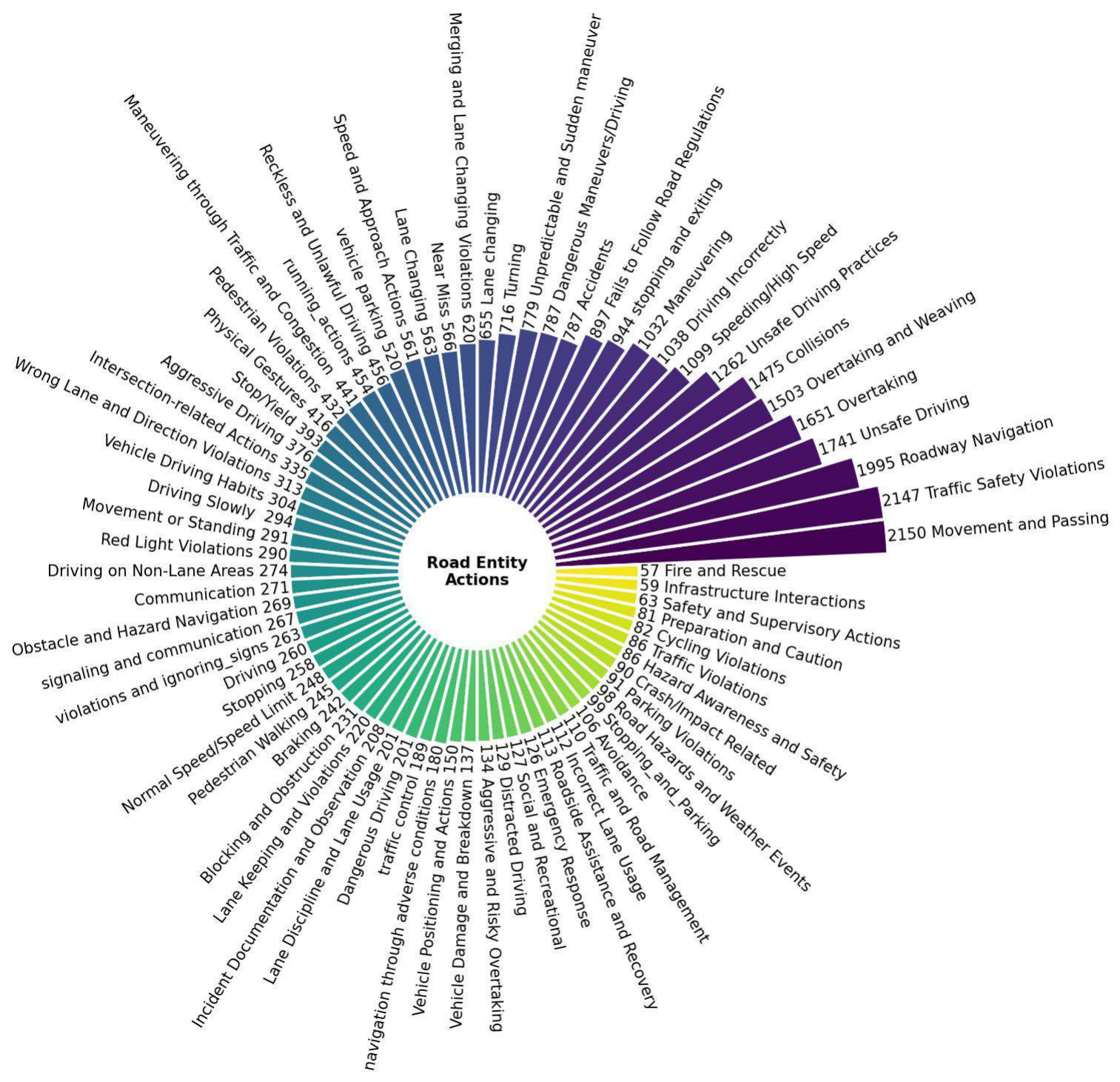}
    \caption{Road Entity Action Video Tags distribution. Tags with fewer than 57 videos are omitted from the radar plot for clarity and to reduce clutter.}
    \label{fig:RoadEntityAction_tags}
\end{figure}

\begin{figure}[!t]
    \centering
    \includegraphics[width=\linewidth]{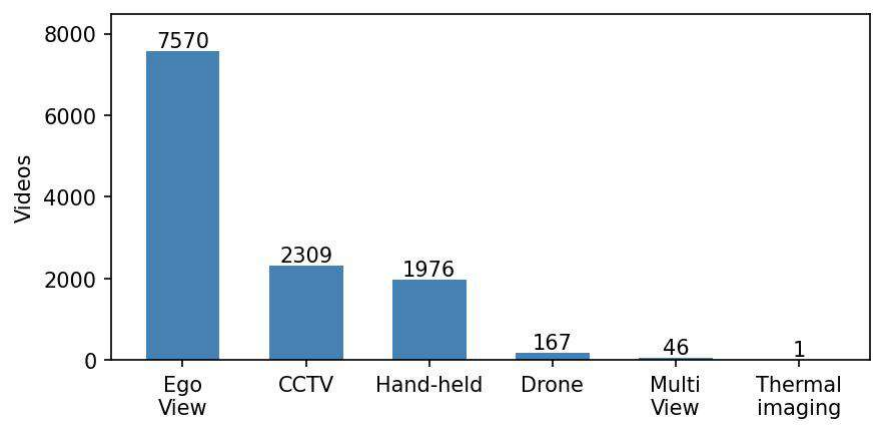}
    \caption{Viewpoint Video Tags distribution.}
    \label{fig:Viewpoint_tags}
\end{figure}

\begin{figure}[!t]
    \centering
    \includegraphics[width=\linewidth]{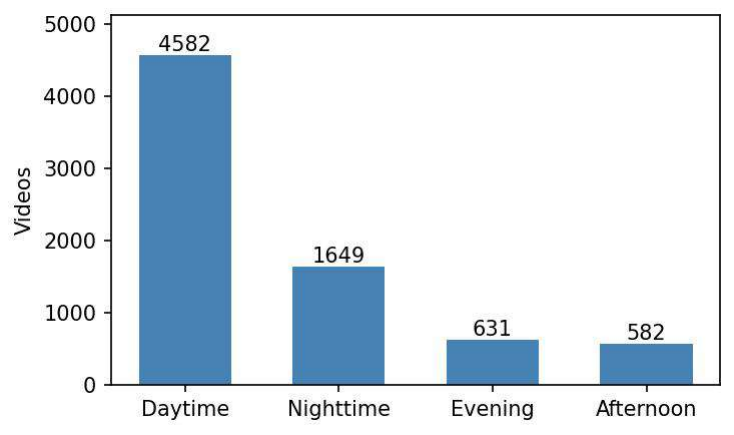}
    \caption{Time of Day Tags distribution.}
    \label{fig:timeofday_tags}
\end{figure}

The resulting video tags or video attributes provide fine-grained details about road scenarios, enabling efficient video retrieval and analysis. Their distribution is shown in \cref{fig:country_tags} - \ref{fig:timeofday_tags}.

\begin{figure}[!t]
    \centering
    \includegraphics[width=\linewidth]{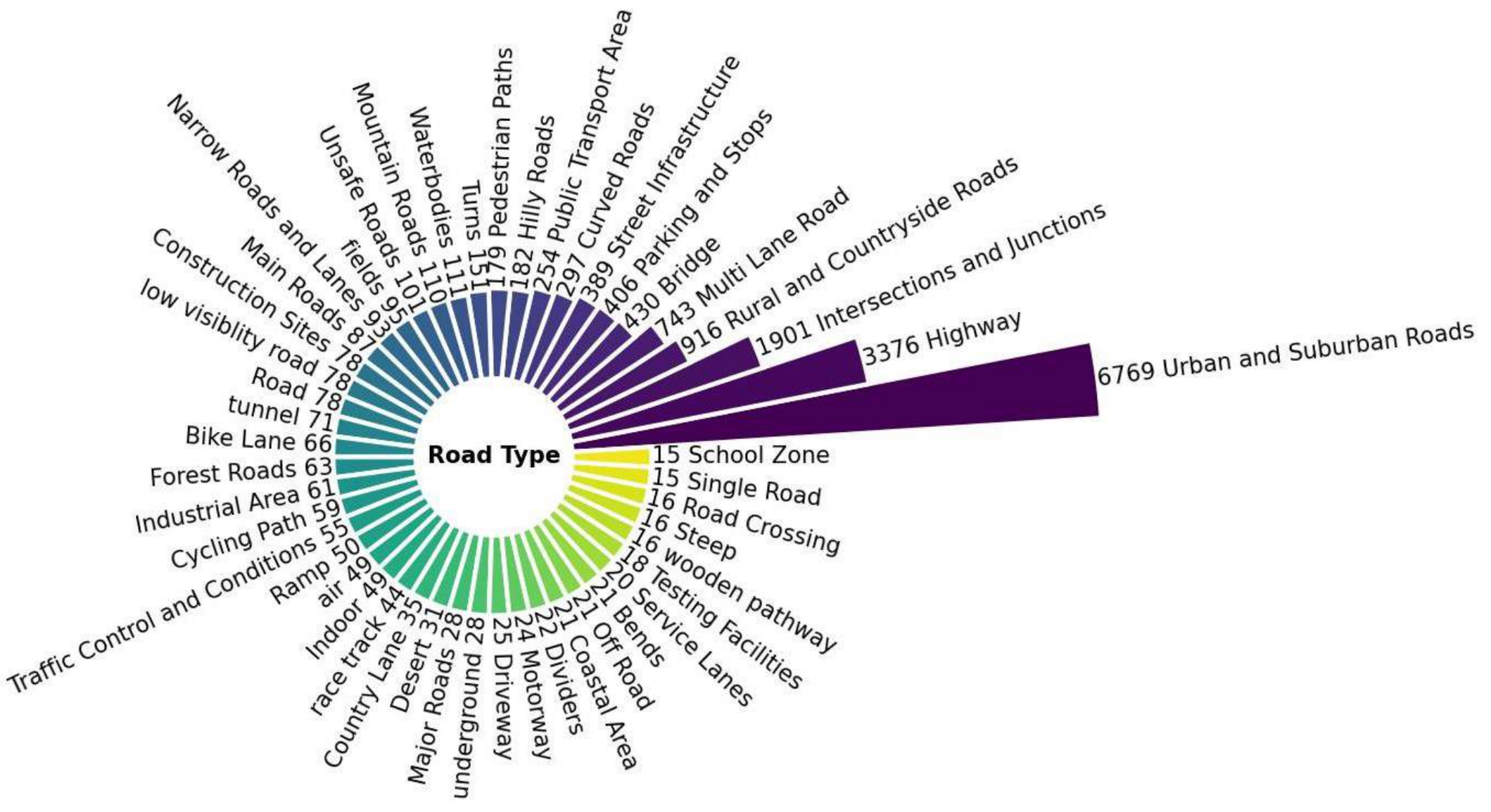}
    \caption{Road Type Video Tags distribution. Tags with fewer than 15 videos are omitted from the radar plot for clarity and to reduce clutter.}
    \label{fig:roadtype_tags}
\end{figure}

\begin{figure}[!t]
    \centering
    \includegraphics[width=\linewidth]{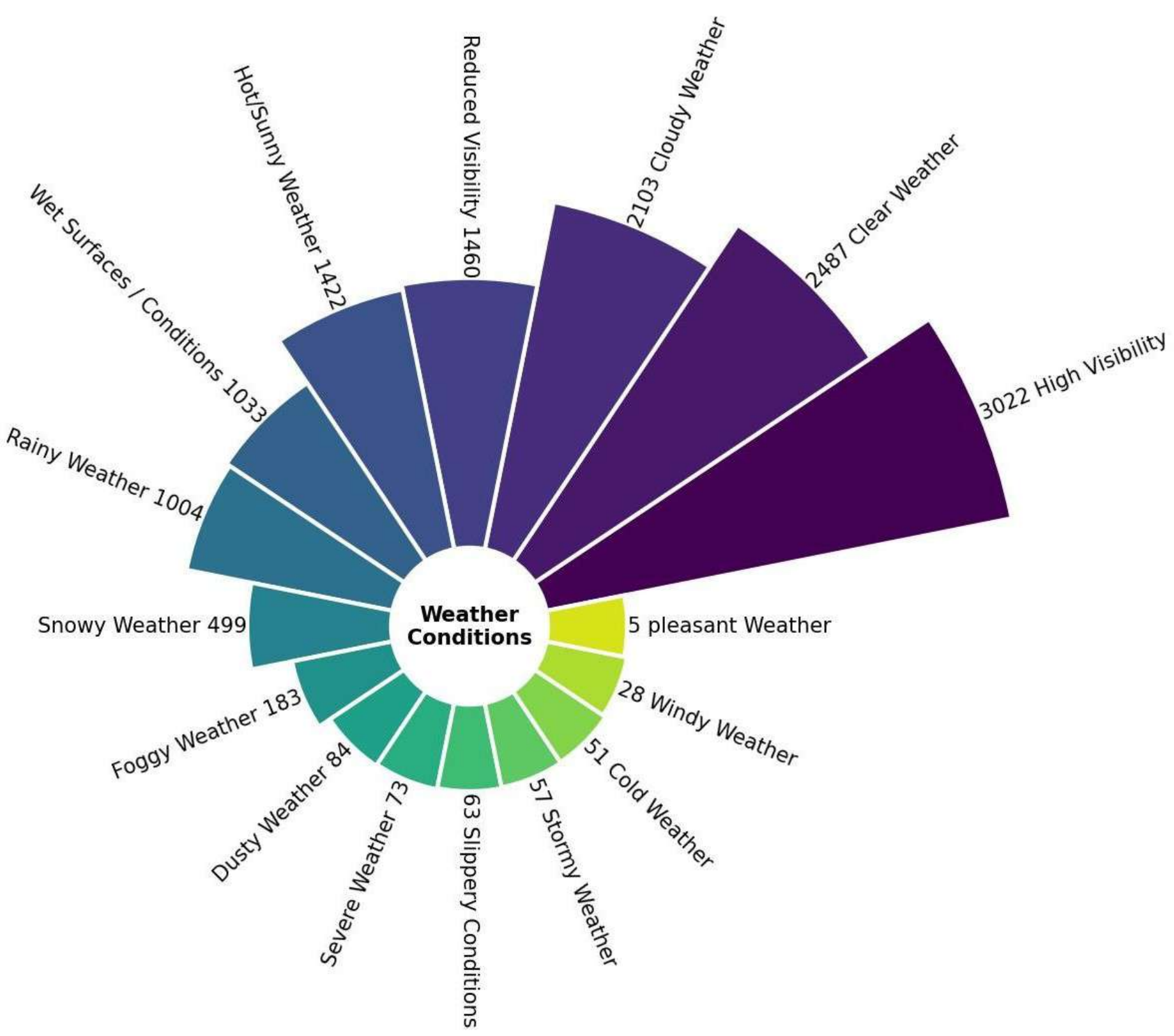}
    \caption{Weather Condition Video Tags distribution. Tags with fewer than five videos are omitted from the radar plot for clarity and to reduce clutter.}
    \label{fig:weather_tags}
\end{figure}

\section{Experiments}
\label{sec:experiments_supp}

\subsection{Data Setup}
\label{sec:data_setup}

For model evaluation, we provide the model with video frames and a task-specific question, following the format: video frames $+$ model's default system prompt (if any) $+$ our task-specific question. An example prompt for LLaVA-Video~\cite{zhang2024videoinstructiontuningsynthetic} in the specified format is given in \cref{fig:std_prompt}. Also, the prompting structure for QA tasks in our dataset is described in \cref{fig:prompt_strategy}.

\subsection{Model Setup}
\label{sec:model_setup_supp}

For evaluating model-generated responses against ground-truth answers in our open-ended QA tasks, we employ GPT-3.5 score~\cite{openai2022gpt35}. Our evaluation method prompts GPT-3.5 to act as an expert assessor, analyzing the semantic alignment between predicted and ground-truth answers. For each QA pair, the system generates a similarity score on a scale of 0 to 100, where higher scores indicate closer alignment with the ground truth. 
To ensure interpretability, each score is accompanied by a detailed explanation justifying the rating. This approach provides transparent insights into the evaluation process while maintaining reproducibility. \cref{fig:prompts-GPT-Eval} demonstrates GPT evaluation prompt (\raisebox{-3.6pt}{\includegraphics[width=0.025\textwidth, height=0.025\textwidth]{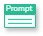}}).
For the Temporal Grounding task, we employ separate metrics (average mAP) better suited to temporal alignment assessment.

\noindent \textit{Evaluation Details}: For zero-shot evaluation of 18 models across 12 tasks on the test set, we utilized a cluster of 16 NVIDIA H100 GPUs. We maintained a batch size of 1 for all model evaluations and used the sampling frame rate and input video resolution parameters as recommended in their respective official repositories. The evaluation process was parallelized across multiple GPUs, resulting in an average evaluation time of 4 hours per model. 
For the closed-source models (Gemini 1.5 Pro [gemini-1.5-pro-latest], GPT-4o), the evaluation on 12 tasks required approximately two days per model due to API rate limitations.

Our evaluation period spanned October-November 2024, with specific access windows for different models: Gemini 1.5 Pro and GPT-4V (November 1-6, 2024). Claude 3.5 Sonnet was used from October 1-29 and GPT-3.5-turbo was used from October 18 to November 10, 2024.

\subsection{Qualitative Analysis}
\label{sec:qualitative}

The qualitative results are shown \cref{fig:QA_time_1} onwards. 

\begin{table}[!t]
    \centering
    \resizebox{\linewidth}{!}{
    \begin{tabular}{ccc}
        \hline
        \textbf{Fine-tuned} & \textbf{DriveLM Planning~\cite{sima2025drivelm}} & \textbf{Lingo-QA Eval~\cite{marcu2024lingoqa}} \\
        \hline
        \ding{56} & 31.7 & 37.0 \\
        \hline
        \ding{52} & 40.1 (\textcolor{darkred}{$+$8.3\%}) & 41.6 (\textcolor{darkred}{$+$4.6\%}) \\
        \hline
    \end{tabular}
    }
    \caption{Performance comparison of LLaVA-OV with and without fine-tuning on RoadSocial dataset.}
    \label{tab:videollm_comparison}
\end{table}

\subsection{RoadSocial's Utility for Planning/AV tasks} 
RoadSocial contains a significant number of egocentric road videos. Planning-related QAs in RoadSocial are distributed across multiple tasks, such as ``Advisory'', ``Counterfactual'', ``Description'' and ``Why'', covering critical and planning-related road events like dangerous driving, near-misses, and defensive driving. Similarly, perception-related QAs for such videos refer to ego-relative important objects (called ``Key Entities" in our dataset). To demonstrate the utility of the mentioned QA types for planning and perception tasks, we fine-tuned a Video-LLM (LLaVA-OV) \cite{li2024llava} on our dataset. We evaluated the model on representative autonomous driving benchmarks: PlanningQA task in DriveLM\cite{sima2025drivelm} and Action/Scenery QA task in Lingo-QA\cite{marcu2024lingoqa}. The substantial improvement in performance after fine-tuning on RoadSocial (\cref{tab:videollm_comparison}) demonstrates our dataset's utility for evaluating video-based planning/AV tasks.

\begin{table}[!t]

\begin{center}
\setlength{\tabcolsep}{3pt}
\newcommand{\applyHeatmap}[1]{%
    \ifdim #1 pt < 20pt \cellcolor{cyan!12}#1\else%
    \ifdim #1 pt < 40pt \cellcolor{cyan!30}#1\else%
    \ifdim #1 pt < 60pt \cellcolor{yellow!35}#1\else%
    \ifdim #1 pt < 80pt \cellcolor{orange!45}#1\else%
    \cellcolor{red!55}#1%
    \fi\fi\fi\fi
}
 \centering
    \resizebox{\linewidth}{!} {
\begin{tabular}{c|cccccccccccc|cccc}
\toprule
\multirow{2}{*}{Fine-tuning} & \multicolumn{3}{c}{Factual} & \multicolumn{4}{c}{Complex} & \multicolumn{3}{c}{Imaginative} & \multicolumn{2}{c|}{Hallucination} & Overall & Overall & Overall & Overall \\

 \cmidrule(rl){2-4} \cmidrule(rl){5-8} \cmidrule(rl){9-11} \cmidrule(rl){12-13}
 dataset & WR & KE & VP & DS & WY & CQ & TG & AD & IN & CF & AV & IC & (ALL) & (RT) & (Generic) & (Specific) \\[1.2pt]

\cmidrule(r){1-1} \cmidrule(rl){2-4} \cmidrule(rl){5-8} \cmidrule(rl){9-11} \cmidrule(rl){12-13} \cmidrule(rl){14-14} \cmidrule(rl){15-15} \cmidrule(rl){16-16} \cmidrule(l){17-17}

\cellcolor{PastePink} \textbf{RoadSocial} & \applyHeatmap{80.6} & \applyHeatmap{63.9} & \applyHeatmap{85.7} & \applyHeatmap{64.0} & \applyHeatmap{68.7} & \applyHeatmap{65.0} & \applyHeatmap{4.49} & \applyHeatmap{74.1} & \applyHeatmap{70.8} & \applyHeatmap{71.7} & \applyHeatmap{95.4} & \applyHeatmap{84.7} & \applyHeatmap{69.1} & \applyHeatmap{67.7} & \applyHeatmap{64.2} & \applyHeatmap{70.6} \\ 

\cellcolor{Gray} Video-only & \applyHeatmap{67} & \applyHeatmap{53.4} & \applyHeatmap{68.3} & \applyHeatmap{32.7} & \applyHeatmap{34.7} & \applyHeatmap{43.5} & \applyHeatmap{0.08} & \applyHeatmap{55.8} & \applyHeatmap{42.1} & \applyHeatmap{51.5} & \applyHeatmap{83.2} & \applyHeatmap{71.4} & \applyHeatmap{50.3} & \applyHeatmap{44.9} & \applyHeatmap{45.3} & \applyHeatmap{43.6} \\ 

\bottomrule
\end{tabular}
}
\end{center}

\caption{LLaVA-OV fine-tuned on RoadSocial 
 with Video and Commentary v/s fine-tuned on only Video-based QA dataset.}
\label{tab:video_only_roadsocial}
\end{table}

\subsection{Video-only QAs falls short} 
We regenerated entire QA dataset using only video-based summaries and used it to fine-tune LLaVA-OV model\cite{li2024llava}. We compared this model with the LLaVA-OV model described in paper (fine-tuned using QA obtained from video and social conversations). The performance gap (\cref{tab:video_only_roadsocial}) highlights the crucial role of social commentary in enhancing QA data quality.


\end{document}